\pdfoutput=1
\PassOptionsToPackage{table}{xcolor}
\documentclass[letterpaper]{article} 
\usepackage{aaai2027}  
\usepackage[hyphens]{url}  
\usepackage{graphicx} 
\usepackage{natbib}  
\usepackage{caption} 
\usepackage{pifont} 
\usepackage{multirow}

\usepackage{amsmath}
\DeclareMathAlphabet{\mathbbold}{U}{bbold}{m}{n}
\usepackage{amssymb}
\usepackage{booktabs}

\usepackage{amsmath,amsfonts,bm}

\def\eqref#1{equation~\ref{#1}}

\def\1{\bm{1}}

\def\rvg{{\mathbf{g}}}

\def\rvr{{\mathbf{r}}}

\def\rvv{{\mathbf{v}}}

\def\rvy{{\mathbf{y}}}

\def\rmF{{\mathbf{F}}}

\def\mF{{\bm{F}}}

\def\mM{{\bm{M}}}

\def\mS{{\bm{S}}}

\DeclareMathAlphabet{\mathsfit}{\encodingdefault}{\sfdefault}{m}{sl}
\SetMathAlphabet{\mathsfit}{bold}{\encodingdefault}{\sfdefault}{bx}{n}

\def\gD{{\mathcal{D}}}

\def\gG{{\mathcal{G}}}

\def\gL{{\mathcal{L}}}

\def\gT{{\mathcal{T}}}

\def\gV{{\mathcal{V}}}

\def\gX{{\mathcal{X}}}
\def\gY{{\mathcal{Y}}}

\def\sR{{\mathbb{R}}}

\newcommand{\E}{\mathbb{E}}

\nocopyright

\title{TRIG: Toward Measuring $\dot{\text{T}}$exture $\dot{\text{R}}$ealism in $\dot{\text{I}}$mage $\dot{\text{G}}$eneration}
\author{
    Zhiwei Xu\textsuperscript{1}\thanks{
        Work done during internship at Futurewei Technologies.
    },
    Jiali Cui\textsuperscript{2},
    Zhiqiang Lao\textsuperscript{2},
    Heather Yu\textsuperscript{2},
    Liang Peng\textsuperscript{2}\thanks{
        Corresponding author.
    }
}
\affiliations{
    \textsuperscript{1}Northwestern University\\
    \textsuperscript{2}Futurewei Technologies
}

\begin{document}

\maketitle

\begin{abstract}
This paper studies the backbone used by metrics for evaluating the quality of generated images. Although generative models have advanced rapidly, some outputs still exhibit unrealistic texture characteristics, and these differences are difficult to measure. Widely used evaluation metrics provide limited insight into texture realism, since they often rely on a backbone trained for another objective, such as image classification or image-text alignment. Such objectives typically encourage invariance to low-level variations that carry texture, so the resulting features primarily encode semantic and global structure. As a result, texture-specific discrepancies may not be explicitly captured. We propose TRIG, a metric directed at the texture realism of generated images. Its backbone is instead the discriminator of a generative adversarial network pre-trained on a texture dataset, which provides a learned prior for the local statistics of natural textures. A plausibility head is fine-tuned to identify local texture degradations, while preserving natural defocus and sensor noise as positive cases. On the features of this backbone, we compute a standard distributional distance between a reference set of real images and a set of generated ones. We apply TRIG to several generative models and observe closer agreement with human realness ratings than baseline metrics.

\end{abstract}


\section{Introduction}
Progress in image generation quality is largely measured by automated metrics. The standard approach compares a set of generated images with a set of real ones in the feature space of a pretrained network, and reports a distance between the two distributions. The Fréchet Inception Distance (FID) \cite{heusel2017gans} and the Kernel Inception Distance (KID) \cite{binkowski2018demystifying} follow this pattern with features from an image classifier, and CLIP Maximum Mean Discrepancy (CMMD) \cite{jayasumana2024rethinking} follows it with embeddings trained for image-text alignment. A separate line of work \cite{sajjadi2018assessing,kynkaanniemi2019improved} reports precision and recall in place of a single distance, in order to distinguish the fidelity of generated samples from their diversity. Although these methods differ in 
\begin{center}
\begin{minipage}{0.99\columnwidth}
\centering
\setlength{\tabcolsep}{2pt}
\begin{tabular}{ccccc}
&
\scriptsize Real &
\scriptsize DiT &
\scriptsize ADM &
\scriptsize StyleGAN-XL \\[2pt]
\raisebox{1.1\height}{\rotatebox[origin=c]{90}{\scriptsize Original Images}} &
\includegraphics[width=0.23\linewidth]{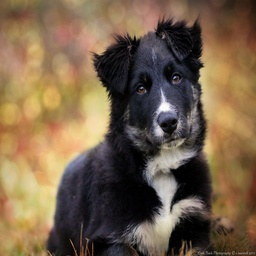} &
\includegraphics[width=0.23\linewidth]{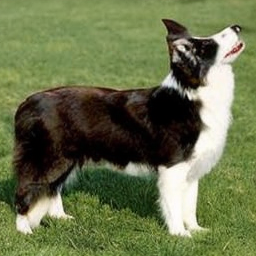} &
\includegraphics[width=0.23\linewidth]{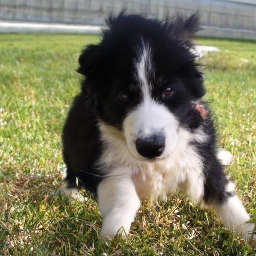} &
\includegraphics[width=0.23\linewidth]{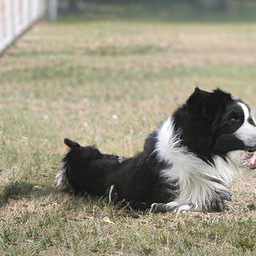} \\[0.5pt]
\raisebox{1.1\height}{\rotatebox[origin=c]{90}{\scriptsize Fourier Spectra}} &
\includegraphics[width=0.23\linewidth]{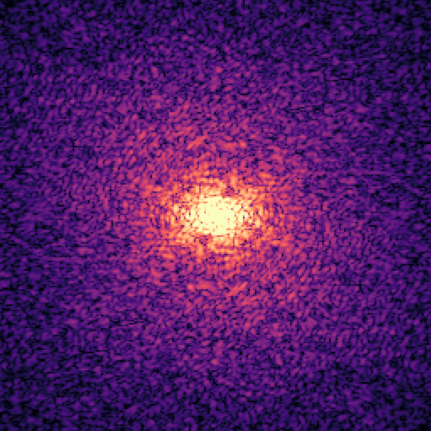} &
\includegraphics[width=0.23\linewidth]{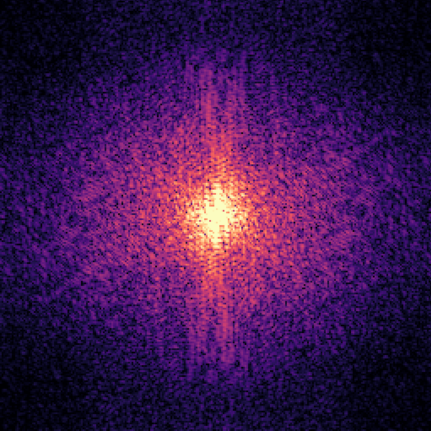} &
\includegraphics[width=0.23\linewidth]{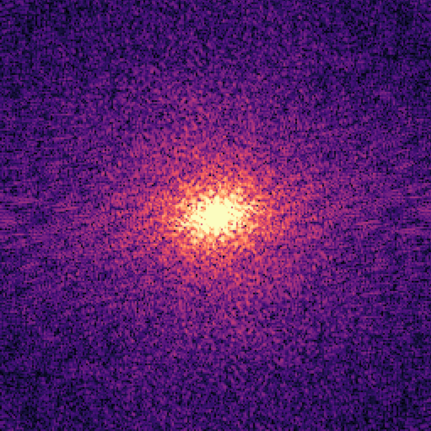} &
\includegraphics[width=0.23\linewidth]{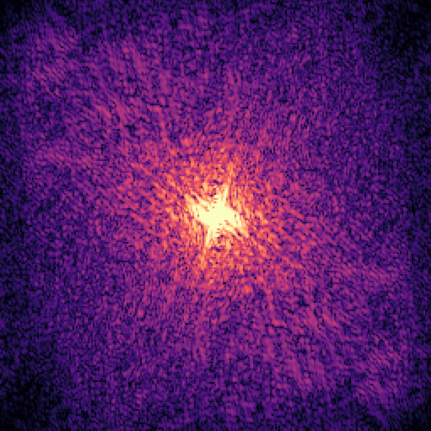}
\end{tabular}
\captionof{figure}{
Real and generated images with their Fourier spectra. Compared with natural photographs, generated images exhibit characteristic spectral artifacts,  particularly in the mid-to-high frequency range, despite appearing visually similar in the image domain.
}
\label{fig:fft_backbone_comparison}
\end{minipage}
\end{center}\vspace{20pt}
how they compare distributions, they primarily rely on representations learned for particular objectives, and thus influence which discrepancies between real and generated images the resulting metrics are sensitive to.

Modern generative models now reproduce scene content and global structure remarkably well, while often differing from photographs in the statistics of local texture. These differences contribute to a perceived lack of realism even when observers cannot point to their source, and they become more evident in the frequency domain (Figure~\ref{fig:fft_backbone_comparison}). Recent work \cite{stein2023exposing} has shown that the choice of representation is a major source of disagreement between automated metrics and human judgment, motivating alternatives based on self-supervised representations \cite{caron2021emerging,oquab2024dinov2} rather than image classification. Although these representations capture richer visual information, they remain optimized for objectives that encourage invariance to low-level appearance changes, including much of the variation that characterizes texture (see discussion in Section.~\ref{sec:current-backbone}). Recent texture-oriented metrics \cite{napolitano2026gram} likewise extract statistics from such backbones, which can be limited for assessing texture realism itself.

Texture realism, however, is not equivalent to sharpness. Real photographs naturally contain blur, sensor noise, and other imperfections arising from the image formation process: backgrounds fall outside the depth of field \cite{levoy1996light}, moving objects produce motion blur \cite{levin2008motion}, and photon-limited imaging introduces visible noise under low illumination \cite{foi2008practical}. Therefore, measuring whether the local texture is consistent with statistics derived from physical image formation remains challenging. 

Motivated by this perspective, we learn a representation specifically for texture realism instead of adapting one trained for another task. We begin with the discriminator of a generative adversarial network (GANs) \cite{goodfellow2014gan,liu2019texture,ePrime2023} trained on textures, whose objective is to distinguish real textures from synthesized ones. The early layers are kept frozen to preserve this learned texture prior, while a lightweight prediction head is trained to produce dense spatial features. Training such prediction head uses only real photographs, where synthetic degradations are adaptively applied. The degradations are chosen to produce texture statistics that are unlikely to arise through physical image formation. As such, the resulting representation produces, at each spatial location, a score reflecting consistency with learned natural-texture statistics. For evaluating generative models, these features are pooled within each image and compared across image sets using a distributional distance. We refer to the resulting metric as TRIG.

We evaluate TRIG on various generative models and compare it with widely used automatic metrics. We also examine our metric for agreement with two human studies, where in both settings, TRIG exhibits closer agreement with human judgments than the competing metrics. We further analyze the spatial responses of the learned representation, showing that both the image-level score and the texture-realism map identify regions whose texture can differ from that of real photographs. Finally, ablation studies examine the contribution of each component of the proposed method.

Our contributions are summarized as follows:
\begin{itemize}
    \item We introduce \textbf{TRIG}, a distributional metric that evaluates generative models through a representation learned specifically for texture realism.
    \item We propose a spatially supervised texture representation by adapting a texture GAN discriminator and fine-tuning it with localized synthetic degradations.
    \item We apply TRIG to various generative models and find that TRIG better agrees with human judgments of texture realism than widely used metrics, while providing spatial maps that localize inconsistent texture.
\end{itemize}

\section{Related Works}\label{sec-related-work}
\subsection{Metrics for image generation.} Most evaluation metrics compare generated and real images in the feature space of a pretrained network. The Inception Score (IS) \cite{salimans2016improved} summarizes the predictions of an image classifier, while FID~\cite{heusel2017gans} and KID~\cite{binkowski2018demystifying} compare feature distributions extracted by the same backbone using Fréchet and kernel distances, respectively. CMMD replaces the classification features with representations from a vision-language model. Complementary metrics report precision and recall, or density and coverage, to separately characterize sample fidelity and diversity \cite{kynkaanniemi2019improved,naeem2020reliable}.


Several studies examine the reliability of these metrics, and most target the \emph{Inception representation} they rely on rather than the comparison itself. \citet{parmar2022aliased} show that low-level operations such as aliased resizing and compression alter the extracted features and thus the reported score, and propose corrected preprocessing. \citet{chong2020effectively} show that the FID estimator is biased in a way that depends on the model being evaluated, and derive a debiased variant. CMMD~\citep{jayasumana2024rethinking} instead changes both components: it replaces Inception with CLIP features and the Fr\'echet distance with the Maximum Mean Discrepancy (MMD). These works patch the backbone, debias the estimator, or substitute a different generic backbone, but none learn
a representation tailored to texture; this is the gap our method addresses.

\subsection{The choice of feature extractor.} Recent work argues that the representation largely determines agreement with human perception. \cite{stein2023exposing} show that classification features correlate poorly with human judgments and recommend replacing them with self-supervised feature extractors, such as DINOv2-ViT-L/14~\cite{oquab2024dinov2}. Gram-MMD \cite{napolitano2026gram} instead targets texture by computing Gram statistics from intermediate features of a pretrained backbone selected through a search over layers. Despite these differences, all of these methods rely on representations learned for object recognition other than texture assessment. We instead train a representation specifically for texture realism by adapting a texture discriminator and fine-tuning it with supervision designed for this task.

\subsection{From texture representation to image quality assessment.} The statistical description of texture has a long history in image processing, originally developed for texture synthesis. \cite{portilla2000parametric} represent a texture by a compact set of multiscale wavelet statistics and re-synthesize textures that match them. \cite{gatys2015texture} and \cite{heitz2021sliced} instead extract features from a pretrained VGG~\cite{simonyan2014very} network, matching either their Gram matrices or, in the case of \cite{heitz2021sliced}, their full feature distributions via a sliced Wasserstein distance. In all cases the goal is to find a representation in which two textures that look alike are close, and to synthesize a texture that matches a target under that representation. This suggests a natural extension to evaluation: if similar textures yield similar representations, then the distance between representations can measure texture similarity. Gram-MMD follows this idea by computing MMD over Gram statistics extracted from selected layers of a pretrained backbone. We adopt the same premise, but argue that the representation should be learned for texture realism instead of a second-order statistic of features from a backbone trained for an unrelated semantic task.

\section{Methods}
\subsection{Texture Quality in Real Imaging}\label{sec:current-backbone}

We study the problem of assessing the texture quality of images produced by generative models. By \textit{texture quality}, we mean the consistency between local texture statistics in images and those in real images. A real image may contain a defocused background, motion blur from a moving subject, or sensor noise in a dimly lit region. These effects can arise naturally during image formation and therefore do not by themselves indicate unrealistic texture. Generated images may instead exhibit local texture discrepancies that are not explained by these natural effects, such as spatially inconsistent loss of high-frequency detail (see Figure.\ref{fig:fft_backbone_comparison} for examples with their Fourier spectra). The question we ask of an image is therefore not how sharp it is, but whether its texture is consistent with the surrounding visual context.

\begin{figure}[t]
\centering
\setlength{\tabcolsep}{1.pt}
\begin{tabular}{ccccc}
\scriptsize Input &
\scriptsize Inception &
\scriptsize CLIP &
\scriptsize DINOv2 &
\scriptsize Ours \\[2pt]
\includegraphics[width=0.185\linewidth]{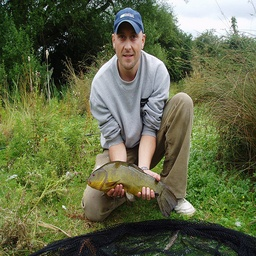} &
\includegraphics[width=0.185\linewidth]{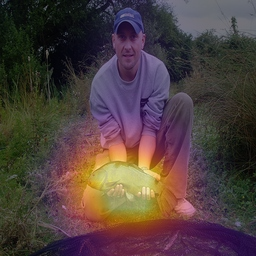} &
\includegraphics[width=0.185\linewidth]{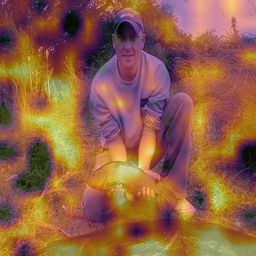} &
\includegraphics[width=0.185\linewidth]{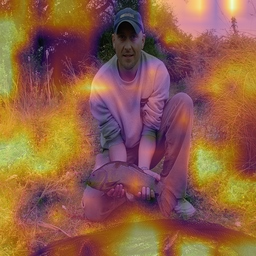} &
\includegraphics[width=0.185\linewidth]{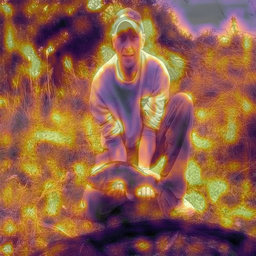} \\[-1pt]
\includegraphics[width=0.185\linewidth]{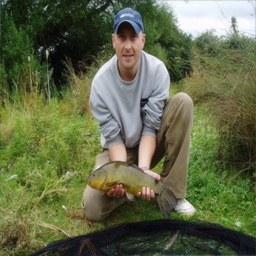} &
\includegraphics[width=0.185\linewidth]{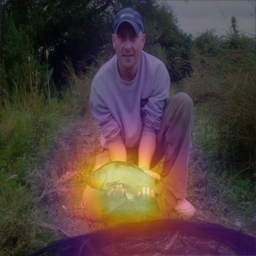} &
\includegraphics[width=0.185\linewidth]{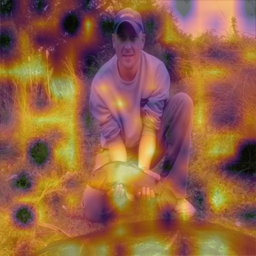} &
\includegraphics[width=0.185\linewidth]{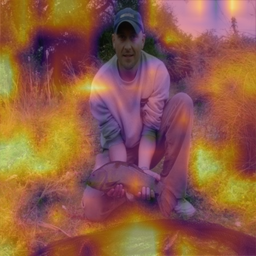} &
\includegraphics[width=0.185\linewidth]{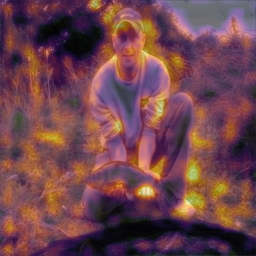} \\[-2pt]
\scriptsize $\Delta_{\mathrm{feat}}=$ &
\scriptsize 0.071 &
\scriptsize 0.088 &
\scriptsize 0.009 &
\scriptsize \textbf{0.131} \\
\end{tabular}
\caption{
Backbone sensitivity to high-frequency texture degradation. Under the same high-frequency suppression, semantic backbones remain relatively texture-invariant, whereas our texture-oriented backbone exhibits the largest feature change and strongest spatial activation redistribution.
}\vspace{10pt}
\label{fig:backbone_texture_sensitivity}
\end{figure}

\noindent\textbf{Current metric backbone.} Widely used metrics give limited access to this property, because the representations they compare were trained for other purposes. FID and KID compare InceptionV3 \citep{szegedy2016rethinking} features learned for object classification, and CMMD compares CLIP embeddings \cite{radford2021learning} learned for image–text alignment. These objectives reward invariance to the low-level variation that carries texture, so the resulting features encode mainly semantic and global structure. As shown in Figure.\ref{fig:backbone_texture_sensitivity}, under the same high-frequency degradation, the features of these backbones exhibit only minor change. As such, we intend to build a representation trained for this task that serves as a texture-aware evaluation protocol.

\subsection{A Backbone with a Texture Prior}

\begin{figure}[t]
\centering
\includegraphics[width=0.185\linewidth]{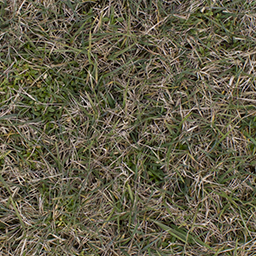} 
\includegraphics[width=0.185\linewidth]{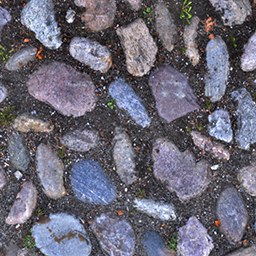} 
\includegraphics[width=0.185\linewidth]{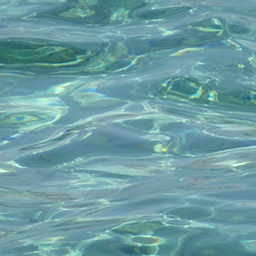} 
\includegraphics[width=0.185\linewidth]{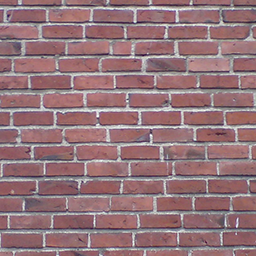} 
\includegraphics[width=0.185\linewidth]{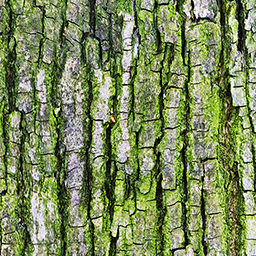}
\caption{Examples of texture images in ~\cite{lin2024dataset}.}
\label{fig:texture_dataset}
\end{figure}

\noindent\textbf{Texture discriminator.} 
A representation for texture quality should be sensitive to the local statistics that distinguish real texture from generated texture. In this work, we consider the discriminator of a generative adversarial network (GANs), which is trained on the real texture dataset \cite{lin2024dataset} and optimized to separate real texture from synthesized texture. As texture images contain little semantic content (see Figure.\ref{fig:texture_dataset} for examples), such a discriminator primarily captures local statistics instead of object semantics, and learned features encode the regularities of real textures, providing a learned prior for assessing texture quality. 

Let $\gD_\phi$ denote the discriminator. The texture images are spatially homogeneous and are cropped independently. This exposes the discriminator to a broad range of real texture statistics while presenting multiple perceptually equivalent but pixel-wise distinct views of the same texture. Together with a generator $\gG_\theta$, the discriminator is trained using the standard adversarial objective \cite{ePrime2023}
\begin{align}
    \gL(\phi)=\E_{z}\{\gD_\phi(\gG_\theta(z))\} - \E_{\gT}\{\E_{x|\gT}(\gD_\phi(x))\}
\end{align}
\begin{equation}
    \gL(\theta)=-\E_z\{\gD_\phi(\gG_\theta(z))\}
\end{equation}
where $\gT$ denotes a texture drawn from the training set and $x$ is a crop sampled from it. After training, $\phi$ is fixed, and the discriminator is used as a feature extractor. 

\begin{figure*}[t]
    \centering
    \includegraphics[width=0.99\linewidth]{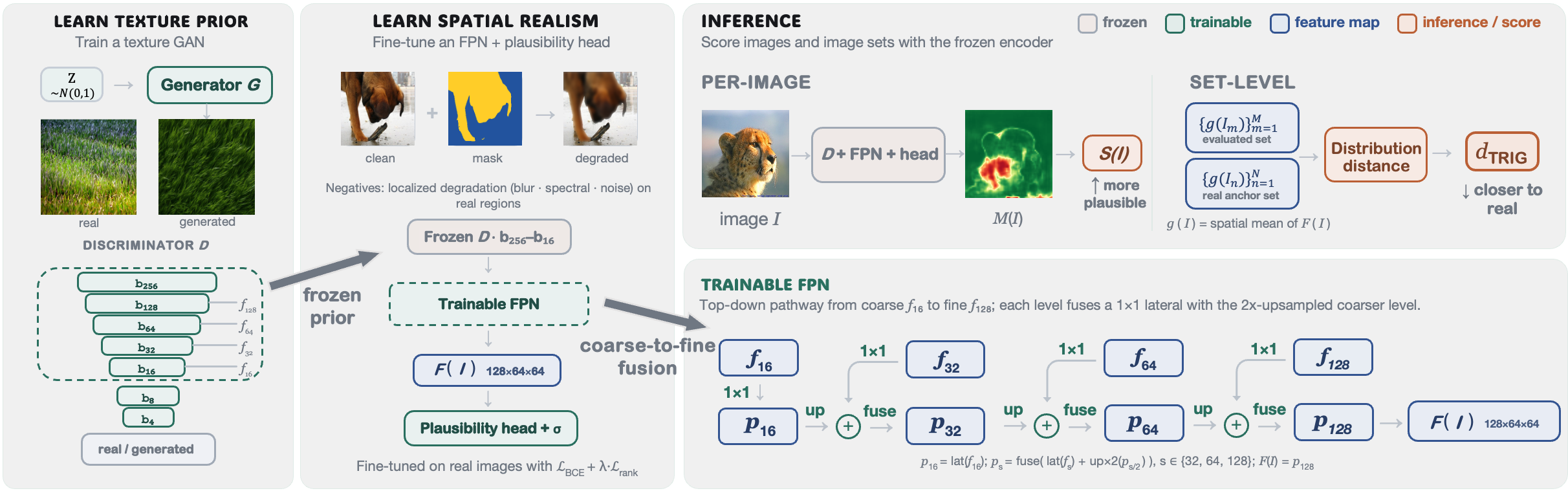}
    \caption{Overview of TRIG. A texture GAN discriminator provides a frozen texture prior. A feature pyramid and plausibility head are fine-tuned on localized synthetic degradations of real images. At inference, the network produces a texture-realism map for per-image scoring and a feature map whose spatial mean is used to compute the set-level distance. The feature pyramid fuses backbone features from coarse to fine.}
    \label{fig:illu}
\end{figure*}

\noindent\textbf{Which blocks to keep.} Figure.\ref{fig:illu} shows the architecture of $\gD_\phi$, which consists of a hierarchy of blocks with progressively decreasing spatial resolution. Prior work \citep{zeiler2014visualizing,luo2016understanding} has observed that as depth increases, the effective receptive field grows and the representation shifts from local image statistics to increasingly global structure, resulting in early blocks primarily capturing edges, color, and local texture, while deeper blocks encode object-level and semantic information. We therefore retain the early blocks, denoted $b_{256}$ through $b_{16}$, and discard the two deepest blocks, $b_8$ and $b_4$. We evaluate this design choice in the ablation study in Appendix.

\subsection{Learning Texture Prior}\label{sec-finetune}
The frozen backbone provides features that carry a texture prior, but not yet a measure of texture quality. We obtain such a measure by adding a trainable plausibility head on top of the frozen blocks and fine-tuning it to distinguish real texture from degraded texture.

\noindent\textbf{Feature pyramid and plausibility head.} Specifically, on top of the frozen blocks, we attach a top-down feature pyramid head. We take the intermediate features $\{f_{16}, f_{32}, f_{64}, f_{128}\}$, where the subscript indexes the selected block. We map each to $128$ channels via a lateral $1\times1$ projection, and we fuse them from coarse to fine, i.e.,
\begin{equation}
    p_{16}=\text{lat}_{16}(f_{16}),\;\;\;\; p_s=\text{fuse}_s(\text{lat}_{s}(f_s)+\text{Up}_{\times2}(p_{s/2}))
\end{equation}
where $s\in\{32, 64, 128\}$, and $\text{Up}_{\times2}$ is a factor-of-two upsampling operation, and $\text{fuse}_s$ is a small convolutional block. The output is a shared feature map $\rmF(I)=p_{128} \in \sR^{128\times64\times64}$. Fusing from coarse to fine lets each location in $\rmF(I)$ draw on context beyond the receptive field of the block alone. A plausibility head reduces $\rmF(I)$ to a map and is defined as
\begin{equation}
    \mM(I)=\text{sigmoid}(\text{Conv}(\text{act}(\text{Conv}(\rmF(I))))
\end{equation}
which predicts the probability that the texture at that location is consistent with real images. The feature pyramid and the plausibility head are the fine-tuned parts of the model.

\noindent\textbf{Degradation.} We construct supervision from real images to train the head: unmodified regions are positive, while synthetically perturbed regions serve as proxy negatives for contextually inconsistent texture alterations. 


The operators span three categories as discussed below: 

\noindent(\textit{i}) \textit{Blur}. The first category applies Gaussian, box, motion blur, and bicubic resampling. Natural defocus and motion blur follow scene depth and object motion and remain positive examples. Our negatives instead apply uniform blur within isolated semantic regions irrespective of depth, encouraging sensitivity to context-inconsistent over-smoothing rather than blur itself \cite{wang2023reconstruct,kamali2025characterizing}. Thus, depth-consistent bokeh should remain compatible with the learned natural-image statistics.

\noindent(\textit{ii}) \textit{Spectral}. The second category operates in the Fourier domain, reproducing alterations a blur cannot. Blur attenuates high frequencies smoothly and monotonically, while spectral operators \emph{selectively} suppress a chosen band, targeting the band-specific spectral deviation of generated images \cite{dzanic2020fourier,Durall_2020_CVPR,
corvi2023intriguing}; or alter \emph{phase} while preserving magnitude, following the primacy of phase in perceptual structure \cite{oppenheim1981importance}.

\noindent(\textit{iii}) \textit{Noise}. The third category adds Gaussian noise with constant variance. Sensor noise is a normal part of real imaging: its dominant photon-counting component is signal-dependent, scaling with local intensity \cite{foi2008practical}. However, uniform, constant-variance Gaussian noise is a negative example of a sensor model. We apply this degradation to model the spurious fine detail some generators introduce as fake sharpness.

We refer to details of these operators in Appendix.\\

\noindent\textbf{Regions and labels.} For each image, we first partition it into regions using SAM2 \cite{ravi2025sam}, where the output masks generally align with object and material boundaries and therefore separate regions of distinct texture. Each region is then either left unchanged or assigned a degradation operator from the pool. This defines the label as $\rvy=1$ for unchanged regions and $\rvy=0$ for degraded regions.

\begin{table*}[t]
\centering
    \resizebox{\textwidth}{!}{
    \setlength{\tabcolsep}{8pt}
    \setlength{\abovecaptionskip}{-5pt}
    \setlength{\belowcaptionskip}{-5pt}
    \renewcommand{\arraystretch}{1.15}
\begin{tabular}{lccccccc}
    \toprule
    \multirow{2}{*}{\textbf{Model}}
& \multicolumn{2}{c}{\textbf{FID} $\downarrow$}
& \multirow{2}{*}{\shortstack{\textbf{KID} $\downarrow$\\$(\times 10^{-3})$}}
& \multirow{2}{*}{\shortstack{\textbf{CMMD} $\downarrow$\\$(\times 10^{-4})$}}
& \multirow{2}{*}{\shortstack{\textbf{Gram-MMD}$^{1}$ $\downarrow$\\$(\times 10^{-2})$}}
& \multirow{2}{*}{\shortstack{\textbf{Gram-MMD}$^{2}$ $\downarrow$\\$(\times 10^{-2})$}}
& \multirow{2}{*}{\textbf{TRIG} $\downarrow$} \\
    & Paper & Reproduced &&&&\\
    \cmidrule(lr){2-3}
    ImageNet-eval & --- & 1.491
    & $0.145{\pm}0.039$ & $0.048$ & $1.512{\pm}0.035$ & $1.561{\pm}0.069$ & $0.105{\pm}0.001$ \\
    \midrule
    SiT-XL/2 (SDE) \cite{ma2024sit} & 2.06 & 2.017
    & $1.860{\pm}0.087$ & $2.785$ & $\mathbf{7.679{\pm}0.028}$ & $6.923{\pm}0.053$ & $0.485{\pm}0.008$ \\
    SiT-XL/2 (ODE) \cite{ma2024sit} & 2.15 & 2.024
    & $1.386{\pm}0.116$ & $3.098$ & $8.129{\pm}0.019$ & $6.797{\pm}0.065$ & $0.476{\pm}0.008$ \\
    DiT-XL/2-G \cite{peebles2023dit} & 2.27 & 2.175
    & $2.404{\pm}0.072$ & $\mathbf{2.753}$ & $8.320{\pm}0.032$ & $6.910{\pm}0.064$ & $0.490{\pm}0.008$ \\
    ADM-G \cite{dhariwal2021diffusion} & 4.59 & 4.487
    & $1.293{\pm}0.054$ & $2.869$ & $9.807{\pm}0.019$ & $7.569{\pm}0.043$ & $1.329{\pm}0.026$ \\
    \midrule
    SoFlow-XL/2 \cite{luo2025soflow} & 2.96 & 2.919
    & $1.014{\pm}0.055$ & $4.474$ & $9.399{\pm}0.029$ & $6.708{\pm}0.023$ & $0.478{\pm}0.008$ \\
    PixelMeanFlow-H/16 \cite{pixelmeanflows} & 2.11 & 2.300
    & $1.437{\pm}0.055$ & $2.496$ & $11.563{\pm}0.025$ & $8.787{\pm}0.017$ & $0.667{\pm}0.010$ \\
    SelfFlow \cite{CheferEsser2026selfflow} & 5.70 & 5.817
    & $1.127{\pm}0.053$ & $3.837$ & $7.751{\pm}0.026$ & $6.380{\pm}0.033$ & $0.632{\pm}0.011$ \\
    \midrule
    VAR-d30 \cite{tian2024var} & 1.92 & 1.954
    & $2.028{\pm}0.066$ & $4.697$ & $9.917{\pm}0.049$ & $7.825{\pm}0.040$ & $\mathbf{0.438{\pm}0.008}$ \\
    RAR-XXL \cite{yu2025rar} & \textbf{1.48} & \textbf{1.482}
    & $1.388{\pm}0.059$ & $7.241$ & $13.898{\pm}0.034$ & $9.177{\pm}0.052$ & $0.981{\pm}0.014$ \\
    LlamaGen-3B \cite{sun2024llamagen} & 2.18 & 2.135
    & $1.954{\pm}0.078$ & $3.070$ & $9.155{\pm}0.032$ & $7.783{\pm}0.019$ & $0.954{\pm}0.022$ \\
    \midrule
    StyleGAN-XL \cite{sauer2022styleganxl} & 2.30 & 2.370
    & $\mathbf{0.949{\pm}0.062}$ & $4.933$ & $7.979{\pm}0.016$ & $\mathbf{5.155{\pm}0.035}$ & $1.194{\pm}0.030$ \\
    BigGAN-deep \cite{brock2019biggan} & 6.95 & 6.897
    & $3.248{\pm}0.042$ & $10.762$ & $12.904{\pm}0.034$ & $7.774{\pm}0.067$ & $1.175{\pm}0.019$ \\
    \bottomrule
    \end{tabular}
    }
    \caption{
Distances between each generative model and the ImageNet-anchor reference set under each metric. FID and KID use Inception-v3 \cite{szegedy2016rethinking} features, CMMD uses CLIP embeddings \cite{radford2021learning}, Gram-MMD$^1$ uses DINOv2 \cite{oquab2024dinov2} features, and Gram-MMD$^2$ uses DC-AE \cite{chen2025dcae} features. The ImageNet-eval row reports the distance between two disjoint sets of real images. Reported values are averaged over five repeated evaluations. For KID, repeated runs account for the random sampling of 1,000-image subsets as in the original implementation. For Gram-MMD, repeated runs arise from randomly sampling subsets of the 50k images to reduce memory usage. CMMD is deterministic, as it compares all 50k CLIP features. For TRIG, repeated runs arise from the random projection directions used in the sliced Wasserstein distance. Bold indicates the smallest distance among the generative models.
}
\label{tab:metric_compare}
\end{table*}
\noindent\textbf{Fine-tuning objective.} The head is trained with a binary cross-entropy loss, i.e., $\gL_{\text{BCE}}$, applied at each spatial location.
\begin{equation}
    -\E_{(u,v)}[\rvy_{u,v}\log\mM(I)_{u,v}+(1-\rvy_{u,v}\log(1-\mM(I)_{u,v}))]
\end{equation}
where the expectation is taken over the spatial locations of $\mM(I)$. 
This loss teaches the head to detect whether a degradation operator has been applied, but not its severity. To encourage the representation to reflect degradation strength, we introduce a ranking loss. For two regions $r$ and $r’$ in the same image that receive the same operator with strengths $\alpha_r$ and $\alpha_{r'}$ satisfying $|\alpha_r-\alpha_{r’}|\ge\delta$, the more heavily degraded region is required to receive a lower score by a margin $\eta$:
\begin{equation}
    \gL_{\text{rank}}=\sum_{
    (\rvr,\rvr')}\max\left(0, \mS(\rvr)-\mS(\rvr')+\eta\right)
\end{equation}
\begin{equation}\label{eq-single-score}
    \text{where}\quad\quad\quad\mS(I)=\E_{(u,v)\in R}\mM(I)_{u,v}
\end{equation}
in which $\mS(I)$ computes the average over a set of region $R$. The comparisons is restricted to regions that received the same operator, since the strengths of different operators are not on a common scale. The total objective is
\begin{equation}
    \gL_{\text{total}}=\gL_{\text{BCE}} + \lambda\cdot\gL_{\text{rank}}
\end{equation}
Ablation studies for the choice of $\lambda,\delta,\eta$ are in Appendix.

\noindent\textbf{From synthetic degradations to generated images.} The supervision above is built from real images and synthetic operators, while the metric is applied to generated images. This rests on the assumption that the two kinds of departure from real texture are related closely enough for a representation trained on the first to be informative about the second. In this work, we provide various empirical evaluations for demonstration (e.g.,  agreement with human rating in Section.\ref{sec-human-agreement}).

\newpage
\subsection{The TRIG Metric}
With the learned backbone, we obtain the feature map $\mF(I)$ that provides a description of the texture at each location. When comparing the output of a generative model against real images, these statistics are of interest. In particular, given an image $I$, we compute
\begin{equation}
    \rvg(I)=\E_{(u,v)}[\mF(I)_{u,v}]
\end{equation}
where the expectation spatially averages the multiscale texture responses into a fixed-length image descriptor.

Let $\gX=\{I_1,\dots,I_{|\gX|}\}$ be a set of images to be evaluated, and $\gY=\{I'_1,\dots,I'_{|\gY|}\}$ a reference set of real images. The metric compares the two sets through the distributions of their descriptors. 

We employ the sliced Wasserstein distance (SWD), which compares two distributions through their one-dimensional projections \citep{bonneel2015sliced,heitz2021sliced}. Specifically, let $\gV=\{\rvv_1,\dots,\rvv_K\}$ be directions drawn from the unit sphere, and projecting the descriptors of each set onto a direction $\rvv_k$ gives two sets of scalars,
\begin{equation}
d_{\text{TRIG}}(\gX,\gY)=\frac{1}{K}\sum_{k=1}^{K}W\left(
\{\rvv_k^{\top}\rvg(I_i)\}_{i=1}^{|\gX|},
\{\rvv_k^{\top}\rvg(I'_j)\}_{j=1}^{|\gY|}
\right)
\end{equation}
where $W$ denotes the one-dimensional Wasserstein distance, and $K$ controls the variance of the estimate. In one dimension, this amounts to sorting the two sets of projections and comparing them in order.

TRIG is this quantity, and a \textit{lower} value indicates texture statistics \textit{closer} to those of real images. The maximum mean discrepancy is a further nonparametric option, adopted by CMMD and by Gram-MMD. It requires a kernel bandwidth whose value affects the resulting score and is commonly set by heuristic \cite{napolitano2026gram}. We report SWD in the main experiments. We also conduct ablation studies for choices of metrics used in Section.\ref{sec:ablation}.

\begin{figure*}[t]
    \centering
    \setlength{\tabcolsep}{1pt}
    \renewcommand{\arraystretch}{0.6}
    \includegraphics[width=0.163\linewidth]{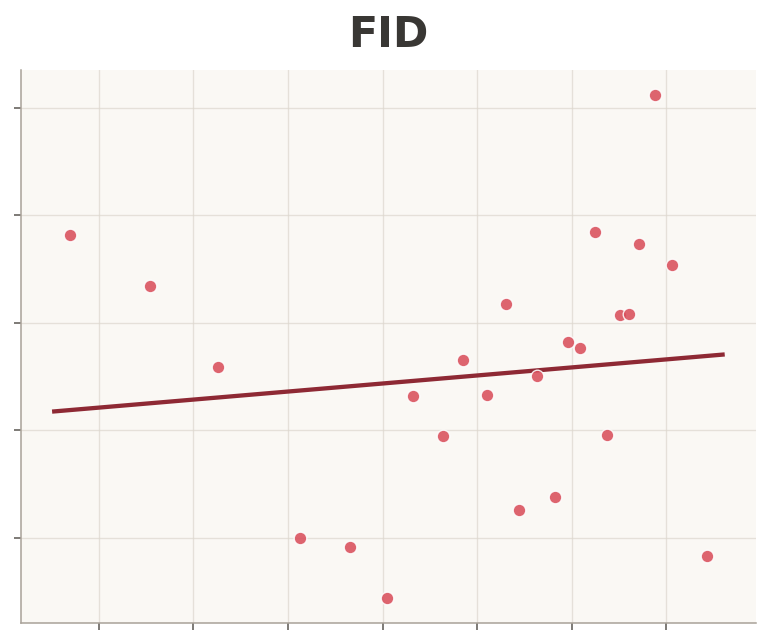}
    \includegraphics[width=0.163\linewidth]{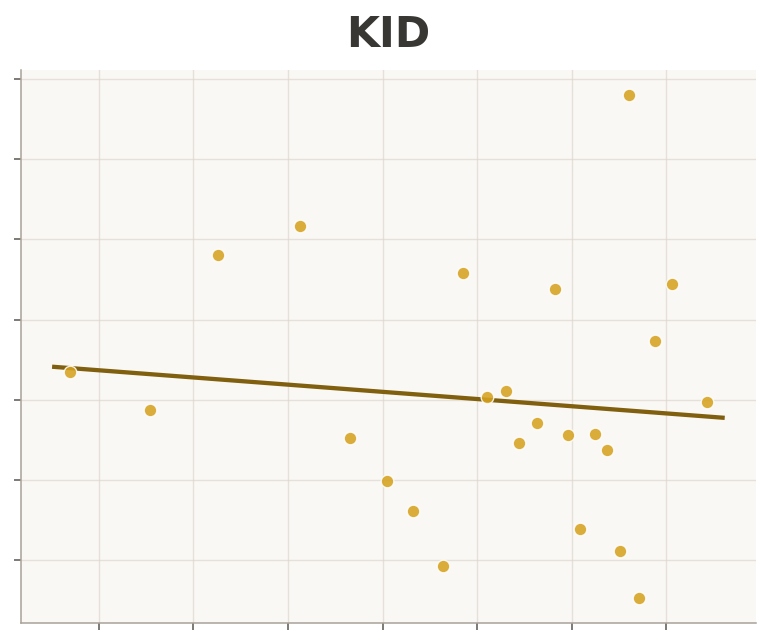}
    \includegraphics[width=0.163\linewidth]{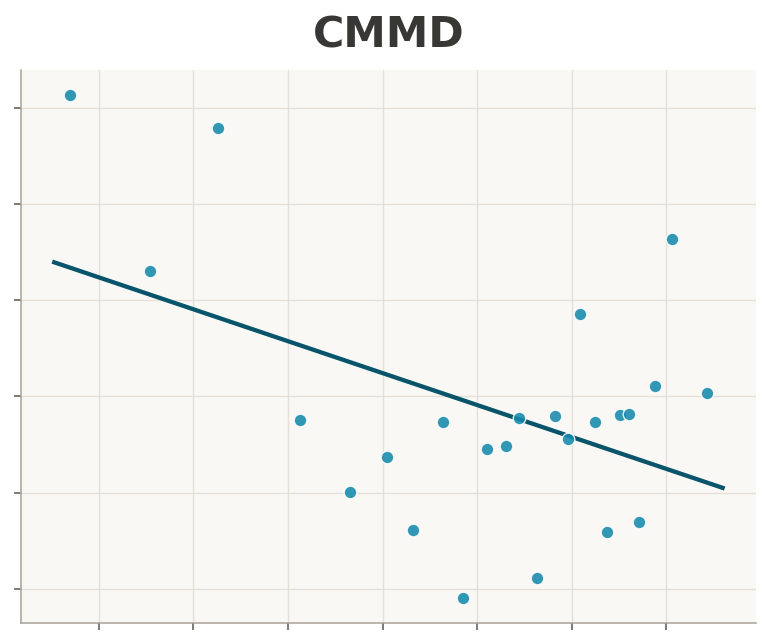}
    \includegraphics[width=0.163\linewidth]{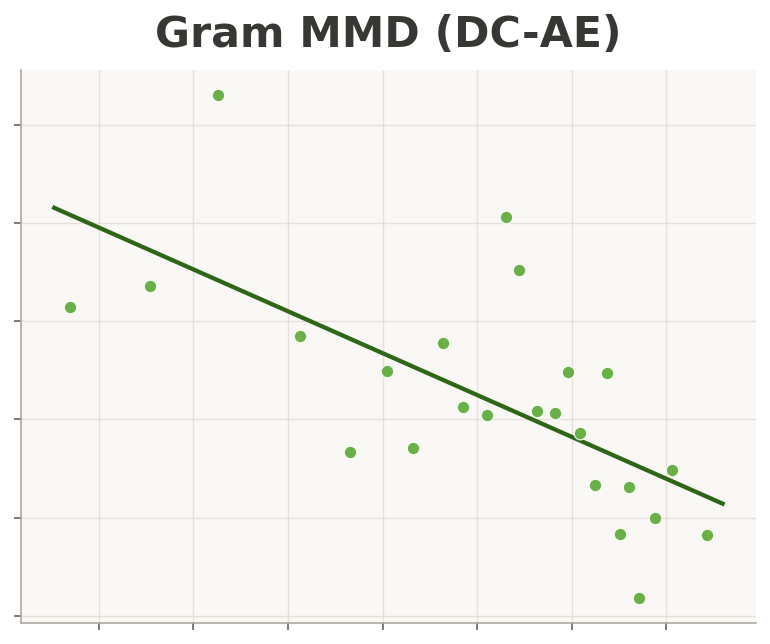}
    \includegraphics[width=0.163\linewidth]{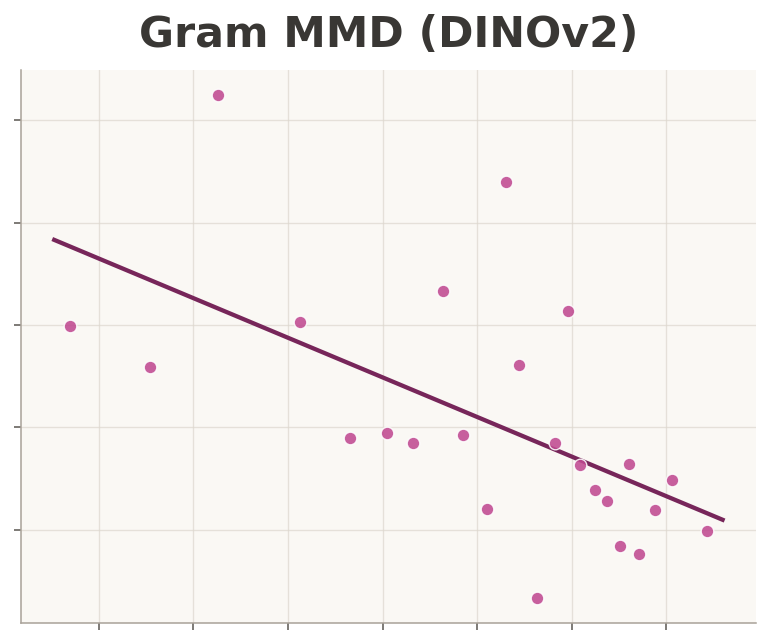}
    \includegraphics[width=0.163\linewidth]{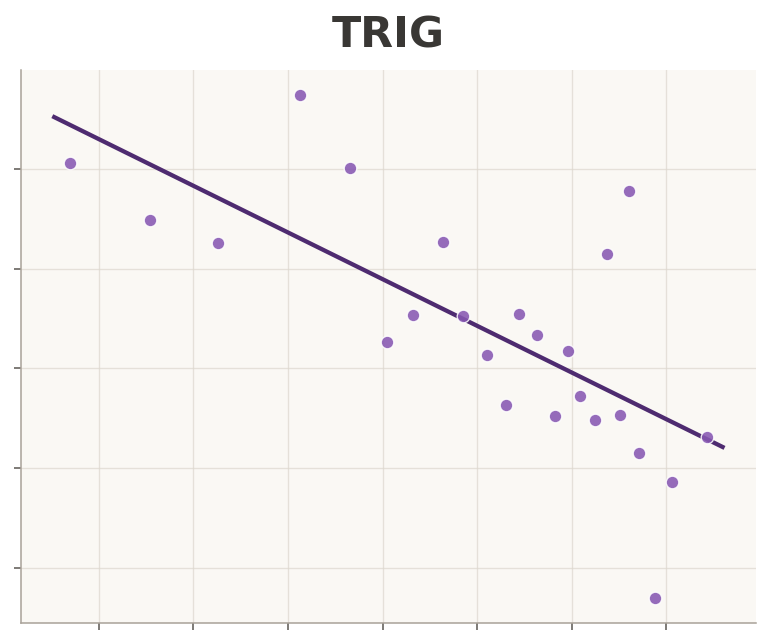}
    \includegraphics[width=1.0\linewidth]{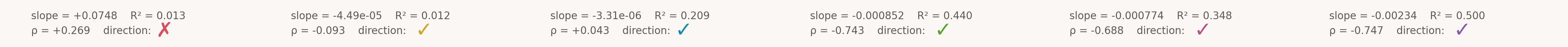}
    \caption{Each metric plots against human mean opinion score (MOS) on the RAISE realness benchmark ($n=24$ groups). 
    A texture-aware distance metric is expected to correlate with perceived realism (MOS) negatively. FID increases rather than decreases with MOS, whereas TRIG exhibits the strongest monotonic relationship with human judgments.}
    \label{fig:exp3_scatter}
\end{figure*}
\section{Experiment}
\subsection{Scores Across Generative Advances} 

We compare various advanced generative models trained on ImageNet $256\times256$ \cite{deng2009imagenet}, including diffusion models, flow-based models, autoregressive models, and GANs. We generate 50,000 images from each model. ImageNet-anchor is the real reference set against which every distance is computed, and ImageNet-eval provides a second real set, so that the distance between the two real sets shows the measured discrepancy on real data. Table.\ref{tab:metric_compare} reports the results. TRIG is shown with the SWD by default. Under every metric, the two real sets are the closest pair, and each generator receives a larger distance from the reference. More experimental details can be found in Appendix.

\subsection{Agreement with Human Realism Rating}
\label{sec-human-agreement}
In Table.\ref{tab:metric_compare}, we apply TRIG to various generative models. Whether these distances agree with human perception of texture realism is a separate question, on which we examine through two human studies in this section. We first adopt RAISE \cite{mukherjee2025raise}, a dataset targets overall perceived realness rather than semantic contents. Each image carries a mean opinion score (MOS) from a human study, collected under instructions to judge realness independently of semantic content, making the dataset well-suited to a texture-oriented metric. Following the protocol of~\cite{napolitano2026gram}, we order images by MOS into 24 groups and compute each distance against a fixed anchor of 1000 COCO images, reporting Spearman
and Kendall $\tau$ agreement against FID, KID, CMMD, and Gram-MMD (DINOv2 and DC-AE variants).

\begin{center}
\begin{minipage}{0.99\columnwidth}
\centering
\resizebox{0.99\columnwidth}{!}{%
\begin{tabular}{lccccc}
\toprule
\multirow{2}{*}{Metric} & \multicolumn{2}{c}{RAISE} & \multicolumn{3}{c}{ImageNet Study}\\
\cmidrule(lr){2-3}\cmidrule(lr){4-6}
 & Spearman & Kendall & Spearman & Kendall & 95\% CI (Spearman)\\
\midrule
FID           & $+0.269$ & $+0.232$ & $+0.467$ & $+0.278$ & $[+0.300,+0.517]$\\
KID           & $-0.093$ & $-0.101$ & $-0.100$ & $-0.056$ & $[-0.300,+0.033]$\\
CMMD          & $+0.044$ & $+0.087$ & $-0.683$ & $-0.500$ & $[-0.800,-0.617]$\\
Gram-MMD$^1$  & $-0.688$ & $-0.500$ & $-0.600$ & $-0.444$ & $[-0.767,-0.467]$\\
Gram-MMD$^2$  & $-0.743$ & $-0.558$ & $-0.283$ & $-0.222$ & $[-0.433,-0.167]$\\
\midrule
\textbf{TRIG} & $\mathbf{-0.747}$ & $\mathbf{-0.594}$ & $\mathbf{-0.750}$ & $\mathbf{-0.611}$ & $\mathbf{[-0.817,-0.550]}$\\
\bottomrule
\end{tabular}%
}\vspace{2pt}
\captionof{table}{Agreement between each metric and human ratings (RAISE MOS / ImageNet aggregated ranking), with bootstrap $95\%$ CIs (Spearman) for the ImageNet study.
Gram-MMD$^{1}$/Gram-MMD$^{2}$ use DINOv2 / DC-AE features. Note that FID correlates in the wrong (positive) direction.}
\label{tab:correlation}
\end{minipage}
\end{center}

A metric that reflects human perception should assign smaller distances to images judged more realistic, yielding a stronger negative correlation. As shown in Table.\ref{tab:correlation}, FID, KID, and CMMD exhibit only weak correlation with human ratings, with FID even showing the opposite trend. In contrast, the two texture-oriented metrics negatively correlate with perceived realism. Figure.\ref{fig:exp3_scatter} shows the same pattern, where the distances produced by FID, KID, and CMMD vary little with the mean opinion score, whereas those of TRIG decrease consistently as perceived realism increases.

Next, to evaluate a subset of the generators in Table\ref{tab:metric_compare}, we performed a human study on their generated images. We recruited 24 independent participants to compare image pairs and select the one with more realistic texture; We collected 4800 pairwise preferences and aggregated them into a ranking of sources. We then measure each metric's agreement with this ranking computed against a fixed anchor of 1000 ImageNet test images, reporting the same coefficients and a Spearman bootstrap CI. As shown in Table\ref{tab:correlation}, TRIG attains the strongest point-estimate agreement, its CI
excluding zero and lying below those of FID, KID, and Gram-MMD. Across the two studies, TRIG more closely reflects human judgments of texture realism than the competing metrics. Additional details of our human study are provided in the Appendix.

\subsection{Where Our Backbone Responds}
\begin{figure}[t]
\centering
\includegraphics[width=\linewidth]{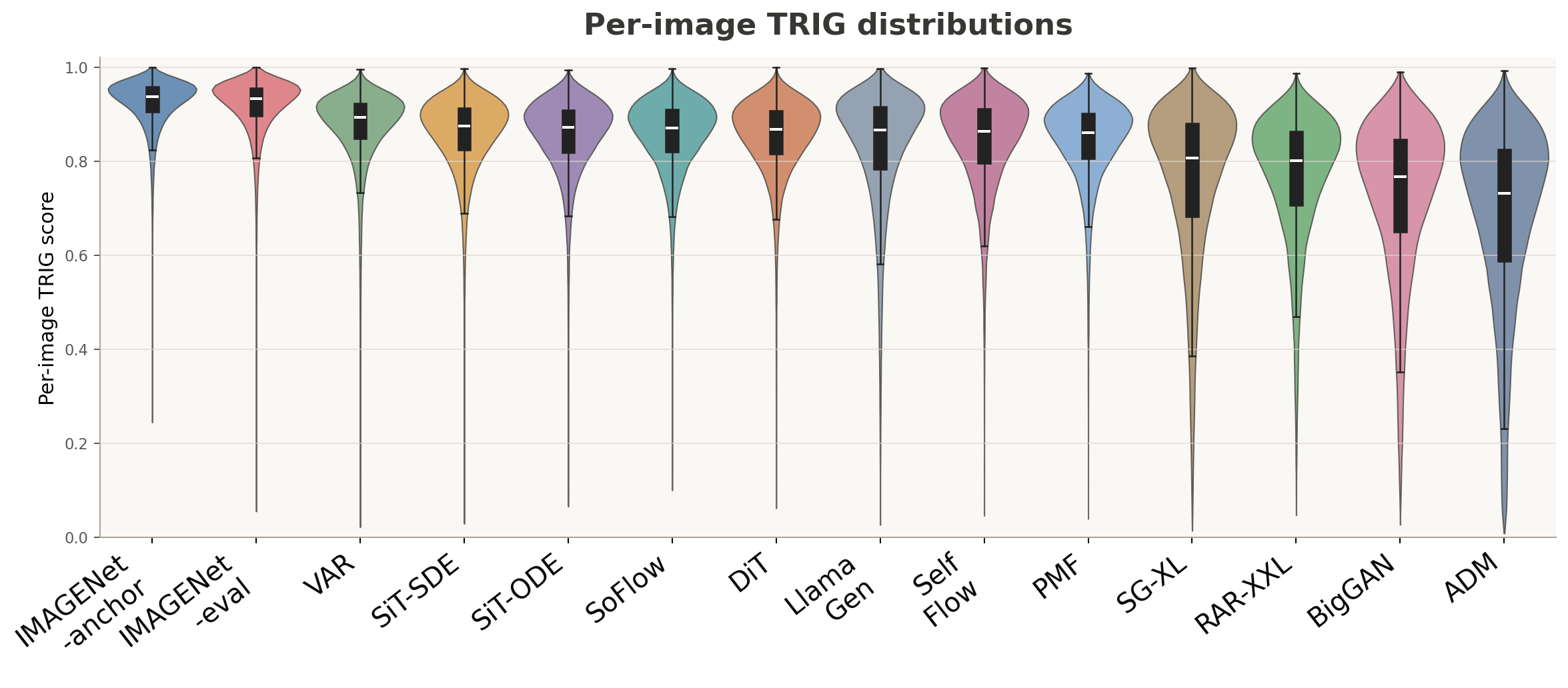}
\vspace{-15pt}
\captionof{figure}{Distribution of the score $\mS(I)$. Each violin shows the kernel density for one set of images, with the inner box marking the interquartile range and the median.}
\label{fig:violin_TRIG}
\vspace{-10pt}
\end{figure}

In this section, we probe where our backbone responds within an image. To assess the spatial distribution of texture realism, we compute $\mS(I)$ (Eq.\ref{eq-single-score}), which is the spatial average of the texture-realism map. Figure.\ref{fig:violin_TRIG} shows the distribution of $\mS(I)$. We observe that the two real sets have the highest medians and the smallest spread, while generated images have a lower median and a wider distribution, with StyleGAN-XL, BigGAN, and ADM rendering notably larger variance. 

We also compute $\mS(I)$ separately over foreground and background regions extracted with Rembg~\cite{gatis_rembg}. Their scores are similar for real images, whereas generated images score lower in the foreground, suggesting stronger departures from natural texture statistics on primary objects (Table~\ref{tab:exp2_sharp}). The corresponding texture-realism maps $\mM(I)$ show uniformly high responses for real images but localized foreground responses for generated images (Figure~\ref{fig:exp2_heatmaps}), further indicating sensitivity to local texture characteristics.

\begin{center}
\begin{minipage}{\columnwidth}
\centering

  \resizebox{\columnwidth}{!}{
    \small
    \setlength{\tabcolsep}{5pt}
    \setlength{\abovecaptionskip}{-5pt}
    \setlength{\belowcaptionskip}{-5pt}
    \begin{tabular}{@{}llcccc@{}}
    \toprule
    Category & Model & full$\uparrow$ & fg$\uparrow$ & bg$\uparrow$ & $\Delta_{\text{fg}-\text{bg}}$ \\
    \midrule
    \multirow{2}{*}{Real}
     & ImageNet-anchor  & \textbf{0.922} & \textbf{0.932} & \textbf{0.918} & $+0.014$ \\
     & ImageNet-eval   & 0.915 & 0.925 & 0.911 & $+0.015$ \\
    \midrule
    \multirow{4}{*}{Diffusion model}
     & SiT-XL/2(SDE)  & 0.855 & 0.833 & 0.858 & $-0.025$ \\
     & SiT-XL/2(ODE)  & 0.852 & 0.827 & 0.856 & $-0.029$ \\
     & DiT-XL/2-G       & 0.849 & 0.824 & 0.853 & $-0.029$ \\
     & ADM-G       & 0.683 & 0.663 & 0.690 & $-0.026$ \\
    \midrule
    \multirow{3}{*}{Flow model}
    & SoFlow-XL/2     & 0.854 & 0.826 & 0.859 & $-0.033$ \\
    & PMF-H/16        & 0.840 & 0.818 & 0.844 & $-0.026$ \\
    & SelfFlow   & 0.840 & 0.804 & 0.846 & $-0.042$ \\
    \midrule
    \multirow{3}{*}{Auto Regressive}
    & VAR-d30        & 0.870 & 0.863 & 0.869 & $-0.006$ \\
    & RAR-XXL        & 0.766 & 0.742 & 0.771 & $-0.029$ \\
    & LlamaGen-3B   & 0.828 & 0.812 & 0.829 & $-0.017$ \\
    \midrule
    \multirow{2}{*}{GAN}
     & StyleGAN-XL & 0.758 & 0.737 & 0.762 & $-0.025$ \\
     & BigGAN-deep      & 0.726 & 0.691 & 0.735 & $-0.044$ \\
    \bottomrule
    \end{tabular}
    }
\captionof{table}{Single-image texture-consistency score. 
}
\label{tab:exp2_sharp}
\end{minipage}
\end{center}

\subsection{Artificial Degradation}
Previous experiments visualize texture-realism maps on generated images, where the locations of texture artifacts are unknown. To further provide a controlled localization sanity check of the maps, we introduce known degradations and examine whether our backbone provides effective responses. 

\begin{center}\vspace{10pt}
\begin{minipage}{0.99\columnwidth}
\centering
\setlength{\tabcolsep}{2pt}
\begin{tabular}{ccccc}
\includegraphics[width=0.23\linewidth]{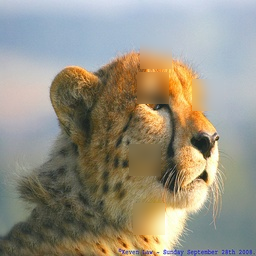} &
\includegraphics[width=0.23\linewidth]{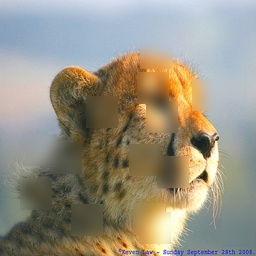} &
\includegraphics[width=0.23\linewidth]{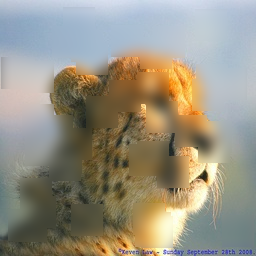} &
\includegraphics[width=0.23\linewidth]{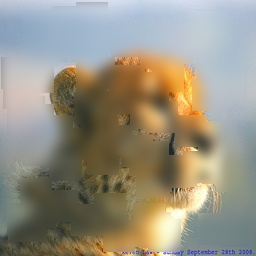} &
\raisebox{1.1\height}{\rotatebox[origin=c]{270}{\scriptsize Blurred images}}
\\
\includegraphics[width=0.23\linewidth]{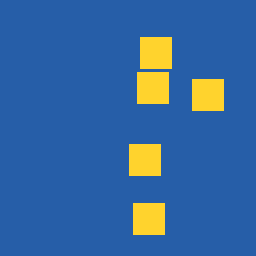} &
\includegraphics[width=0.23\linewidth]{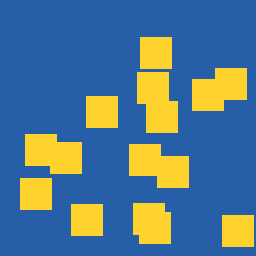} &
\includegraphics[width=0.23\linewidth]{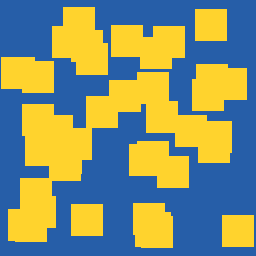} &
\includegraphics[width=0.23\linewidth]{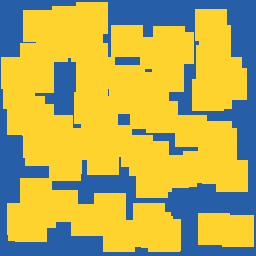} &
\raisebox{1.1\height}{\rotatebox[origin=c]{270}{\scriptsize Ground truth}}
\\
\includegraphics[width=0.23\linewidth]{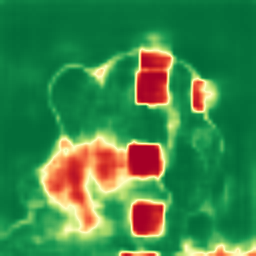} &
\includegraphics[width=0.23\linewidth]{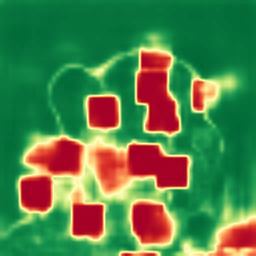} &
\includegraphics[width=0.23\linewidth]{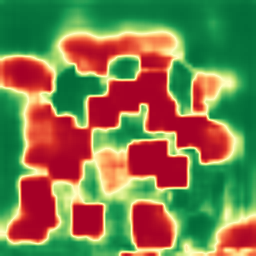} &
\includegraphics[width=0.23\linewidth]{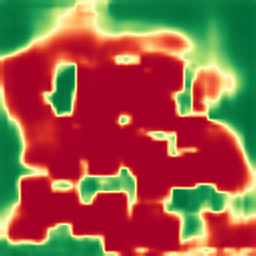} &
\raisebox{2.0\height}{\rotatebox[origin=c]{270}{\scriptsize Heatmap}}
\end{tabular}
\captionof{figure}{
Gaussian blur applied at increasing coverage.
}
\label{fig:manual_attack}
\end{minipage}
\end{center}\vspace{10pt}

Specifically, we apply Gaussian blur to real images with increasing levels of coverage. Figure.\ref{fig:manual_attack} presents results. It can be observed that the resulting map assigns lower scores to the degraded regions while remaining consistent over other regions. Given the known blur regions, we find our backbone can effectively respond to the degradation. More results and held-out degradations can be found in the Appendix.


\begin{figure}[t]
\centering
\setlength{\tabcolsep}{1pt}
\renewcommand{\arraystretch}{0.4}
\begin{tabular}{@{}c@{\hspace{2pt}}*{4}{c}@{}}
 & \parbox[c]{0.105\textwidth}{\centering\scriptsize ImageNet} &
\parbox[c]{0.105\textwidth}{\centering\scriptsize SoFlow} &
\parbox[c]{0.105\textwidth}{\centering\scriptsize SiT(ODE)} &
\parbox[c]{0.105\textwidth}{\centering\scriptsize StyleGAN-XL} \\
\end{tabular}
\begin{tabular}{@{}c@{\hspace{2pt}}cccc@{}}
\raisebox{0.5\height}{\rotatebox{90}{\scriptsize Original}} &
\includegraphics[width=0.105\textwidth]{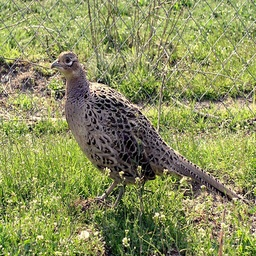} &
\includegraphics[width=0.105\textwidth]{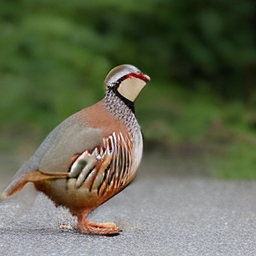} &
\includegraphics[width=0.105\textwidth]{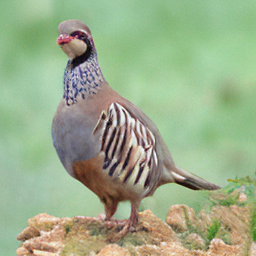} &
\includegraphics[width=0.105\textwidth]{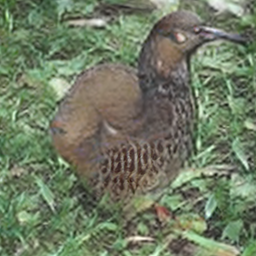} \\
\raisebox{0.5\height}{\rotatebox{90}{\scriptsize $+$Heatmap}} &
\includegraphics[width=0.105\textwidth]{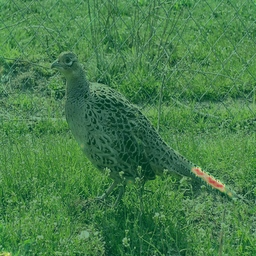} &
\includegraphics[width=0.105\textwidth]{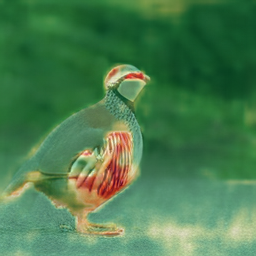} &
\includegraphics[width=0.105\textwidth]{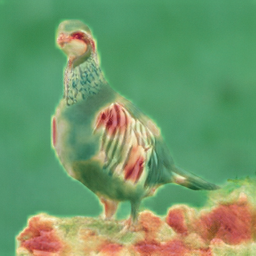} &
\includegraphics[width=0.105\textwidth]{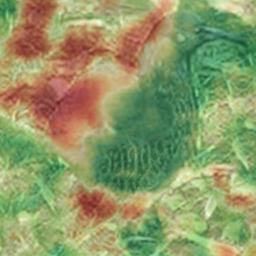}
\end{tabular}
\caption{Texture-realism maps by our backbone. The top row shows the original images and the bottom row the corresponding heatmap (overlaped). Red color highlights regions that deviate from the natural texture statistics detected. The real image remains nearly uniform, whereas generated images show responses on textured regions.}
\label{fig:exp2_heatmaps}
\end{figure}

\section{Ablation Study}\label{sec:ablation}
Table~\ref{tab:ablation} reports the correlation with the RAISE benchmark and our ImageNet human study for several variants of TRIG.

\noindent\textbf{Pre-trained data.} The backbone is a discriminator trained on textures, and we examine how pre-training data affect the performance. We observed that pretraining on textures consistently gives the strongest agreement with human judgment, whereas random initialization or pretraining on other datasets  (e.g., FFHQ \cite{karras2019style}), even directly on ImageNet, reduces performance. This therefore suggests the importance of our learned texture prior.

\noindent\textbf{Distribution distance.} We next fix the representation and vary the distance used to compare image sets. The SWD and MMD produce comparable results, whereas the Fréchet distance (FD) performs worse. The similar behavior of the two nonparametric distances suggests that agreement with human judgment is determined primarily by the learned representation rather than by the particular distributional distance.

\noindent\textbf{Degradation.} Finally, we examine the supervision used to train the plausibility head. Removing the noise degradations reduces agreement on RAISE and reverses the sign on ImageNet Study, indicating that noise provides an informative supervision. Removing blur also lowers agreement on both studies, while removing the frequency operators has a smaller effect; both contribute, but less critically than noise. Applying degradations without region masks likewise reduces agreement, suggesting that preserving object and material boundaries helps the representation learn realistic local texture. These results indicate that all degradation type and the spatial placement contribute to the learned representation.

We refer to more ablation studies of design choices in Appendix.
\begin{table}[t]
\centering
  \resizebox{0.92\columnwidth}{!}{
\setlength{\tabcolsep}{6pt}
\renewcommand{\arraystretch}{1.15}
\begin{tabular}{lcccc}
\toprule
\textbf{Variant} & \multicolumn{2}{c}{RAISE} & \multicolumn{2}{c}{ImageNet Study} \\
 & Spearman & Kendall & Spearman & Kendall \\
\midrule
TRIG (main)          & $-0.747$ & $-0.594$ & $-0.750$ & $-0.611$ \\
\midrule
\rowcolor{gray!15}
\multicolumn{5}{c}{\textit{Pre-trained Data}} \\
Random               & $-0.577$ & $-0.420$ & $+0.300$ & $+0.333$ \\
FFHQ                 & $-0.558$ & $-0.406$ & $-0.667$ & $-0.500$ \\
ImageNet             & $+0.349$ & $+0.225$ & $+0.283$ & $+0.222$ \\
\midrule
\rowcolor{gray!15}
\multicolumn{5}{c}{\textit{Distribution distance}} \\
TRIG-FD              & $-0.590$ & $-0.435$ & $-0.600$ & $-0.444$ \\
TRIG-MMD             & $-0.770$ & $-0.645$ & $-0.767$ & $-0.611$ \\
\midrule
\rowcolor{gray!15}
\multicolumn{5}{c}{\textit{Degradation}} \\
No Noise             & $-0.637$ & $-0.478$ & $+0.300$ & $+0.388$ \\
No Freq.              & $-0.650$ & $-0.449$ & $-0.600$ & $-0.444$ \\
No Blur              & $-0.494$ & $-0.340$ & $-0.517$ & $-0.333$ \\
No Mask              & $-0.485$ & $-0.326$ & $-0.733$ & $-0.556$ \\
\bottomrule
\end{tabular}
}
\caption{Ablation study measured by correlation with the RAISE benchmark and our human study.}
\label{tab:ablation}
\end{table}

\section{Conclusion}
We presented TRIG, a metric for evaluating generative models through texture realism. Instead of relying on representations trained for other vision tasks, TRIG learns a representation specifically for assessing the local texture. Compared to competing metrics, TRIG aligns more closely with human judgments of texture realism and can also produce spatial maps that localize inconsistent texture. As texture represents one aspect of image quality, we view TRIG as a complementary protocol to existing evaluation metrics.

\bibliography{aaai2027}

\appendix

\paragraph{Overview.}
This supplementary material provides additional details and results supporting the main paper. We first visualize the texture prior learned through GAN pretraining (Section~\ref{sec:supp_texture_prior}). We then describe the full training protocol, including the degradation operators, the region-based degradation procedure, optimization settings, and dataset splits (Section~\ref{sec:supp_training}). Then we discuss the implementation and computational properties of the three set-level distribution distances considered in the paper (Section~\ref{sec:supp_distances}). Section~\ref{sec:supp_ablation} reports additional
ablations supporting our choice of fine-tuned layers and training hyperparameters. We then provide complete details of the ImageNet human study, covering study design, the ethics statement, preference aggregation, reliability analysis, bootstrap confidence intervals, and image sourcing and randomization (Section~\ref{sec:supp_human_study}), and describe how we use the RAISE dataset and sample the COCO anchor set (Section~\ref{sec:supp-raise-grouping}). Finally, we probe TRIG further to check its sensitivity to texture quality perturbations (Section~\ref{sec:supp-semantic-texture}). We introduce more qualitative texture-realism maps responding to seen and unseen degradations across generators and image classes (Section~\ref{sec:supp_heatmaps}).

\section{What Priors Have Learned?}
\label{sec:supp_texture_prior}

\begin{figure}[t]
\centering
\setlength{\tabcolsep}{1pt}
\begin{tabular}{ccccc}

\includegraphics[width=0.185\linewidth]{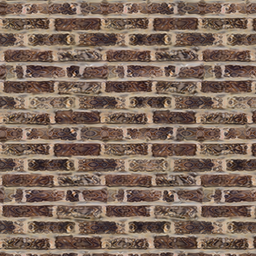} &
\includegraphics[width=0.185\linewidth]{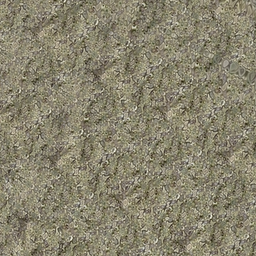} &
\includegraphics[width=0.185\linewidth]{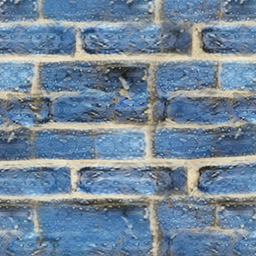} &
\includegraphics[width=0.185\linewidth]{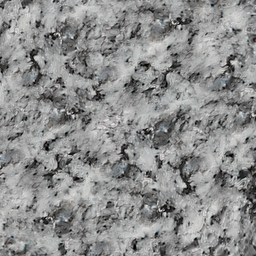} &
\includegraphics[width=0.185\linewidth]{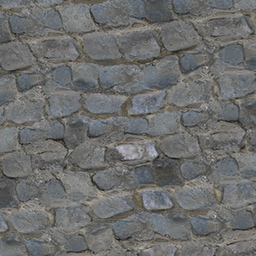}\\[-1pt]

\includegraphics[width=0.185\linewidth]{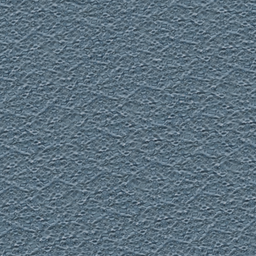} &
\includegraphics[width=0.185\linewidth]{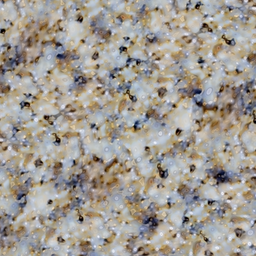} &
\includegraphics[width=0.185\linewidth]{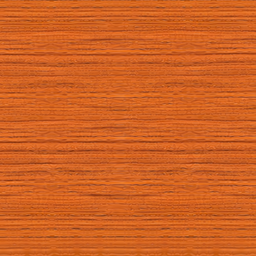} &
\includegraphics[width=0.185\linewidth]{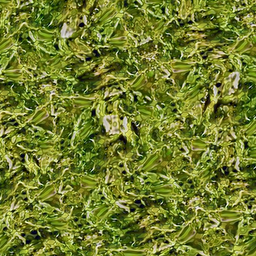} &
\includegraphics[width=0.185\linewidth]{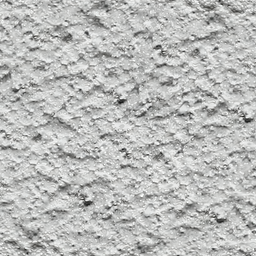} \\[-1pt]

\includegraphics[width=0.185\linewidth]{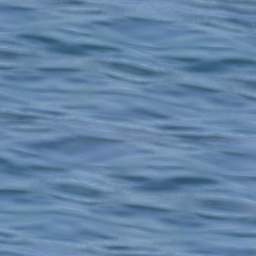} &
\includegraphics[width=0.185\linewidth]{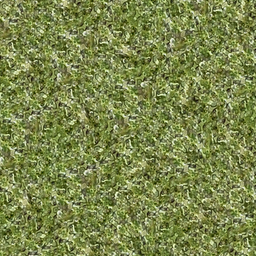} &
\includegraphics[width=0.185\linewidth]{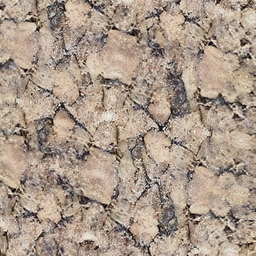} &
\includegraphics[width=0.185\linewidth]{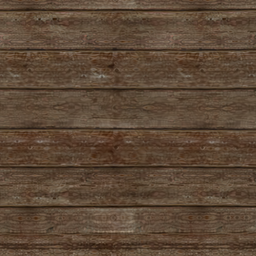} &
\includegraphics[width=0.185\linewidth]{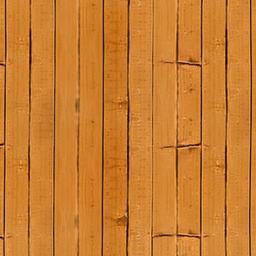} \\[-1pt]

\includegraphics[width=0.185\linewidth]{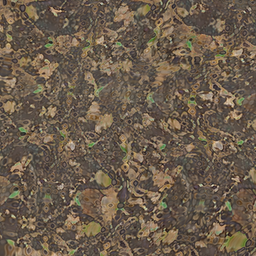}  &
\includegraphics[width=0.185\linewidth]{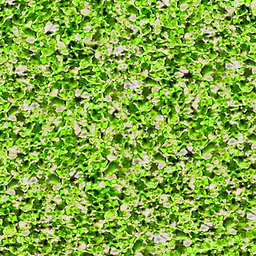} &
\includegraphics[width=0.185\linewidth]{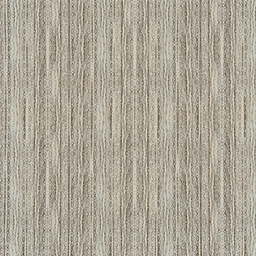} &
\includegraphics[width=0.185\linewidth]{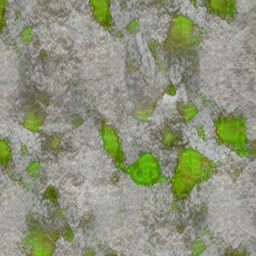} &
\includegraphics[width=0.185\linewidth]{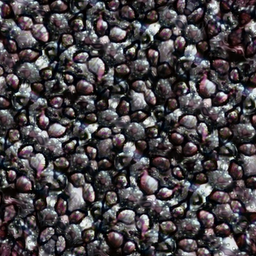} \\[-1pt]

\end{tabular}

\caption{Texture examples generated by the pretrained GAN generator pretrained on NUUR-Texture2953~\cite{lin2024dataset} dataset, paired with the discriminator we used. 
}
\label{fig:generated_textures}
\end{figure}

In Figure~\ref{fig:generated_textures} we present some examples generated by the generator in the StyleGAN3 we used as a source of texture prior, with the paper's ~\cite{ePrime2023} official implementations https://github.com/JueLin/stylegan3-texture-analysis-synthesis.git. Those textures cover several categories such as bricks, wall, water, grass, and wood with realistic details, illustrating the diversity of texture appearances encountered during adversarial pretraining and motivate using the associated discriminator as a texture-oriented initialization.

\section{Training Details}
\label{sec:supp_training}

\subsection{Degradation Pool Details}
\label{app:degradation}

In this section, we provide the exact parameterization of the degradation pool $\mathcal{P}$ used to train the plausibility head (summarized in Table~\ref{tab:degradation_pool}). The pool contains seven operators across three categories, with nine strength variants in total. All severities are normalized to $[0,1]$ for the ordering loss.

\paragraph{Category 1: Blur (optical and resolution loss).}
In this category, we apply uniform or directional low-pass operators modeling AI-generated defocus, motion, and resolution loss.
\begin{itemize}
\item \textbf{Gaussian blur}: convolution with an isotropic Gaussian kernel, applied at
two strength ranges, $\sigma\in[3,8]$ (light) and $\sigma\in[8,15]$ (heavy);
severity $=\sigma/15$.
\item \textbf{Box blur}: uniform $k\times k$ averaging kernel, $k\in\{7,11,15\}$;
severity $=k/15$.
\item \textbf{Motion blur}: $1\times L$ horizontal averaging kernel (linear motion),
$L\in[10,25]$; severity $=L/25$.
\item \textbf{Bicubic resample}: bicubic downsampling to resolution $R/s$ followed by
upsampling back to $R$, with $s\in[2,4]$ (i.e.\ $256\!\to\!\{128\dots64\}\!\to\!256$);
severity $=(s-1)/3$.
\end{itemize}

\paragraph{Natural defocus versus artificial over-smoothing.}
We stress that blur is not intrinsically unrealistic. Natural shallow depth of field follows lens optics and scene depth, producing defocus patterns and focus transitions
that remain consistent with scene geometry. Such patterns are retained in the undegraded real photographs used as positive samples. In contrast, our proxy degradation applies a single blur kernel uniformly within a selected region,
without access to depth, optical parameters, or motion, thereby introducing contextually inconsistent loss of detail. The multiscale feature pyramid can evaluate this local frequency change together with its surrounding structure,
rather than responding to low-frequency content alone.

\begin{figure}[t]
\centering
\setlength{\tabcolsep}{2pt}
\renewcommand{\arraystretch}{1.0}
    \begin{tabular}{@{}cccc@{}}
        \textbf{\small ImageNet}
        & \textbf{\small BG blur}
        & \textbf{\small VAR}
        & \textbf{\small ADM} \\[2pt]

        \includegraphics[width=0.23\linewidth]{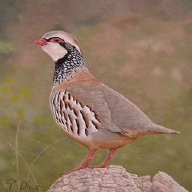}
        &
        \includegraphics[width=0.23\linewidth]{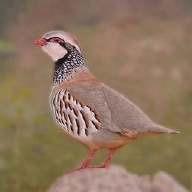}
        &
        \includegraphics[width=0.23\linewidth]{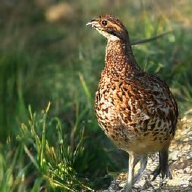}
        &
        \includegraphics[width=0.23\linewidth]{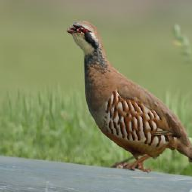}
        \\[3pt]

        \textbf{\scriptsize $S=0.967$}
        & \textbf{\scriptsize $S=0.462$}
        & \textbf{\scriptsize $S=0.881$}
        & \textbf{\scriptsize $S=0.717$}
        \\[2pt]

        \includegraphics[width=0.23\linewidth]{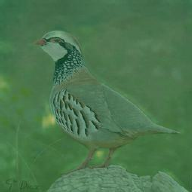}
        &
        \includegraphics[width=0.23\linewidth]{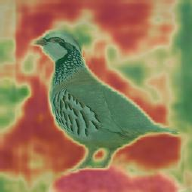}
        &
        \includegraphics[width=0.23\linewidth]{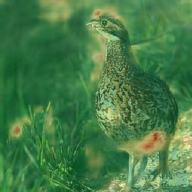}
        &
        \includegraphics[width=0.23\linewidth]{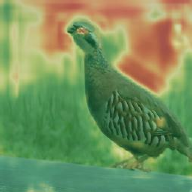}
    \end{tabular}

    \caption{\textbf{Natural defocus versus artificial over-smoothing.}
    The top row shows real and generated images containing background defocus, and the bottom row shows their corresponding texture-realism maps. Green indicates greater consistency with natural-image texture statistics, while red indicates lower texture consistency. The texture realism scores $S$ are reported above the heatmaps.}

    \label{fig:natural_vs_artificial_blur}
\end{figure}

Figure~\ref{fig:natural_vs_artificial_blur} illustrates this behavior. The original ImageNet photograph already contains natural background defocus and receives a uniformly high texture realism response ($S=0.967$). Applying additional uniform blur to its background reduces the score to $S=0.462$, with the lower responses concentrated primarily in the artificially over-smoothed region. A VAR image with coherent depth-of-field appearance similarly retains a relatively high score ($S=0.881$), whereas the ADM example, whose background exhibits less spatially consistent texture, produces a stronger localized response and a lower score ($S=0.717$).
Although qualitative, these examples indicate that our backbone does not treat blur itself as evidence of unrealistic texture; it responds more strongly when the loss of detail is inconsistent with the surrounding spatial context.

\paragraph{Category 2: Spectral / texture collapse.}
Frequency-domain operators reproduce the selective high-frequency degradation characteristic of latent/VAE-based generators. Unlike Category~1, these do not apply a
uniform low-pass.
\begin{itemize}
\item \textbf{Frequency suppression}: in the 2D FFT, mask a fraction of the mid/high-frequency band (radius $\rho >\tau_0$, i.e. $\Lambda(\rho)=\rho\,\mathbbold{1}[\rho\ge\tau_0]$ ) while preserving low-frequency structure, with strength $\alpha\in[0.5,0.9]$; severity $=\alpha/0.9$. This produces the selective high-frequency absence of VAE-bottleneck images rather than a uniform rolloff.
\item \textbf{High-frequency phase smoothing}: after the FFT, isolate the high-frequency region (radius $\rho>\tau_1$, $\tau_1$ randomly selected from $[0.15,0.35]$, radius is normalized by Nyquist frequency), smooth the phase of those components with a Gaussian ($\sigma\in[3,8]$) while leaving low-frequency phase and all magnitudes untouched, then invert. This mimics the reduced high-frequency phase randomness of generated images, whose magnitude spectrum may appear plausible while the phase is over-smooth; severity $=\sigma/8$.
\end{itemize}

\paragraph{Category 3: Noise.}
\begin{itemize}
\item \textbf{Signal-independent Gaussian noise}: additive noise $x+\mathcal{N}(0,\sigma)$ with $\sigma$ constant across intensities, at two strengths, $\sigma\in[0.02,0.05]$
(light) and $\sigma\in[0.06,0.10]$ (heavy), in $[0,1]$ photometric scale; severity $=\sigma/0.10$. Because its variance is independent of local intensity, it departs
from physically plausible (signal-dependent) sensor noise.
\end{itemize}

\begin{table}[t]
    \centering
    
    \setlength{\tabcolsep}{5pt}
    \setlength{\abovecaptionskip}{4pt}
    \setlength{\belowcaptionskip}{-2pt}
    \resizebox{0.95\columnwidth}{!}{
        \centering
        \begin{tabular}{@{}ll@{}}
        \toprule
        \textbf{Operator} & \textbf{Definition} \\
        \midrule
        \rowcolor{gray!15}
        \multicolumn{2}{@{}l}{\textit{Category 1: Blur (loss of local detail)}} \\
        Gaussian          & $I * G_\sigma$ \\
        Box               & $I * B_k$ ($k\times k$ uniform) \\
        Motion            & $I * M_L$ ($1\times L$ horizontal) \\
        Bicubic resample  & $\operatorname{Up}(\operatorname{Down}(I,s),s)$ ($s$ is the downscaling factor)\\
        \midrule
        \rowcolor{gray!15}
        \multicolumn{2}{@{}l}{\textit{Category 2: Spectral (frequency-domain alteration)}} \\
        Freq.\ suppress   & $(1-\alpha \Lambda)\hat{I}$ on $\rho>\tau_0$\\
        Phase smooth      & $\Phi \!\to\! \Phi * G_\sigma$ on $\rho>\tau_1$ \\
        \midrule
        \rowcolor{gray!15}
        \multicolumn{2}{@{}l}{\textit{Category 3: Noise (added high-frequency content)}} \\
        Gaussian noise    & $I + \epsilon,\ \epsilon\!\sim\!\mathcal{N}(0,\sigma^2\mathbf{I})$ i.i.d. per pixel \\
        \bottomrule
        \end{tabular}
    
    }
    \caption{Degradation pool $\mathcal{P}$. $G_\sigma$: Gaussian kernel; $\hat{I}=A\odot e^{j\Phi}$: 2D FFT with magnitude $A$, phase $\Phi$; $\rho$: frequency radius; $\tau_0,\tau_1$: frequency thresholds. $\Lambda$ the frequency mask, $\alpha$ its coefficient.}
    \label{tab:degradation_pool}
\end{table}

\subsection{Degradation Application Protocol}
Degradations are applied per region: each image is partitioned using SAM2. If SAM2 returns fewer than three valid regions, we instead use an N×N regular grid, with $N\in [4,8,16,32,64]$. For each image, we sample the clean-region probability $p_{clean}\!\sim\!\mathcal{U}(0.3,0.7)$, after which each region is independently left unchanged with probability $p_{clean}$. Otherwise, one of the seven degradation operators is sampled uniformly and applied at a randomly sampled strength. Unchanged regions receive label 1, and synthetically perturbed regions receive label 0. Severity is not used as a regression target, since strengths are not comparable across operator types; instead, the ranking loss (Eq.6 in main paper) enforces monotonicity \emph{within} an operator type, requiring that for two same-type regions with
$|\Delta\text{severity}|\ge\delta$ the more degraded region scores lower. The Gaussian-blur and noise operators each include two strength ranges to guarantee that same-type pairs exist for the ranking loss.

\begin{figure}[t]
\centering
\setlength{\tabcolsep}{2pt}
\renewcommand{\arraystretch}{1.0}
\begin{tabular}{@{}c c c c c@{}}
\textbf{\scriptsize Original} &
\textbf{\scriptsize SAM2} &
\textbf{\scriptsize Overlay} &
\textbf{\scriptsize Degraded} &
\\[2pt]
\includegraphics[width=0.23\linewidth]{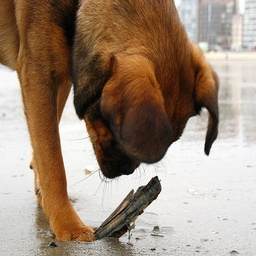} &
\includegraphics[width=0.23\linewidth]{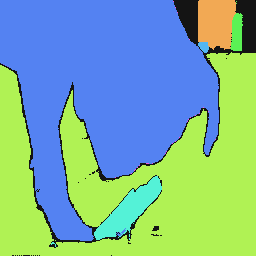} &
\includegraphics[width=0.23\linewidth]{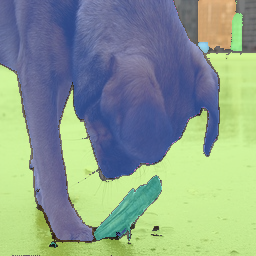} &
\includegraphics[width=0.23\linewidth]{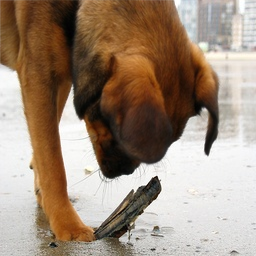} &
\raisebox{1.1\height}{\rotatebox[origin=c]{270}{\scriptsize Freq.\ suppress}}
\\[3pt]
\includegraphics[width=0.23\linewidth]{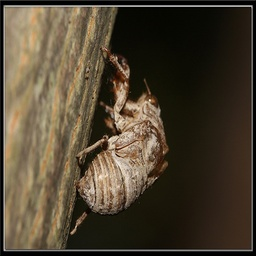} &
\includegraphics[width=0.23\linewidth]{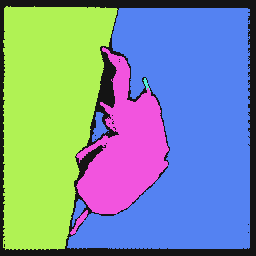} &
\includegraphics[width=0.23\linewidth]{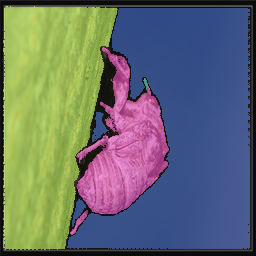} &
\includegraphics[width=0.23\linewidth]{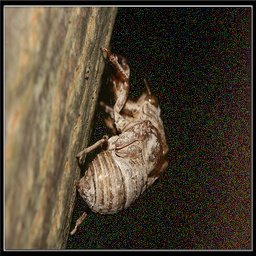} &
\raisebox{1.0\height}{\rotatebox[origin=c]{270}{\scriptsize Gaussian.\ noise}}
\\[3pt]
\includegraphics[width=0.23\linewidth]{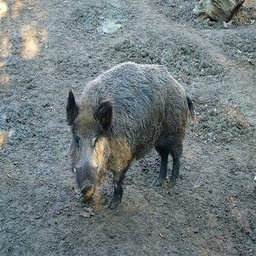} &
\includegraphics[width=0.23\linewidth]{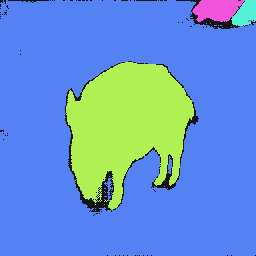} &
\includegraphics[width=0.23\linewidth]{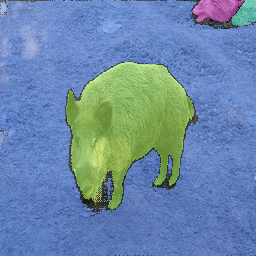} &
\includegraphics[width=0.23\linewidth]{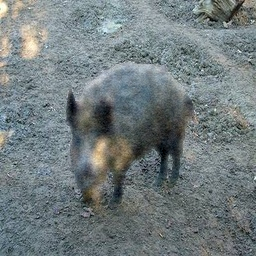} &
\raisebox{1.2\height}{\rotatebox[origin=c]{270}{\scriptsize Phase smooth}}
\\[3pt]
\includegraphics[width=0.23\linewidth]{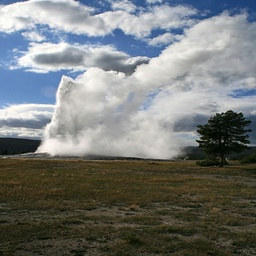} &
\includegraphics[width=0.23\linewidth]{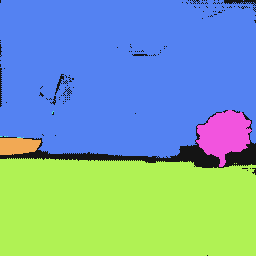} &
\includegraphics[width=0.23\linewidth]{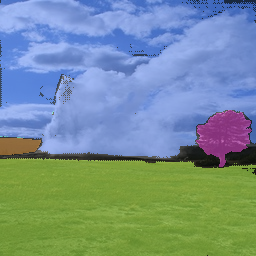} &
\includegraphics[width=0.23\linewidth]{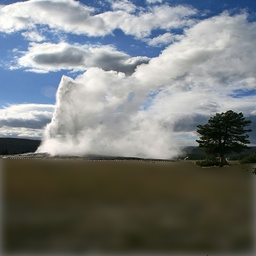} &
\raisebox{3.0\height}{\rotatebox[origin=c]{270}{\scriptsize Blur}}
\end{tabular}
\caption{Region-level degradation for fine-tuning. Each row shows the original image, its SAM2 region segmentation, the segmentation overlay, and the result after a degradation operator sampled from the pool is applied within the selected regions. One operator is shown per row: frequency suppression, Gaussian noise, phase smoothing, and blur.}
\label{fig:degradation_examples}
\end{figure}

\paragraph{Role of synthetic degradations.}
We emphasize that the degradation operators used for fine-tuning are proxy perturbations rather than an exhaustive definition of unrealistic texture. Blur and noise may occur naturally in real photographs, and their presence alone does not imply poor texture realism. Our operators instead expose the backbone to controlled attenuation, alteration, or amplification of local texture statistics resembling artifacts observed in generated images. Accordingly, the binary labels indicate whether a synthetic perturbation was applied, rather than whether the resulting region must be universally perceptually inconsistent. We show some examples of our training data by overlaying SAM2 mask on real image and apply various degradation to one selected region, as shown in Figure~\ref{fig:degradation_examples}. The transfer from these proxy perturbations to generated-image texture realism is evaluated independently through the two human studies reported in the main paper.

\subsection{Backbone Implementation and Training Configuration}
\label{sec:complete-config}

\paragraph{Architecture.}
The backbone comprises StyleGAN3 discriminator blocks
$\{\mathrm{b256,b128,b64,b32,b16}\}$. For a $256\times256$ input, the FPN receives features from $\mathrm{b16}$, $\mathrm{b32}$, $\mathrm{b64}$, and $\mathrm{b128}$ with shapes $512\times8\times8$, $512\times16\times16$, $512\times32\times32$, and $512\times64\times64$, respectively. Each is projected to 128 channels by
a $1\times1$ lateral convolution. Beginning at $8\times8$, each top-down stage bilinearly upsamples the coarser map by 2, adds the corresponding lateral map, and applies a $3\times3$, 128-channel convolution followed by LeakyReLU with slope 0.2. A $3\times3$ convolution from 128 to 64 channels, LeakyReLU,
and a $3\times3$ convolution from 64 to 1 channel produce the
$1\times64\times64$ logit map. Global average pooling of the final 128-channel FPN map gives the 128-dimensional image descriptor, while the single-image texture-consistency score is the spatial mean of the sigmoid output map.

\paragraph{Frozen and trainable components.}
Discriminator blocks $\mathrm{b256}$--$\mathrm{b16}$ are frozen, while the FPN and texture-realism head, containing 747{,}009 parameters, are trainable. Blocks $\mathrm{b8}$ and $\mathrm{b4}$ remain frozen and are not used in the FPN forward pass.

\paragraph{Preprocessing.}
Images are converted to RGB, resized directly to $256\times256$ without preserving aspect ratio using Lanczos interpolation, converted to floating-point tensors, and mapped channel-wise from $[0,1]$ to $[-1,1]$. We use no cropping, five-crop evaluation, flipping, or color augmentation. SAM2 region masks are precomputed, with regular grids used as a fallback. 

\paragraph{Optimization and objective.}
We use AdamW with learning rate $2\times10^{-4}$, weight decay $10^{-4}$, batch size 8, and global gradient-norm clipping at 1.0. Cosine annealing is applied over 50{,}000 steps and ends at $2\times10^{-6}$, or 1\% of the initial learning rate. The objective is
\[
\mathcal L
=\mathcal L_{\mathrm{BCE}}+\lambda\mathcal L_{\mathrm{ord}},
\qquad \lambda=0.5.
\]
Clean map cells are labeled 1 and degraded cells 0. For region pairs receiving the same degradation type and differing in normalized severity by at least $\delta=0.1$, we use
\[
\mathcal L_{\mathrm{ord}}
=\max(0,p_{\mathrm{severe}}-p_{\mathrm{mild}}+\eta),
\qquad \eta=0.05.
\]

\paragraph{Hyperparameter selection.}
The final settings $\lambda=0.5$, $\eta=0.05$, $\delta=0.1$, and 50{,}000 training steps were selected using the internal five-images-per-class ImageNet-training validation protocol. Selection considered validation BCE and ordering losses together with the held-out noise-SRCC monitor. Neither
the final human-study ranking nor the reported metric--MOS correlations in RAISE were used for hyperparameter selection. We confirm our hyperparameters selection with additional ablation study in Section~\ref{sec:supp_ablation}. Training configurations of our backbone are summarized in Table~\ref{tab:training-config}.

\begin{table}[t]
\centering
\small
\setlength{\tabcolsep}{4pt}
\renewcommand{\arraystretch}{1.08}
\begin{tabular}{@{}ll@{}}
\toprule
Component & Configuration \\
\midrule
Input & $256\times256$ RGB, $[-1,1]$ \\
Trainable parameters & 747{,}009 \\
Optimizer & AdamW \\
Learning rate & $2\times10^{-4}$ \\
Weight decay & $10^{-4}$ \\
Batch size & 8 \\
Training steps & 50{,}000 \\
LR schedule & Cosine to $2\times10^{-6}$ \\
Gradient clipping & Global norm 1.0 \\
$\lambda,\delta,\eta$ & $0.5,\ 0.1,\ 0.05$ \\
Hardware & One NVIDIA H100 (80\,GB) \\
Peak GPU memory & $\approx 12{,}000$ MiB \\
Training time & Approximately 10 h \\
\bottomrule
\end{tabular}
\caption{Our backbone training configuration.}
\label{tab:training-config}
\end{table}

\begin{table}[t]
\centering
\small
\setlength{\tabcolsep}{4pt}
\renewcommand{\arraystretch}{1.08}
\begin{tabular}{@{}ll@{}}
\toprule
Software & Version \\
\midrule
Python & 3.10.20 \\
PyTorch & 2.7.0+\texttt{cu128} \\
torchvision & 0.22.0+\texttt{cu128} \\
CUDA build & 12.8 \\
NumPy & 1.26.4 \\
SciPy & 1.15.3 \\
Pillow & 12.2.0 \\
pandas & 2.3.3 \\
scikit-learn & 1.7.2 \\
tqdm & 4.68.2 \\
SAM2 model & \texttt{sam2.1-hiera-small} \\
\bottomrule
\end{tabular}
\caption{Software configuration.}
\label{tab:software-config}
\end{table}

\paragraph{Computational resources and runtime.}
Training uses one NVIDIA H100 GPU, with peak GPU memory around
12{,}000\,MiB. The precise H100 form factor, CPU model, system RAM, operating-system build, and driver version were not retained. Training the main model for 50{,}000 steps takes approximately 10 hours. SAM2 masks are precomputed and their cost is therefore excluded from training time; we
refer to the SAM2 paper~\cite{ravi2025sam} for the computational requirements of mask generation. FPN feature extraction runs at approximately 213.09 images/s, so processing a 50{,}000-image source requires about 234 seconds, excluding initialization, data loading, and disk-cache overhead. Image scoring uses the same forward pass and has approximately the same runtime. Software dependencies are summarized in Table~\ref{tab:software-config}.

\paragraph{Dataset splits and leakage control.}
The ImageNet-anchor set we used to evaluate the distances from each source is sampled from official ImageNet validation split, 50 per class, which is already separate from ImageNet training and have never been seen in our backbone training. ImageNet-eval acts as a reference to calibrate the distance between ImageNet-anchor and all sources (ImageNet-eval should be closer to anchor than other sources). The ImageNet-eval is the first 50 samples encountered per class when iterating the Hugging Face ImageNet training split without shuffling, which is held out from our training set. 

Table~\ref{tab:image-splits} specifies the composition of
every image set. Our backbone is trained exclusively on ImageNet training images and does not use ImageNet-anchor, ImageNet-eval, or any generated images used in evaluation. Within each lexicographically sorted class directory, the final five images are reserved for validation and all remaining files are available for training, yielding 5{,}000 validation and 1{,}105{,}841 training images. This internal validation set is distinct from the official 50{,}000-image ImageNet-anchor and ImageNet-eval.

\begin{table}[t]
\centering
\small
\setlength{\tabcolsep}{4pt}
\begin{tabular}{p{0.18\linewidth}p{0.54\linewidth}r}
\toprule
Set & Exact definition & Size \\
\midrule
Backbone train set &
Images from ImageNet official train set, excluding the final five
lexicographically sorted filenames in each class and ImageNet-eval &
1{,}105{,}841 \\
Backbone val set &
Final five lexicographically sorted training images per class &
5{,}000 \\
ImageNet-anchor &
A subset of images from ImageNet official val set, with 50 images per class &
50{,}000 \\
ImageNet-eval & 
A subset of images from ImageNet official train set, with 50 images per class & 50{,}000\\
Evaluated sources &
12 generated-image directories, each with 50 images per class &
50{,}000/source \\
NUUR-Texture2953 &
Separate real-texture dataset absent from all evaluation, anchor, and human-study manifests &
2{,}953 \\
\bottomrule
\end{tabular}
\caption{Image splits and leakage controls.}
\label{tab:image-splits}
\end{table}

Our backbone is initialized from publicly released GAN weights
pretrained on NUUR-Texture2953~\cite{lin2024dataset}. This dataset contains only real homogeneous texture images and is disjoint from all ImageNet and generated-image manifests used in our experiments. Neither the backbone pretraining nor our backbone fine-tuning uses outputs from any evaluated generator. Thus, TRIG is not trained on generator-specific artifacts from the models it evaluates.

\section{Set-Level Distances: FD, MMD, and SWD}
\label{sec:supp_distances}

\begin{table*}[t]
\centering
\small
\setlength{\tabcolsep}{6pt}
\renewcommand{\arraystretch}{1.25}
\newcommand{\yes}{\textcolor{green!55!black}{\ding{51}}}
\newcommand{\no}{\textcolor{red}{\ding{55}}}
\begin{tabular}{llll}
\toprule
 & \textbf{Fr\'echet distance} & \textbf{MMD distance} & \textbf{SWD (ours)} \\
\midrule
\multirow{3}{*}{\shortstack[l]{Inception\\embeddings}}
 & \no~Weak texture cues     & \no~Weak texture cues     & \no~Weak texture cues \\
 & \no~Normality assumption  & \yes~Distribution-free    & \yes~Distribution-free \\
 & \yes~Scales, $O(n)$       & \no~$O(n^2)$, subset-limited & \yes~Scales, $O(n\log n)$ \\
\midrule
\multirow{3}{*}{\shortstack[l]{CLIP / DINOv2\\embeddings}}
 & \yes~Semantic embeddings  & \yes~Semantic embeddings  & \yes~Semantic embeddings \\
 & \no~Normality assumption  & \yes~Distribution-free    & \yes~Distribution-free \\
 & \yes~Scales, $O(n)$       & \no~$O(n^2)$, subset-limited & \yes~Scales, $O(n\log n)$ \\
\midrule
\multirow{3}{*}{\shortstack[l]{Texture backbone\\(ours)}}
 & \yes~Texture-aware embeddings & \yes~Texture-aware embeddings & \yes~Texture-aware embeddings \\
 & \no~Normality assumption      & \yes~Distribution-free        & \yes~Distribution-free \\
 & \yes~Scales, $O(n)$           & \no~$O(n^2)$, subset-limited  & \yes~Scales, $O(n\log n)$ \\
\bottomrule
\end{tabular}
\caption{Design space of set-level generative-evaluation metrics along two axes: the backbone (rows) and the distributional distance (columns). Along the distance axis, the
Fr\'echet distance scales cheaply but assumes Gaussian features; MMD is distribution-free but costs $O(n^2)$ and is limited to subsets at large $n$; SWD is both distribution-free
and $O(n\log n)$, scaling to the sample sizes at which FID is reported.}
\label{tab:design_space}
\end{table*}

Given feature sets $\{x_i\}_{i=1}^{m}$ and $\{y_j\}_{j=1}^{m}$ with empirical distributions
$\mu$ and $\nu$ in $\mathbb{R}^d$, a set-level metric reduces them to a scalar distance.
The three distances we consider differ along two independent axes summarized in
Table~\ref{tab:design_space}: the \emph{backbone} that produces the features, and the
\emph{distance} that compares their distributions. Here we detail the three distances.

\paragraph{Fr\'echet distance (FD).}
FD models each set as a multivariate Gaussian, $\mu\!\approx\!\mathcal{N}(m_x,\Sigma_x)$
and $\nu\!\approx\!\mathcal{N}(m_y,\Sigma_y)$, and compares only their first two moments:
\begin{equation}
\begin{aligned}
\operatorname{FD}^2(\mu,\nu)
= &
\lVert \mathbf m_x-\mathbf m_y\rVert_2^2
\\+&
\operatorname{Tr}\!\left(
\mathbf\Sigma_x+\mathbf\Sigma_y
-
2\left(
\mathbf\Sigma_x^{1/2}
\mathbf\Sigma_y
\mathbf\Sigma_x^{1/2}
\right)^{1/2}
\right)    
\end{aligned}
\end{equation}
Estimating $(m,\Sigma)$ costs $O(nd^2)$ and the closed form is independent of $n$, so FD
scales cheaply to large sample sizes. However, it is exact only when the features are
Gaussian; any higher-order structure is ignored, and its finite-sample estimator is biased
in a model-dependent way. FID applies FD to Inception features.

\paragraph{Maximum mean discrepancy (MMD).}
MMD is non-parametric: it embeds each distribution into a reproducing-kernel Hilbert space
via a kernel $k$ and compares the mean embeddings,
\begin{equation}
\begin{aligned}
\mathrm{MMD}^2(\mu,\nu) = {}& \mathbb{E}_{x,x'}[k(x,x')]+\mathbb{E}_{y,y'}[k(y,y')] \\
&-2\,\mathbb{E}_{x,y}[k(x,y)] .
\end{aligned}
\end{equation}
With a characteristic kernel, $\mathrm{MMD}(\mu,\nu)=0$ iff $\mu=\nu$, so MMD is sensitive
to differences of all orders and makes no Gaussian assumption. Its unbiased estimator sums
over all pairs, costing $O(n^2 d)$, which becomes expensive at the large sample sizes at
which FID is reported; it also requires choosing a kernel and bandwidth. KID, CMMD, and
Gram-MMD are MMD variants under different kernels and backbones.

\paragraph{Sliced Wasserstein distance (SWD).}
SWD is also non-parametric but avoids both the Gaussian assumption of FD and the kernel
choice of MMD. It compares two distributions through their one-dimensional projections onto
random directions $\mathbf{v}\in\mathbb{S}^{d-1}$:
\begin{equation}
\mathrm{SWD}_2^2(\mu,\nu)=\int_{\mathbb{S}^{d-1}}
W_2^2\!\big(P^{\mathbf{v}}_{\#}\mu,\,P^{\mathbf{v}}_{\#}\nu\big)\,d\mathbf{v},
 P^{\mathbf{v}}(x)=\mathbf{v}^{\top}x ,
\end{equation}
which we approximate with $K$ sampled directions,
\begin{equation}
\widehat{\mathrm{SWD}}_2^2(\mu,\nu)=\frac{1}{K}\sum_{k=1}^{K}
W_2^2\!\big(P^{\mathbf{v}_k}_{\#}\mu,\,P^{\mathbf{v}_k}_{\#}\nu\big).
\end{equation}
Each one-dimensional $W_2$ has a closed form obtained by sorting the two sets of projected values and matching them in order, giving an $O(n\log n)$ cost per projection. The population SWD is a proper metric and is sensitive to the full distribution; we approximate it using $K$ random projections. It requires no kernel and scales to the large
sample sizes ($n\sim5\times10^4$) at which FID is reported. This is the distance used by TRIG (Eq.~10 in the main paper), applied to the image descriptors $\mathbf{g}(I)$, with $K$ controlling the variance of the estimate. We therefore adopt SWD as our default set-level distance and report FD and MMD as ablations (Table~4 in the main paper), where the ranking is robust to the choice of distance.

\paragraph{Implementation Details of Distances.} We compute TRIG (SWD) by drawing \(K=1000\) directions by normalizing independent standard-Gaussian vectors to unit length. When the two sets have unequal sizes, their sorted projected samples are linearly interpolated as empirical quantile functions onto the common endpoint-inclusive grid \(q_\ell\in[0,1]\), with \(L=\min(n,m)\). In our experiments, we standardize every feature dimension using the mean and standard deviation of the anchor set only first. When we apply MMD in this paper, we use the Gaussian RBF kernel. Features are standardized using anchor-only statistics as above. The bandwidth is fixed once
from the anchor:
\[
 \gamma_{med}=\left(2\,\operatorname{median}_{i<j}
 \lVert\widetilde{\mathbf{x}}_i-\widetilde{\mathbf{x}}_j\rVert_2^2\right)^{-1}.
\]
Thus the bandwidth is shared across all model comparisons and does not depend on an evaluation set. We apply FD to our texture backbone like FID, which means directly using the raw (not anchor-standardized or \(L_2\)-normalized) image-level FPN features. KID and CMMD follow the configuration of the published version.

\paragraph{Computational considerations.}
As Table~\ref{tab:design_space} indicates, FD is efficient but assumes Gaussian features; MMD removes that assumption but scales as $O(n^2)$ in the number of samples; SWD is both
distribution-free and kernel-free while scaling more smoothly than the quadratic MMD estimator. Cost also depends on the feature dimension. The Gram-based metrics operate on very high-dimensional descriptors---$295{,}296$-d for Gram-MMD with DINOv2 and $524{,}800$-d with DC-AE, while CMMD uses $768$-d CLIP embeddings and FID/KID use $2{,}048$-d Inception features. TRIG instead compares $128$-d descriptors
$\mathbf{g}(I)$, one to three orders of magnitude smaller, so both the projection step of SWD and the pairwise computations of the baseline distances are far cheaper on our representation. The three distances differ only in how they compare distributions, orthogonal to the backbone axis; pairing any of them with a texture-aware backbone (bottom
row) therefore isolates the effect of the representation from that of the distance, while our compact descriptor keeps the overall metric inexpensive to evaluate at the sample sizes (50k) used by FID to provide a fair comparison.

This scaling limit is not merely theoretical: the Gram-based estimators exhausted GPU memory on the full $50$k sample size, so we computed Gram-MMD over subsets ($10$k for the
DINOv2 variant and $5$k for the higher-dimensional DC-AE variant) and repeated each evaluation over six random seeds ($42$--$47$) to report the mean and standard deviation in
Table 1 of the main paper. SWD, by contrast, runs on the full sample at $O(n\log n)$, with randomness from selection of $K$ directions.

\section{Additional Ablation Study of TRIG}
\label{sec:supp_ablation}

Table 4 in the main paper reports a representative subset of ablations. Here we provide other additional variants evaluated on both studies, reporting the correlation with
RAISE MOS and with our ImageNet human ranking (Table~\ref{tab:app_full_abl}), grouped by the component each variant probes. In the main paper, we choose to only fine-tune FPN layer with $\lambda= 0.5$, $\delta=0.1$, and $\eta=0.05$ for 50{,}000 steps. In this table we show several variants that remain moderately negative on RAISE \emph{reverse sign} on the ImageNet study, indicating some failures the external benchmark RAISE alone would not reveal. 

\paragraph{Fine-tuned layers.}
Fine-tuning the full FPN head is best. We have variants restricting training to a single backbone block
(\texttt{b16 only}, \texttt{b32 only}) sharply reduces RAISE agreement ($-0.339$, $-0.138$), confirming that multi-scale fusion, not any single block, carries the signal. Extending
fine-tuning into the backbone (\texttt{FPN+b16}, \texttt{FPN+b16+b32}) does not improve over the FPN-only default, so we keep the backbone frozen to preserve the texture prior rather than overwriting it.

\paragraph{Ordering loss and training length.}
The metric is robust to the ordering-loss hyperparameters. Varying the loss weight $\lambda$, margin $\eta$, and severity threshold $\delta$, or removing the ordering loss entirely,
changes the correlation only mildly and never reverses sign, suggesting the ordering loss refines but is not essential. Agreement improves with training and saturates by $\sim$50k steps, which we adopt as the default.

\begin{table}[t]
\centering\small
\setlength{\tabcolsep}{4pt}
\renewcommand{\arraystretch}{1.1}
\begin{tabular}{lcccc}
\toprule
\multirow{2}{*}{Variant} & \multicolumn{2}{c}{RAISE} & \multicolumn{2}{c}{ImageNet} \\
 & Spear. & Kend. & Spear. & Kend. \\
\midrule
TRIG (main)          & $\mathbf{-0.747}$ & $\mathbf{-0.594}$ & $\mathbf{-0.750}$ & $\mathbf{-0.611}$ \\
\midrule
\multicolumn{5}{@{}l}{\textit{Fine-tuned layers}} \\
FPN+b16              & $-0.730$ & $-0.551$ & $-0.667$ & $-0.556$ \\
FPN+b16+b32          & $-0.650$ & $-0.500$ & $-0.650$ & $-0.500$ \\
b16 only             & $-0.339$ & $-0.217$ & $-0.733$ & $-0.556$ \\
b32 only             & $-0.138$ & $-0.094$ & $-0.717$ & $-0.500$ \\
\midrule
\multicolumn{5}{@{}l}{\textit{Ordering loss weight $\lambda$, main $\lambda{=}0.5$}} \\
$\lambda{=}0.2$      & $-0.717$ & $-0.536$ & $\mathbf{-0.750}$ & $\mathbf{-0.611}$ \\
$\lambda{=}0.8$      & $-0.665$ & $-0.522$ & $-0.600$ & $-0.444$ \\
$\lambda{=}0$        & $-0.659$ & $-0.500$ & $-0.733$ & $-0.556$ \\
\midrule
\multicolumn{5}{@{}l}{\textit{Ordering margin $\eta$, main $\eta{=}0.05$}} \\
$\eta{=}0.1$         & $-0.746$ & $-0.565$ & $-0.650$ & $-0.500$ \\
$\eta{=}0.02$        & $-0.717$ & $-0.551$ & $-0.733$ & $-0.556$ \\
$\eta{=}0$           & $-0.710$ & $-0.536$ & $-0.617$ & $-0.500$ \\
\midrule
\multicolumn{5}{@{}l}{\textit{Threshold $\delta$, main $\delta{=}0.1$}} \\
$\delta{=}0.2$       & $\mathbf{-0.747}$ & $-0.565$ & $-0.667$ & $-0.556$ \\
$\delta{=}0.05$      & $-0.698$ & $-0.536$ & $-0.667$ & $-0.556$ \\
$\delta{=}0$         & $-0.690$ & $-0.500$ & $-0.650$ & $-0.500$ \\
\midrule
\multicolumn{5}{@{}l}{\textit{Training steps, main 50k}} \\
20k                  & $-0.600$ & $-0.486$ & $-0.600$ & $-0.444$ \\
30k                  & $-0.645$ & $-0.500$ & $-0.733$ & $-0.556$ \\
40k                  & $-0.734$ & $\mathbf{-0.594}$ & $\mathbf{-0.750}$ & $\mathbf{-0.611}$ \\
60k                  & $-0.739$ & $-0.551$ & $\mathbf{-0.750}$ & $\mathbf{-0.611}$ \\
80k                  & $-0.716$ & $-0.515$ & $\mathbf{-0.750}$ & $\mathbf{-0.611}$ \\
\bottomrule
\end{tabular}
\caption{Additional ablation across both studies. Values are Spearman/Kendall correlations with human ratings (lower/more negative $=$ better agreement).}
\label{tab:app_full_abl}
\end{table}

\section{ImageNet Study Details}
\label{sec:supp_human_study}

\subsection{Perceptual Judgments: 2AFC}

Unlike some classification tasks which can be supervised by clear ground truth, the perceived realism depends more on subjective preference. To fill this gap, there is a long-standing paradigm from psychophysics for quantifying perception. The \emph{two-alternative forced choice} (2AFC) protocol, dating back to Fechner's foundational work in 1860s~\cite{fechner1860}, presents an observer with two stimuli and asks which exhibits a given perceptual attribute more strongly. Because a relative pairwise judgment avoids the calibration and scale-drift problems of absolute rating scales, 2AFC has become a workhorse for subjective image-and video-quality assessment, valued for its high discrimination power and reliability ~\cite{ding2021comparison, prashnani2018pieapp}. 

Another consideration in forced-choice studies is whether to permit an ``undecided''/tie response: allowing ties yields more faithful data when stimuli are genuinely indistinguishable, at the cost of discarding those trials from the decisive signal~\citep{men2023relaxed}. We therefore adopted a 2AFC-style pairwise comparison augmented with a tie option to obtain a ground-truth ordering of image sources by perceived \emph{texture realism}, and the procedure for correlating each distribution metric against it.

\subsection{Study Design}
Participants performed a blind, anonymous pairwise comparison of texture realism. Table~\ref{tab:human_setup} summarizes the protocol. Each trial draws one ImageNet class, two distinct sources, and one random image from each source within that class; the two images are randomly assigned to the left/right panes. Every session uses a fresh random seed, so no two participants see the same trial sequence. All relevant images and human judgments \textbf{will be released} together with our source code.

\begin{table}[t]
\centering\small
\begin{tabular}{@{}ll@{}}
\toprule
Property & Value \\
\midrule
Participants (sessions)     & 24 \\
Trials per participant      & 200 \\
Total comparisons           & 4{,}800 \\
Image sources               & 10 (2 real + 8 generators) \\
Responses per trial         & Left / Tie / Right \\
Session duration            & $\approx$10-20 min \\
\bottomrule
\end{tabular}
\caption{Human study protocol. Sources are two disjoint real ImageNet sets and eight generators (ADM, BigGAN, DiT, PMF, StyleGAN-XL, SiT-ODE, SiT-SDE, SoFlow). The two real
sets serve as a sanity check: because both are genuinely real, participants should find them indistinguishable, so a strong preference between them would indicate unreliable or
biased judging.}
\label{tab:human_setup}
\end{table}

\paragraph{Task framing.} Participants judge which image has more realistic \emph{surface texture}, such as fur, skin, foliage, fabric, grain, rather than which image is sharper. When both sides appear equally real or equally fake, participants select \emph{Tie}. Sources are hidden and the interface is fully blind.

\paragraph{Calibration.} Before the scored trials, each participant completes a guided walkthrough with the identical interface: they decide each example, after which the intended answer and a short explanation are revealed (covering ``judge texture, not content,'' and some tie cases).

\begin{figure}[t]
\centering
\includegraphics[width=\linewidth]{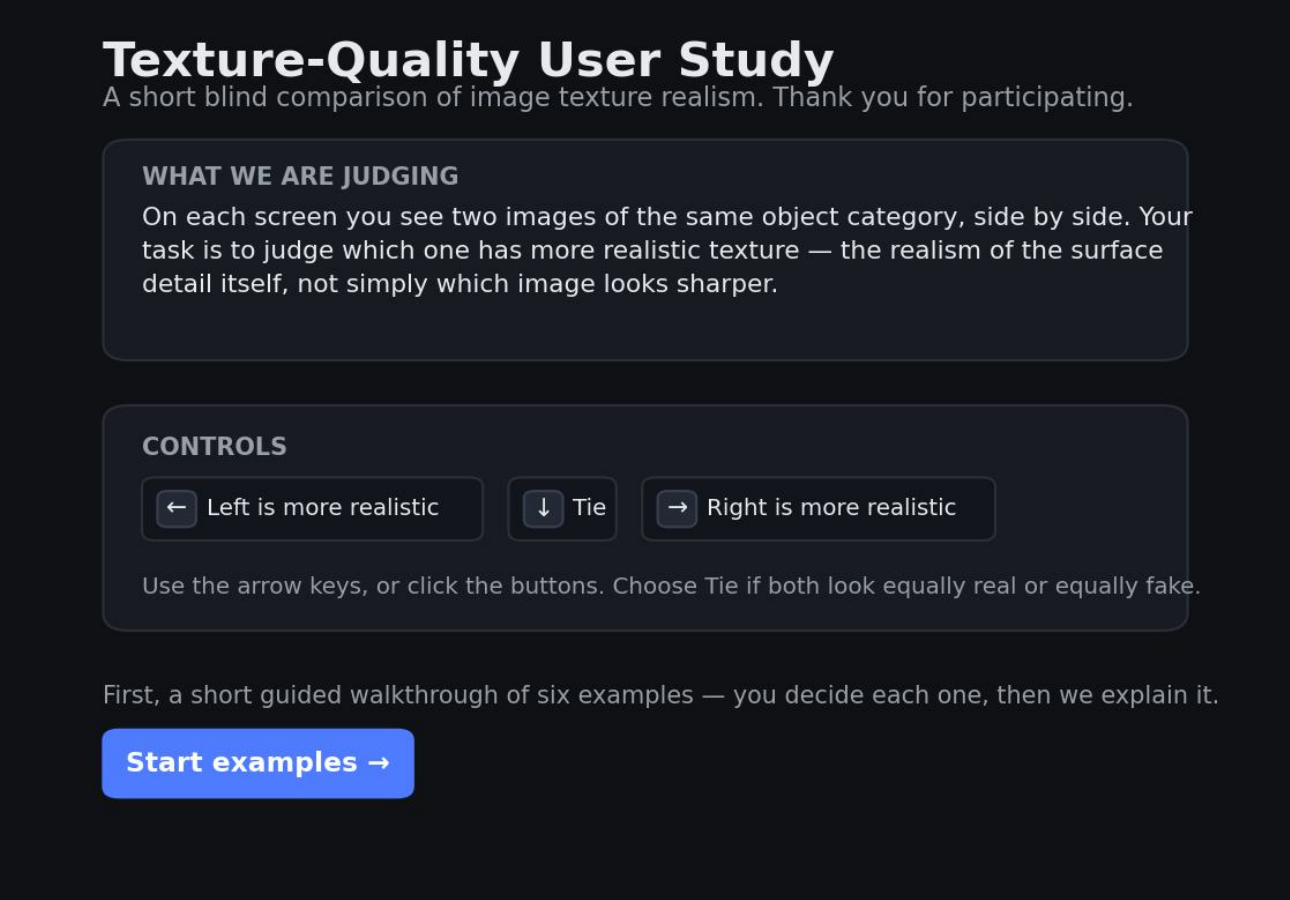}
\includegraphics[width=\linewidth]{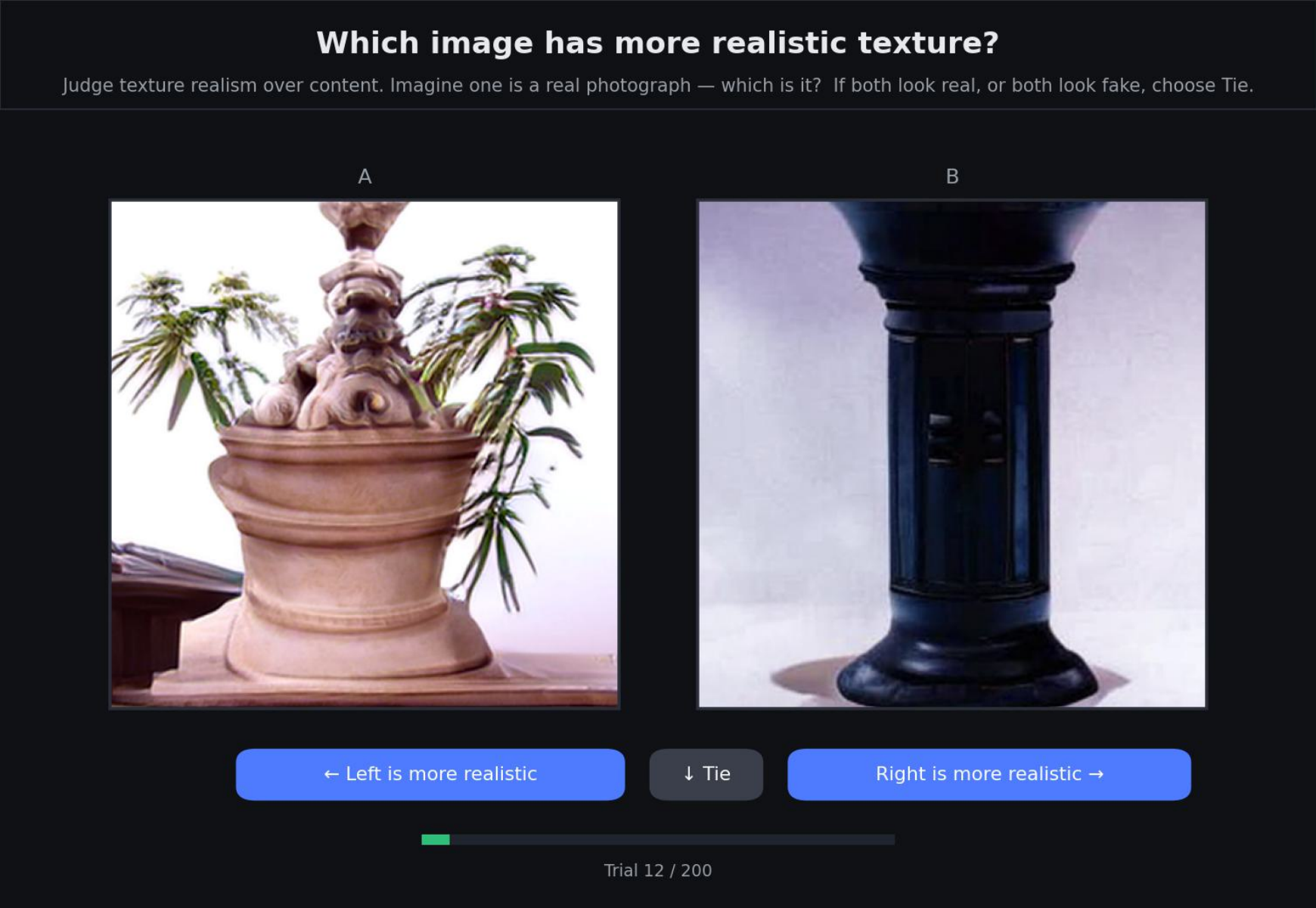}
\caption{\textbf{Study interface.} \emph{Top:} instruction screen defining the task and the Left/Tie/Right controls. \emph{Bottom:} a trial screen showing two same-class images with the three response buttons and a progress counter.}
\label{fig:human_ui}
\end{figure}


\paragraph{Ethics statement.} This work includes a human perceptual study to evaluate the realism of generated images. Participants were informed about the purpose of the study and voluntarily provided their judgments. No personally identifiable information or sensitive personal data were collected, and responses were analyzed only in aggregate. Participants were free to stop the study at any time. All participants have normal or corrected to normal vision. No payments are provided. The images used in the study were generated by publicly available generative models or obtained from publicly available datasets, and the study involved only perceptual comparisons without exposing participants to sensitive or harmful content. We believe the risks associated with this study are minimal.

\subsection{Aggregating Pairwise Judgments}
We collected 24 completed sessions, and no participant was excluded. We convert the 4{,}800 pairwise judgments into a per-source realness score in two ways, which agree strongly (Spearman $+0.983$ over the eight generators). Correlations are computed over n=9 non-anchor sources: eight generators and one real set. We only use one of real sets because the two real sets are theoretically identical, and they only serve as a sanity check.

\paragraph{Win-rate (primary).} For each source,
$\text{win-rate}=(\text{wins}+0.5\cdot\text{ties})/\text{games}$, counted over all opponents.
Higher values indicate greater perceived realism.

\begin{figure}[t]
\centering
\includegraphics[width=\linewidth]{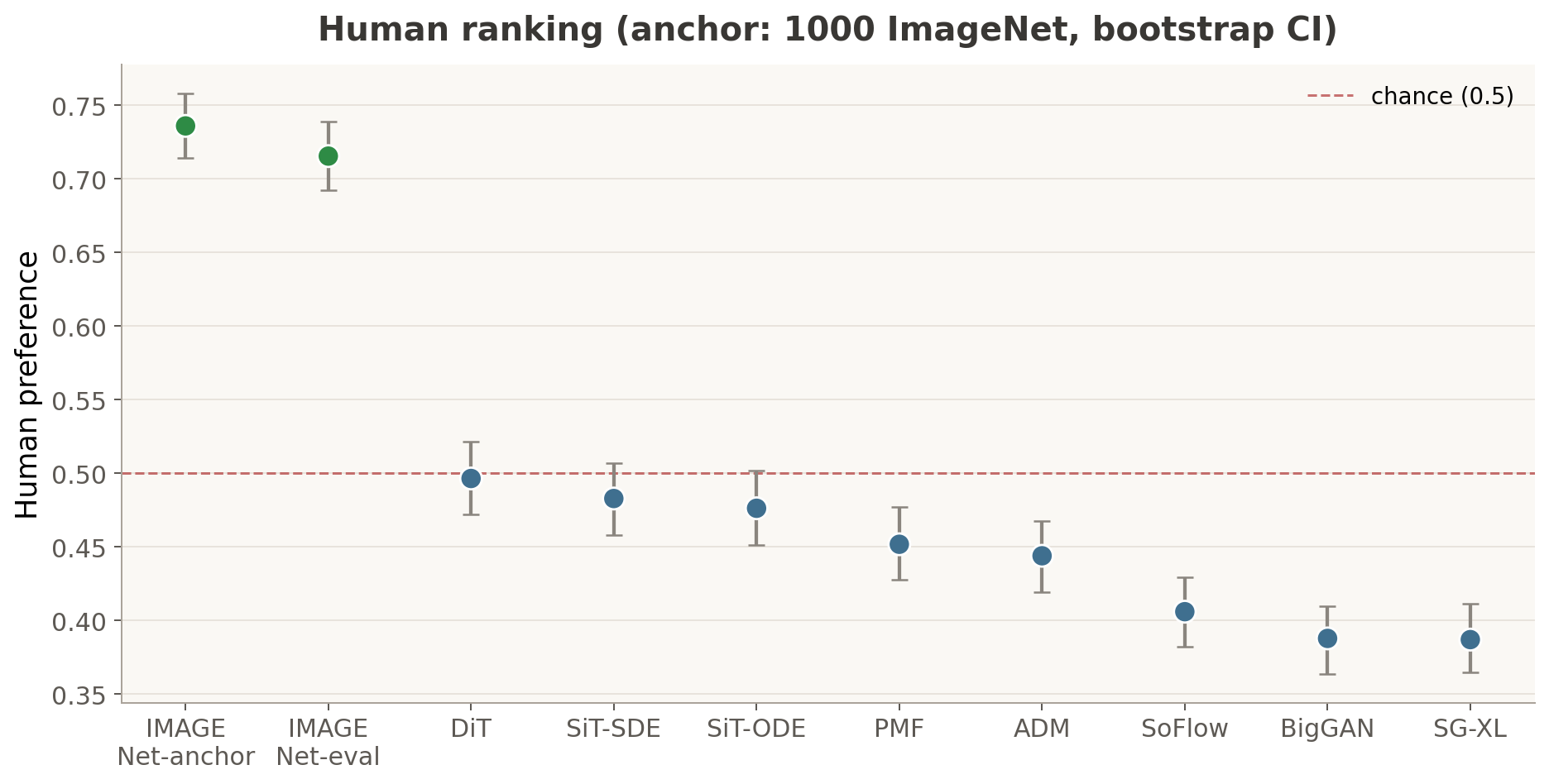}
\caption{\textbf{Human preference ranking.} Per-source win-rate for all ten sources, sorted, with $95\%$ bootstrap confidence intervals. The two real sets (green) sit well above the chance line ($0.5$); the eight generators cluster at or below chance with overlapping intervals, showing that generator-vs-generator preference is weaker while real-vs-generated is decisive.}
\label{fig:human_ranking}
\end{figure}

\paragraph{Bradley--Terry.} We also fit a Bradley--Terry~\cite{hunter2004bt} model by the Hunter MM maximum likelihood estimator, yielding mean-centered log-strengths $\beta$ (higher $=$ more real).

Both scores place the two real sets clearly at the top and the generators near or below the $0.5$ chance line, consistent with humans reliably distinguishing real from generated while struggling to separate one generator from another. We report the resulting human ranking with trial-level bootstrap confidence intervals in Figure~\ref{fig:human_ranking}.

\paragraph{Split-half reliability.}
We assessed the reliability of the source ranking using 2,000 random participant-level splits with a fixed seed. In each repetition, the 24  participants were divided into two disjoint groups of 12, source win rates were recomputed independently in each half, and the resulting rankings were compared using Spearman's rank correlation. For the nine sources included in the primary ranking analysis, the median split-half correlation was $\rho = 0.917$, with a central 95\% range of $[0.767,\,0.983]$ across random splits. Applying the Spearman--Brown correction yielded an estimated full-study reliability of $0.957$, with a central 95\% range of $[0.868,\,0.992]$. Reliability remained high when considering all ten sources ($0.969$ corrected) or only the eight generated-image sources ($0.937$ corrected), indicating that the source ordering is stable across independent subsets of participants.
  
\subsection{Generator-vs-Generator Preferences}
\begin{figure}[t]
\centering
\setlength{\tabcolsep}{2pt}
\renewcommand{\arraystretch}{1.0}
\begin{tabular}{cccc}

\includegraphics[width=0.20\linewidth]{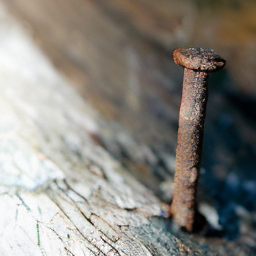} &
\includegraphics[width=0.20\linewidth]{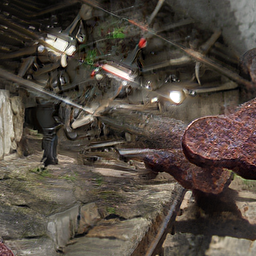} &
\includegraphics[width=0.20\linewidth]{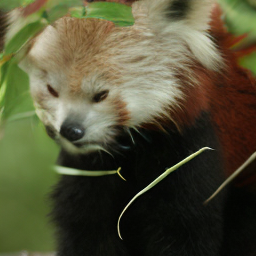} &
\includegraphics[width=0.20\linewidth]{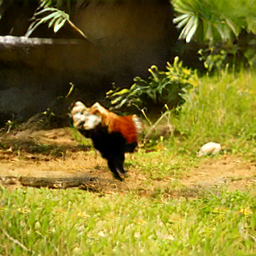} \\
{\scriptsize \textcolor{green!55!black}{\textbf{DiT}\,\ding{51}}} & {\scriptsize SG-XL} &
{\scriptsize \textcolor{green!55!black}{\textbf{DiT}\,\ding{51}}} & {\scriptsize SG-XL} \\[6pt]

\includegraphics[width=0.20\linewidth]{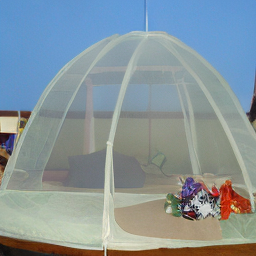} &
\includegraphics[width=0.20\linewidth]{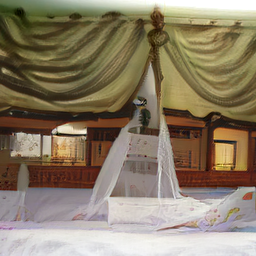} &
\includegraphics[width=0.20\linewidth]{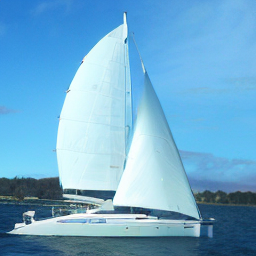} &
\includegraphics[width=0.20\linewidth]{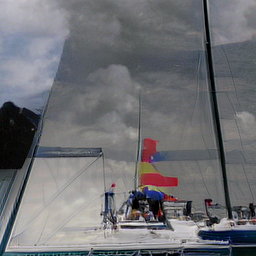} \\
{\scriptsize \textcolor{green!55!black}{\textbf{DiT}\,\ding{51}}} & {\scriptsize SG-XL} &
{\scriptsize \textcolor{green!55!black}{\textbf{DiT}\,\ding{51}}} & {\scriptsize SG-XL} \\[6pt]

\includegraphics[width=0.20\linewidth]{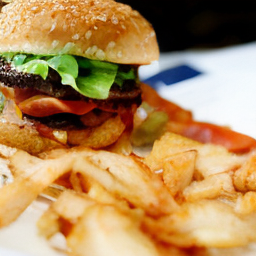} &
\includegraphics[width=0.20\linewidth]{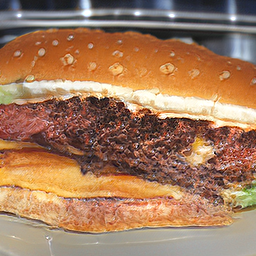} &
\includegraphics[width=0.20\linewidth]{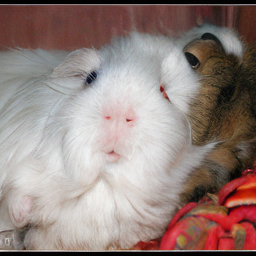} &
\includegraphics[width=0.20\linewidth]{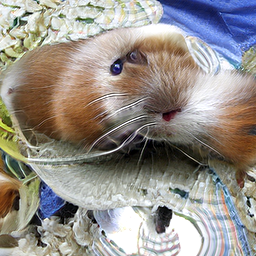} \\
{\scriptsize \textcolor{green!55!black}{\textbf{DiT}\,\ding{51}}} & {\scriptsize SG-XL} &
{\scriptsize \textcolor{green!55!black}{\textbf{DiT}\,\ding{51}}} & {\scriptsize SG-XL} \\[6pt]

\includegraphics[width=0.20\linewidth]{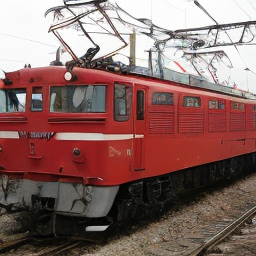} &
\includegraphics[width=0.20\linewidth]{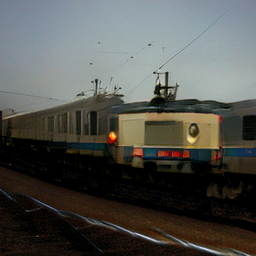} &
\includegraphics[width=0.20\linewidth]{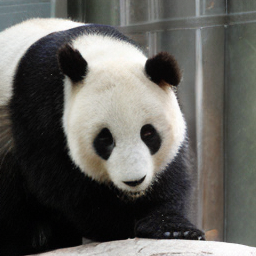} &
\includegraphics[width=0.20\linewidth]{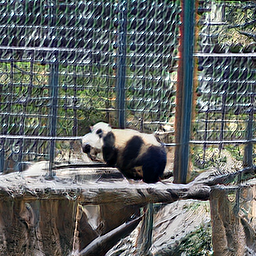} \\
{\scriptsize \textcolor{green!55!black}{\textbf{DiT}\,\ding{51}}} & {\scriptsize SG-XL} &
{\scriptsize \textcolor{green!55!black}{\textbf{DiT}\,\ding{51}}} & {\scriptsize SG-XL} \\
\end{tabular}
\caption{\textbf{DiT preferred over StyleGAN-XL.} 8 representative same-class trials
(sampled from the human study logs) in which participants judged DiT (left of each pair, preferred)
to have more realistic texture than StyleGAN-XL (right).}
\label{fig:dit_sg_xl}
\end{figure}

\begin{figure}[t]
\centering
\setlength{\tabcolsep}{2pt}
\renewcommand{\arraystretch}{1.0}
\begin{tabular}{cccc}

\includegraphics[width=0.20\linewidth]{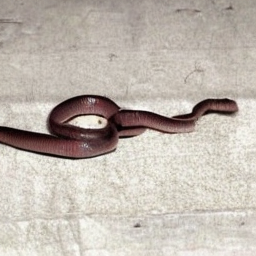} &
\includegraphics[width=0.20\linewidth]{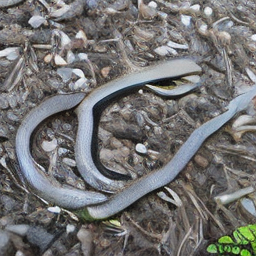} &
\includegraphics[width=0.20\linewidth]{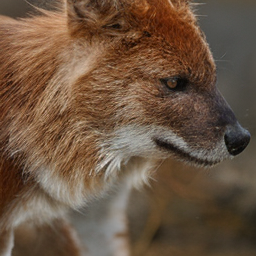} &
\includegraphics[width=0.20\linewidth]{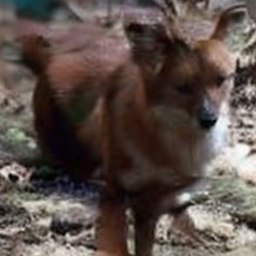} \\
{\scriptsize \textcolor{green!55!black}{\textbf{SiT-SDE}\,\ding{51}}} & {\scriptsize SoFlow} &
{\scriptsize \textcolor{green!55!black}{\textbf{SiT-SDE}\,\ding{51}}} & {\scriptsize SoFlow} \\[6pt]

\includegraphics[width=0.20\linewidth]{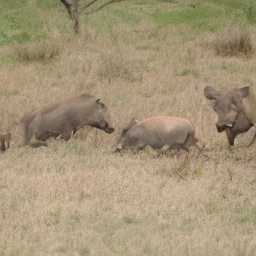} &
\includegraphics[width=0.20\linewidth]{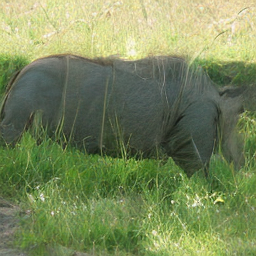} &
\includegraphics[width=0.20\linewidth]{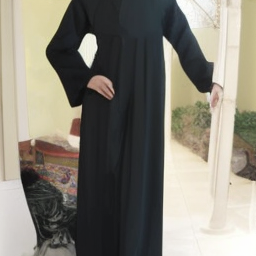} &
\includegraphics[width=0.20\linewidth]{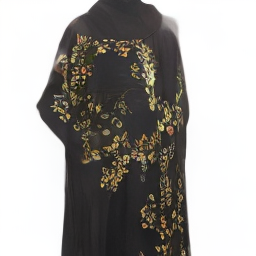} \\
{\scriptsize \textcolor{green!55!black}{\textbf{SiT-SDE}\,\ding{51}}} & {\scriptsize SoFlow} &
{\scriptsize \textcolor{green!55!black}{\textbf{SiT-SDE}\,\ding{51}}} & {\scriptsize SoFlow} \\[6pt]

\includegraphics[width=0.20\linewidth]{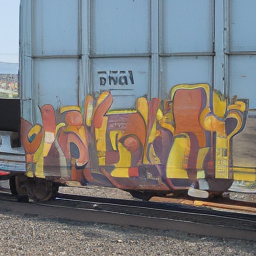} &
\includegraphics[width=0.20\linewidth]{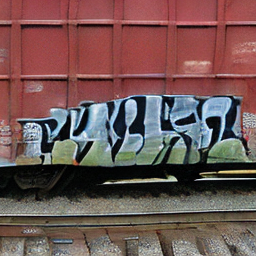} &
\includegraphics[width=0.20\linewidth]{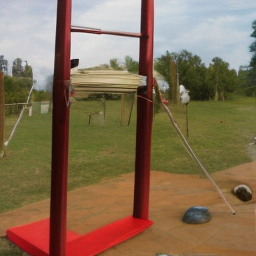} &
\includegraphics[width=0.20\linewidth]{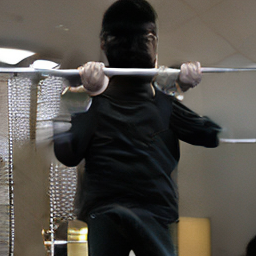} \\
{\scriptsize \textcolor{green!55!black}{\textbf{SiT-SDE}\,\ding{51}}} & {\scriptsize SoFlow} &
{\scriptsize \textcolor{green!55!black}{\textbf{SiT-SDE}\,\ding{51}}} & {\scriptsize SoFlow} \\[6pt]

\includegraphics[width=0.20\linewidth]{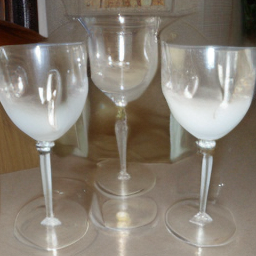} &
\includegraphics[width=0.20\linewidth]{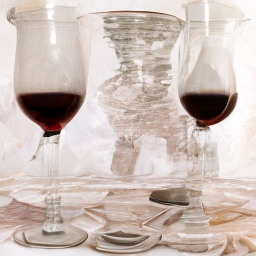} &
\includegraphics[width=0.20\linewidth]{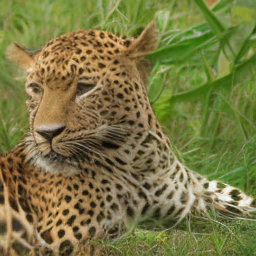} &
\includegraphics[width=0.20\linewidth]{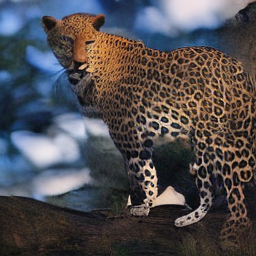} \\
{\scriptsize \textcolor{green!55!black}{\textbf{SiT-SDE}\,\ding{51}}} & {\scriptsize SoFlow} &
{\scriptsize \textcolor{green!55!black}{\textbf{SiT-SDE}\,\ding{51}}} & {\scriptsize SoFlow} \\[6pt]
\end{tabular}
\caption{\textbf{SiT-SDE preferred over SoFlow.} 8 representative same-class trials in
which participants judged SiT-SDE (left, preferred) to have more realistic texture than
SoFlow (right).}
\label{fig:sit_vs_soflow}
\end{figure}

Although generators are harder to separate than real-vs-generated pairs, yielding partially overlapping intervals in Figure~\ref{fig:human_ranking}, a small number of contrasts are nonetheless consistent among decisive (non-tie) judgments: 
\begin{itemize}
\item \textbf{DiT $>$ StyleGAN-XL}: across 102 trials, participants chose DiT 42 times and
StyleGAN-XL 16 times, with 44 ties.
\item \textbf{SiT-SDE $>$ SoFlow}: across 83 trials, participants chose SiT-SDE 27 times and
SoFlow 18 times, with 38 ties.
\end{itemize}
Figures~\ref{fig:dit_sg_xl}--\ref{fig:sit_vs_soflow} show representative decisive trials for each contrast, where the texture-realism preference is visible.

\subsection{Bootstrap Confidence Intervals}
We quantify uncertainty in the metric--human correlations using two non-parametric bootstrap procedures. Our primary analysis uses a trial-level bootstrap that samples 4{,}800 individual trials with replacement, and we also report a participant-level cluster bootstrap that samples the 24 participants with replacement while retaining all trials from
each selected participant. Table~\ref{tab:bootstrap_ci} shows that the conclusions are stable across both resampling procedures. TRIG achieves the strongest point-estimate correlation with human judgments, $\rho=-0.750$, with a primary bootstrap interval of $[-0.817,-0.550]$; its interval excludes zero under both procedures.

CMMD attains the closest agreement among baselines ($\rho=-0.683$), with an interval overlapping TRIG's, yet it is near zero on RAISE ($+0.044$). This is not surprising: although we frame the task around texture rather than content, the two are distinct but not perfectly separable, and several participants reported that when both images look clearly fake, offering no decisive textural cue, they fall back on semantic impressions. This fallback arises in the generator-vs-generator comparisons that are ambiguous because neither image has good texture; when a genuine texture difference exists, observers judge on texture and TRIG tracks them. The residual semantic component thus helps a semantic metric like CMMD on the ambiguous ranking task, but it is absent on RAISE's per-image ratings, where CMMD fails. Among the evaluated metrics, TRIG shows the most consistent point-estimate agreement across both studies.

\begin{table}[t]
\centering

\small
\resizebox{\columnwidth}{!}{
    \small
    \setlength{\tabcolsep}{5pt}
    \setlength{\abovecaptionskip}{-5pt}
    \setlength{\belowcaptionskip}{-5pt}
\begin{tabular}{lrrrr}
\toprule
Metric
& $\rho$
& Trial CI
& Participant CI \\
\midrule
FID
& $+0.467$
& $[+0.300,+0.517]$
& $[+0.267,+0.533]$ \\

KID
& $-0.100$
& $[-0.300,+0.033]$
& $[-0.283,+0.017]$ \\

Gram-MMD (DINOv2)
& $-0.600$
& $[-0.767,-0.467]$
& $[-0.733,-0.500]$ \\

Gram-MMD (DCAE)
& $-0.283$
& $[-0.433,-0.167]$
& $[-0.450,-0.167]$ \\

CMMD
& $-0.683$
& $[-0.800,-0.617]$
& $[-0.783,-0.617]$ \\

\textbf{TRIG}
& $\mathbf{-0.750}$
& $\mathbf{[-0.817,-0.550]}$
& $\mathbf{[-0.817,-0.567]}$ \\
\bottomrule
\end{tabular}
}
\caption{Spearman correlation with the human source ranking and percentile 95\% confidence intervals under different bootstrap procedures. The primary interval uses trial resampling, and an alternative resampling of participants. All
correlations are computed over $n=9$ non-anchor sources. More negative values indicate better agreement with human judgments.}
\label{tab:bootstrap_ci}
\end{table}

\subsection{Real Reference for Distribution Metrics}
The human study contains 9{,}509 distinct images across 4{,}800 pairs: 976 from ImageNet-anchor, 994 from ImageNet-eval, 963 from ADM, 946 from BigGAN, 965 from DiT, 935 from PMF, 975 from StyleGAN-XL, 895 from SiT-ODE, 918 from SiT-SDE, and 942 from SoFlow. For each source, its distributional
distance is computed using the distinct images of that source shown to participants. To score each generator, we use a held-out ImageNet anchor sampled with a fixed seed from ImageNet official test set that is disjoint
from the human-study images. It contains one randomly selected image from each of the 1{,}000 ImageNet classes and has no overlap with either the human-study images or our backbone training set. We denote this subset \textbf{ImageNet-Study-Anchor} ($\mathcal{A}$). For each metric we compute $\operatorname{dist}(\mathcal{A}, s)$ between the anchor and each of the eight generator sources $s$, and correlate the resulting distances with the human scores.Because lower distances indicate greater realism, negative correlations are expected. TRIG produces nearly identical correlations whether the anchor is the ImageNet images shown in the study or the disjoint ImageNet-Study-Anchor, indicating that the result is robust to the choice of reference images. We use 1{,}000 anchor images to give balanced coverage of all classes and to match the size of the COCO anchor used in the RAISE study.

\subsection{Generated-Image Sources}
\label{sec:generated-sources}

The human study includes eight generators, while the broader evaluation includes four additional models. Every source is class-balanced, with 50 images from each of the 1{,}000 ImageNet classes (50{,}000 images total). Integer labels follow the ImageNet-1k class order used by each official
checkpoint; no text prompts or class-name remapping are used. Images are generated and evaluated at $256\times256$.

\begin{table*}[t]
\centering
\tiny
\setlength{\tabcolsep}{2.5pt}
\renewcommand{\arraystretch}{1.03}
\begin{tabular}{@{}
p{0.10\linewidth}
p{0.24\linewidth}
p{0.37\linewidth}
p{0.23\linewidth}
@{}}
\toprule
Model & Official code & Sampling & Parameters \\
\midrule

ADM-G &
https://github.com/openai/guided-diffusion &
Class-conditional DDIM sampling with 1000 steps. &
Classifier guidance 1.0; FP16. \\

BigGAN-deep &
https://github.com/huggingface/pytorch-pretrained-BigGAN &
One synthesis pass with truncated-normal latents and one-hot class labels. &
Truncation 1.0. \\

DiT-XL/2 &
https://github.com/facebookresearch/DiT &
Class-conditional latent diffusion with 250 DDPM steps. &
CFG 1.5. \\

SiT-XL/2 ODE &
https://github.com/willisma/SiT &
Class-conditional latent-flow sampling with adaptive Dopri5. &
CFG 1.5; linear transport; velocity prediction;
$\mathrm{atol}=10^{-6}$, $\mathrm{rtol}=10^{-3}$. \\

SiT-XL/2 SDE &
https://github.com/willisma/SiT &
Euler--Maruyama SDE sampling with 250 steps. &
CFG 1.5; \texttt{diffusion\_form=sigma}; norm 1.0;
final estimator \texttt{Mean}, size 0.04. \\

StyleGAN-XL &
https://github.com/autonomousvision/stylegan-xl &
One synthesis pass with one-hot class labels and constant synthesis noise. &
Truncation $\psi=1.0$. \\

SoFlow-XL/2 &
https://github.com/zlab-princeton/SoFlow &
One-step conditional solution operator in SD-VAE latent space
($\mathrm{NFE}=1$). &
Official inference configuration; training
\texttt{cfg\_drop\_rate=0.1}. \\

pMF-H/16 &
https://github.com/Lyy-iiis/pMF &
One-step pixel-space MeanFlow sampling. &
Interval CFG $\omega=7.0$ for $t\in[0.1,0.7]$. \\

\bottomrule
\end{tabular}
\caption{Sampling configurations for the eight generators included in the
human study. Sampling steps, guidance settings, and other hyperparameters
follow the corresponding official implementations.}
\label{tab:human-generator-sources}
\end{table*}

\begin{table*}[t]
\centering
\tiny
\setlength{\tabcolsep}{2.5pt}
\renewcommand{\arraystretch}{1.03}
\begin{tabular}{@{}
p{0.10\linewidth}
p{0.24\linewidth}
p{0.37\linewidth}
p{0.23\linewidth}
@{}}
\toprule
Model & Official code & Sampling & Parameters \\
\midrule

Self-Flow &
https://github.com/black-forest-labs/Self-Flow &
Class-conditional SDE sampling with 250 steps. &
CFG 1.0; guidance interval $[0,0.7]$. \\

VAR-d30 &
https://github.com/FoundationVision/VAR &
Coarse-to-fine next-scale autoregressive sampling with hard token selection. &
CFG 1.5; top-$k=900$; top-$p=0.96$;
\texttt{more\_smooth=False}. \\

RAR-XXL &
https://github.com/bytedance/1d-tokenizer &
Randomized autoregressive token sampling followed by VQGAN decoding. &
Temperature 1.02; guidance 8.0; guidance exponent 1.2. \\

LlamaGen 3B &
https://github.com/FoundationVision/LlamaGen &
Autoregressive sampling of $24^2$ VQ tokens followed by VQ decoding.
The $384^2$ outputs are bilinearly resized to $256^2$. &
CFG 1.65; temperature 1.0; top-$k=0$; top-$p=1.0$;
\texttt{cfg\_interval=-1}. \\

\bottomrule
\end{tabular}
\caption{Additional generators in the twelve-source evaluation. Sampling
steps, guidance settings, and other hyperparameters follow the corresponding
official implementations.}
\label{tab:additional-generator-sources}
\end{table*}

\paragraph{Licenses.}
The official ADM, BigGAN, SiT, StyleGAN-XL, SoFlow, pMF, VAR, and LlamaGen repositories use MIT licenses; Self-Flow and RAR use Apache 2.0, and DiT uses CC BY-NC 4.0. These are code-repository licenses; checkpoint and ImageNet terms may impose additional restrictions.

\paragraph{Randomness.}
Training, data sampling, and degradation sampling use seed 42 for Python, NumPy, and PyTorch. Fresh-anchor selection and SWD projection directions also use seed 42, while the bootstrap and split-half reliability analyses use seed 0. Each human-study session instead draws and stores a fresh 32-bit seed from \texttt{os.urandom}; thus, the 24 sessions do not share a fixed trial sequence. The recorded session seeds are:
\begin{quote}
\scriptsize
\raggedright
223375562, 1957381840, 688338878, 418787041, 1215994554,
1159794409, 1273772766, 1753782588, 15166438, 3281918733,
505716948, 3855776757, 1026885437, 2735518460, 2925330459,
1146088416, 2906388500, 437931220, 766905075, 3604687666,
2538265070, 2955329442, 2129010511, and 146665620.
\end{quote}

\section{RAISE Distribution-Level Grouping Protocol}
\label{sec:supp-raise-grouping}

RAISE provides mean opinion scores (MOS) for individual images but does \emph{not} define the 24 groups used in our distribution-level evaluation. Following the evaluation
construction of Gram-MMD~\cite{napolitano2026gram}, we form these groups ourselves from the 480 AI-generated RAISE images. The groups below are therefore quality strata introduced by
our protocol; they are \emph{not} official RAISE categories, subsets, or train/test splits.

\begin{figure*}[t]
\centering
\setlength{\tabcolsep}{2pt}
\renewcommand{\arraystretch}{0.8}
\begin{tabular}{cccc}
\includegraphics[width=0.23\linewidth]{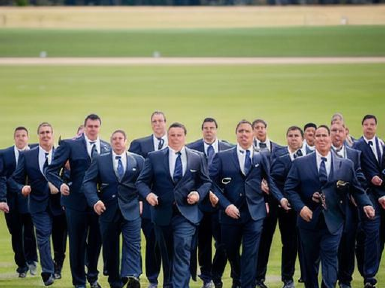} &
\includegraphics[width=0.23\linewidth]{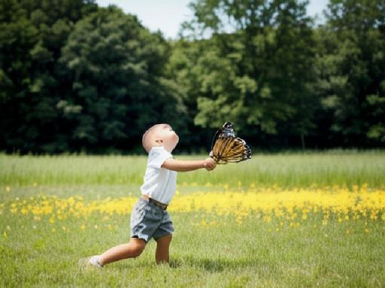} &
\includegraphics[width=0.23\linewidth]{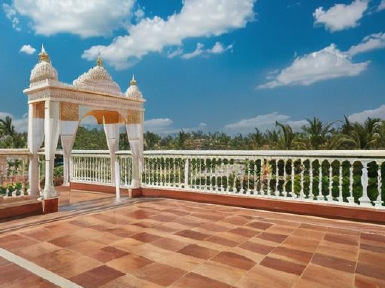} &
\includegraphics[width=0.23\linewidth]{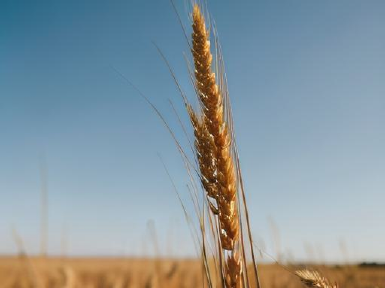} \\
\multicolumn{2}{c}{\scriptsize low MOS (less realistic)} &
\multicolumn{2}{c}{\scriptsize high MOS (more realistic)} \\
\end{tabular}
\caption{\textbf{Example images from the RAISE benchmark~\cite{mukherjee2025raise}.} AI-generated images shown across the perceived-realness range, from low mean opinion score (left) to high (right). Our protocol sorts the 480 AI-generated images by MOS and partitions them into 24 quality strata (Section~\ref{sec:supp-raise-grouping}).}
\label{fig:raise_examples}
\end{figure*}

Let $m_i$ denote the MOS of AI-generated image $x_i$. We sort all 480 images in ascending MOS order,
\begin{equation}
m_{(1)} \leq m_{(2)} \leq \cdots \leq m_{(480)},
\end{equation}
and divide the ordered list into 24 consecutive, non overlapping groups of 20 images,
\begin{equation}
G_j = \left\{x_{(20(j-1)+1)},\ldots,x_{(20j)}\right\}, \qquad j=1,\ldots,24 .
\end{equation}
The human target for group $G_j$ is its mean MOS,
$\overline{m}_j = \tfrac{1}{20}\sum_{x_i\in G_j} m_i$. This construction is deterministic, uses no random seed, and depends only on the human MOS to determine group membership. We provide some examples in the dataset with low MOS and high MOS in \ref{fig:raise_examples}.

For every evaluated distance $D$, we compare each group against a single fixed reference distribution $A$ of 1{,}000 images sampled without replacement from MS-COCO \texttt{val2017} (seed 42), giving $d_j = D(A, G_j)$. We then compute Spearman's $\rho$ and  Kendall's $\tau$ between the 24 distances $\{d_j\}_{j=1}^{24}$ and the group-mean MOS values $\{\overline{m}_j\}_{j=1}^{24}$; the sample size for these correlations is thus $n=24$, not $n=480$. Because all reported metrics are distances (lower is better), agreement with increasing perceived realness appears as a \emph{negative} correlation.

This protocol tests whether a distribution metric is monotonic across MOS-defined quality strata; it is not an image-level correlation analysis, and the MOS-based sorting is not an
independently randomized partition. The same 24 groups and the same COCO anchor are used for every metric, so metric comparisons are paired and differ only in the scoring function.

\section{Does the metric respond to semantics or texture?}
\label{sec:supp-semantic-texture}
\begin{figure*}[t]
\centering
\setlength{\tabcolsep}{1pt}
\renewcommand{\arraystretch}{0.6}
\begin{tabular}{@{}ccc@{}}
\includegraphics[width=0.33\linewidth]{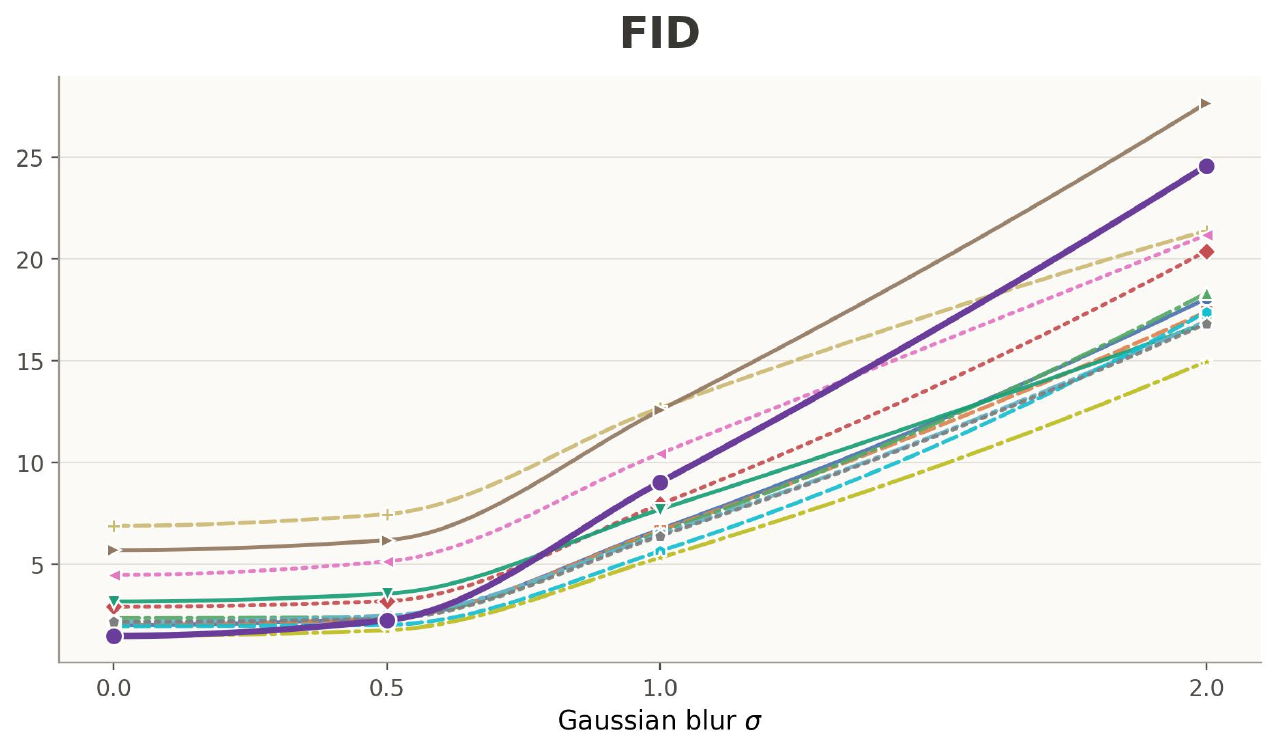} &
\includegraphics[width=0.33\linewidth]{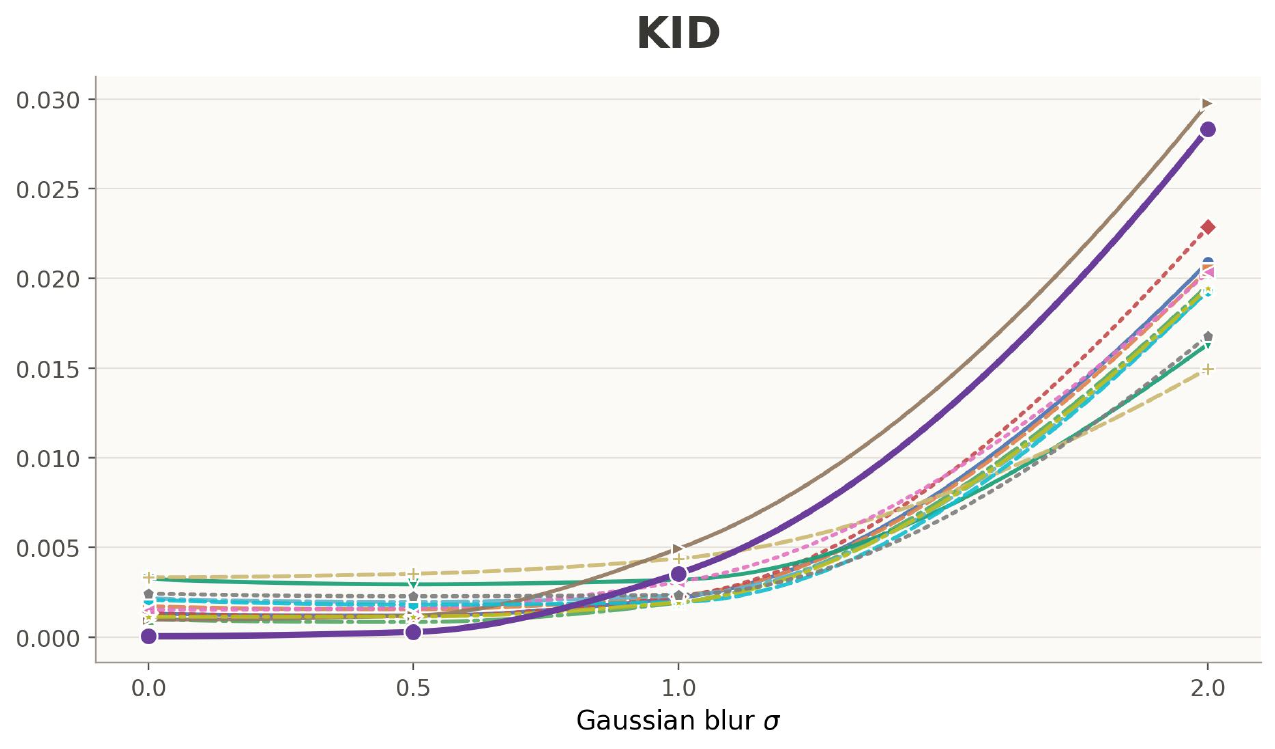} &
\includegraphics[width=0.33\linewidth]{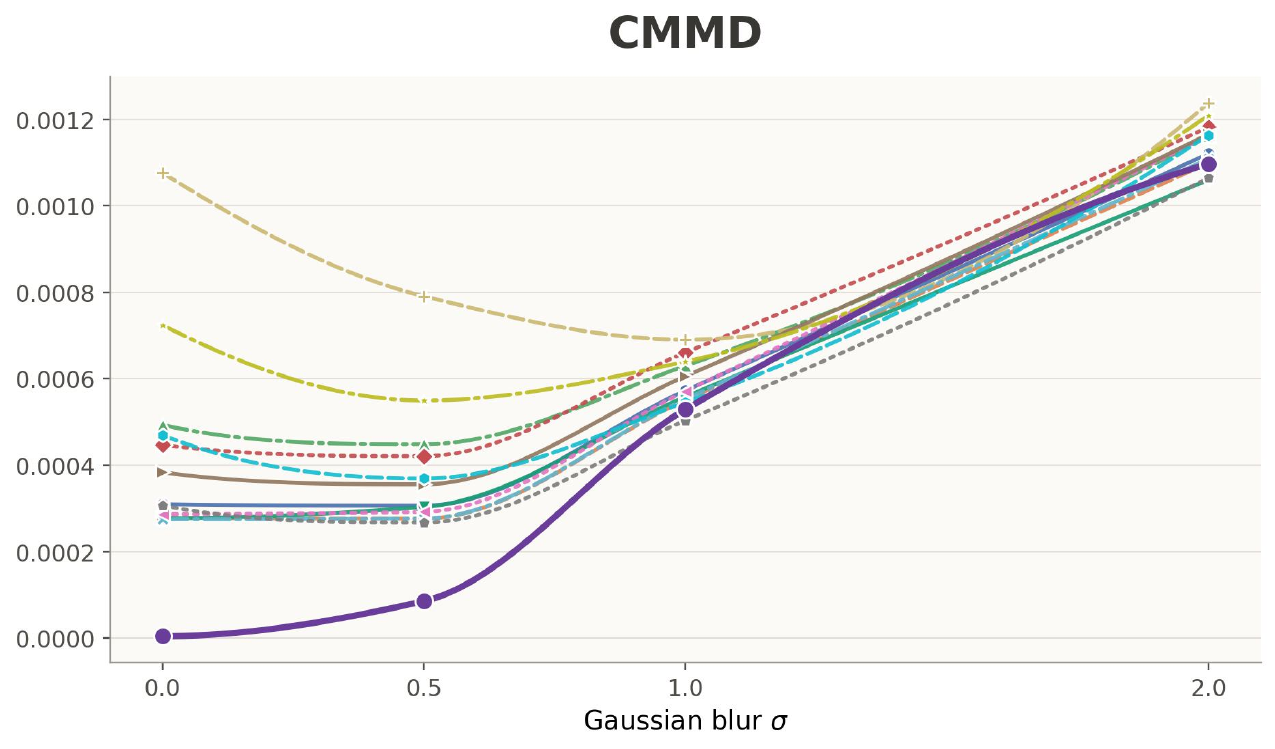} \\
\includegraphics[width=0.33\linewidth]{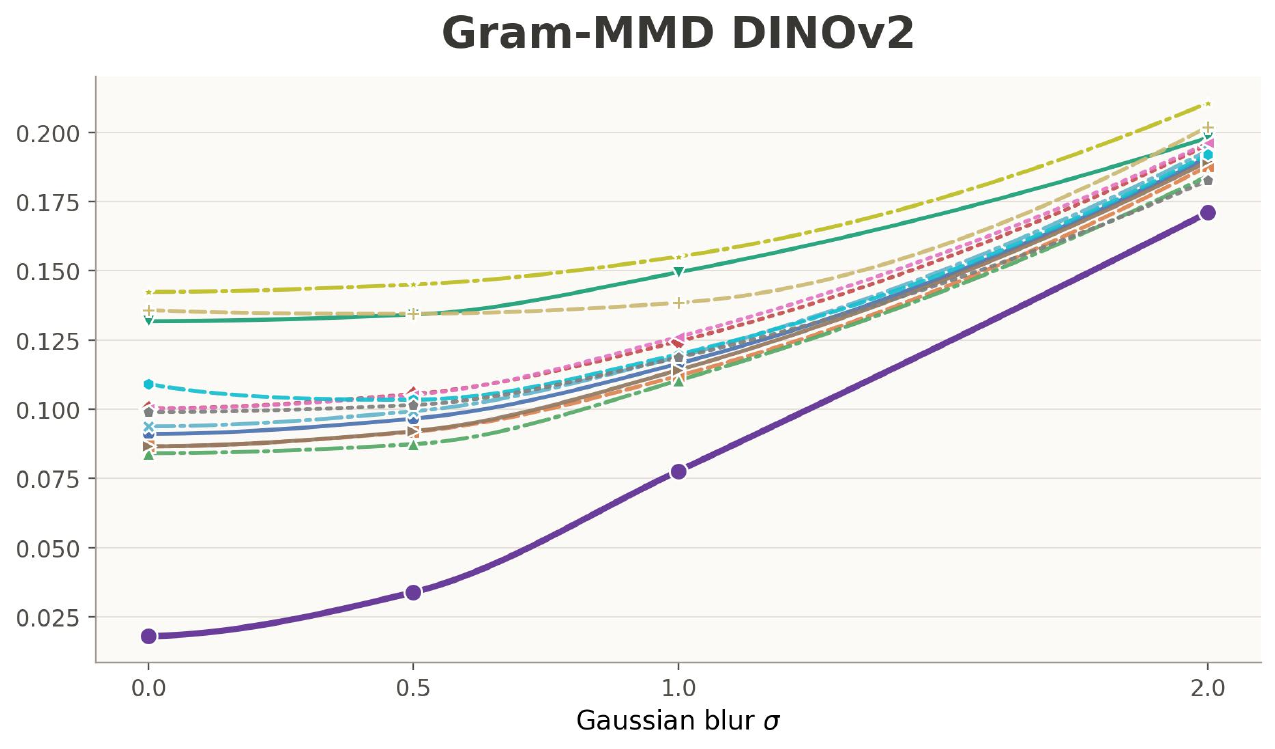} &
\includegraphics[width=0.33\linewidth]{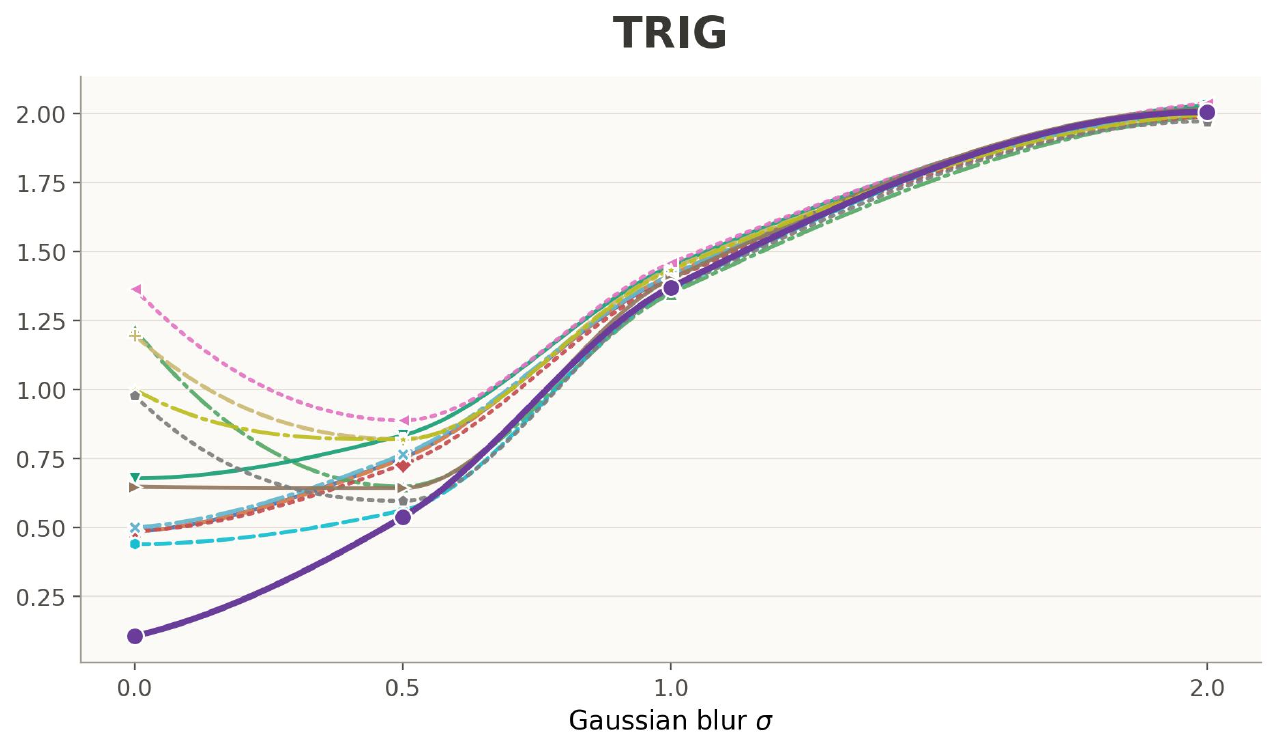} &
\includegraphics[width=0.33\linewidth]{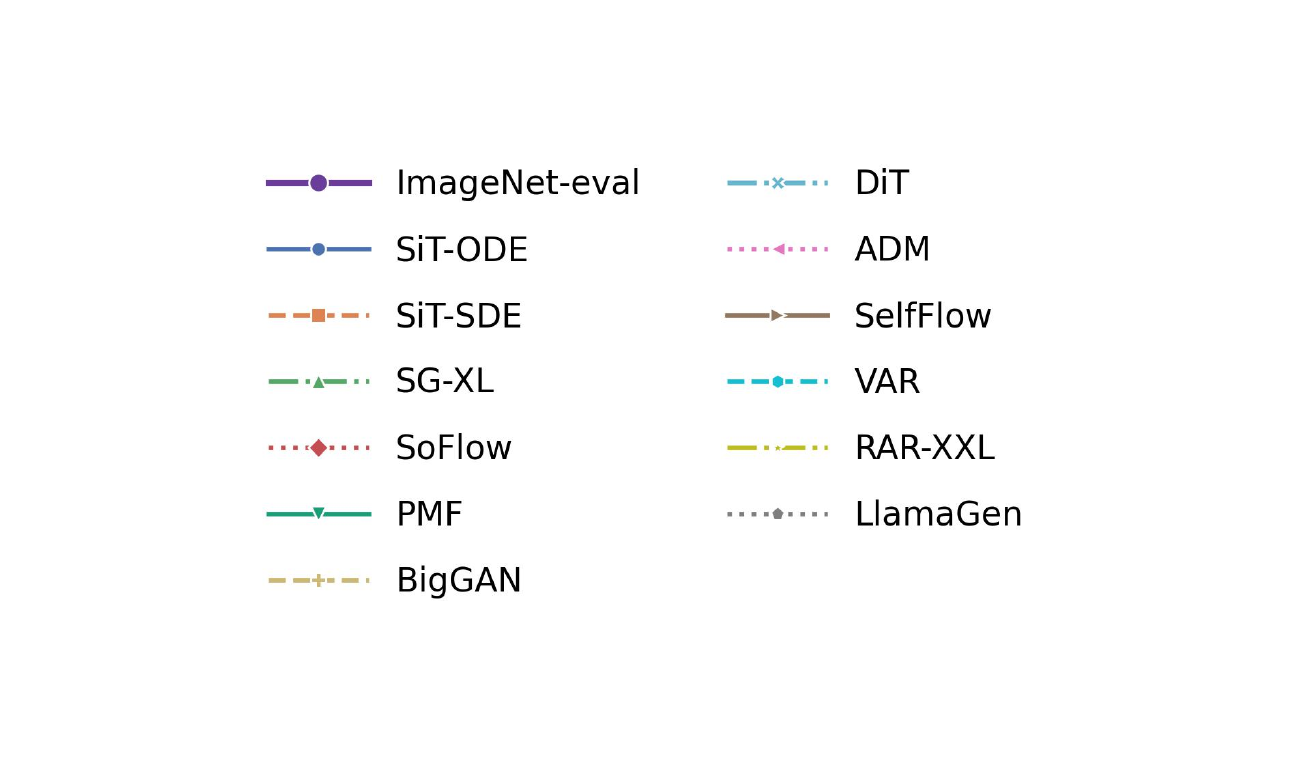} 
\\
\end{tabular}
\caption{\textbf{Metric score vs.\ blur strength for the real ImageNet set and 12 generators.} As blur becomes severe, FID and KID rate the generated sets as \emph{more} realistic than the real set (wrong direction); CMMD and Gram-MMD (DINOv2) keep separating the sources even when texture is largely destroyed; TRIG's scores converge as texture detail is removed, consistent with a metric driven by texture quality.}
\label{fig:exp6_blur}
\end{figure*}

We probe what each metric responds to by adding Gaussian blur of increasing strength to all sources (full 50k). Blur preserves semantic content while progressively destroying texture detail, so a texture-oriented metric should react strongly to it, whereas a semantics-oriented metric should be comparatively insensitive. Figure~\ref{fig:exp6_blur} shows each metric's score for the real ImageNet-eval set and 12 generators as blur strength increases; the metrics respond in markedly different ways. FID and KID, built on an Inception classifier that encodes ``what is in the image,'' are nearly invariant to mild blur and then respond in the \textbf{wrong} direction: under heavy blur they rate the real ImageNet set as less realistic than the generated sets, revealing that their features track semantic content rather than texture. CMMD and Gram-MMD continue to separate the sources even under strong blur; since blur has by then removed most texture information, this residual separability indicates that they too rank largely by semantic or global structure rather than texture. TRIG behaves oppositely: as blur destroys texture detail, the scores of all sources converge, suggesting its sensitivity to texture realism. This convergence is the expected behavior of a metric driven by texture quality, providing evidence that TRIG is particularly sensitive to texture degradation rather than relying primarily on semantic content.

\section{Where does the backbone respond?}
\label{sec:supp_heatmaps}

\subsection{Response to Manual Degradations}
\begin{figure*}[t]
\centering
\setlength{\tabcolsep}{2pt}
\begin{tabular}{cccccccc}

Original &Gaussian blur & Box blur & Motion blur & Bicubic resampling & Freq. suppress & Phase smooth & Noise \\
\includegraphics[width=0.12\linewidth]{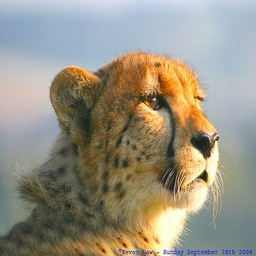} &
\includegraphics[width=0.12\linewidth]{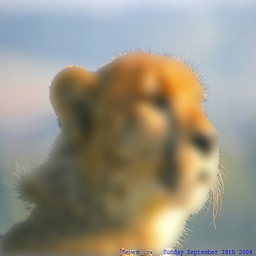} &
\includegraphics[width=0.12\linewidth]{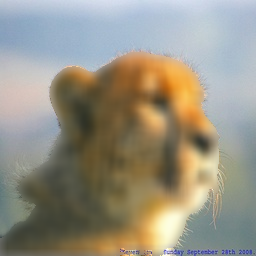} &
\includegraphics[width=0.12\linewidth]{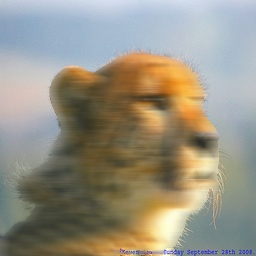} &

\includegraphics[width=0.12\linewidth]{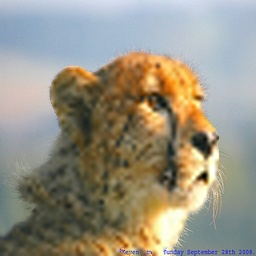} &
\includegraphics[width=0.12\linewidth]{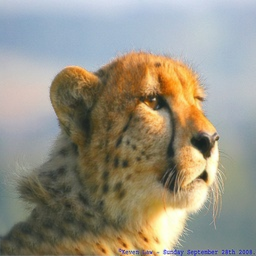} &
\includegraphics[width=0.12\linewidth]{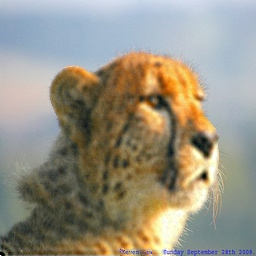} &
\includegraphics[width=0.12\linewidth]{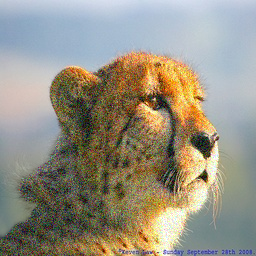} \\[0.5pt]

\multicolumn{8}{c}{\scriptsize Heatmaps} \\[3pt]

\includegraphics[width=0.12\linewidth]{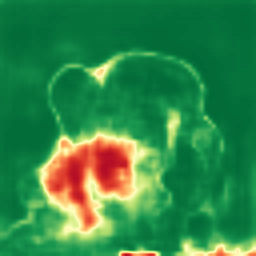} &
\includegraphics[width=0.12\linewidth]{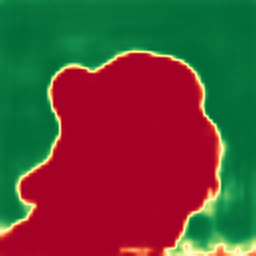} &
\includegraphics[width=0.12\linewidth]{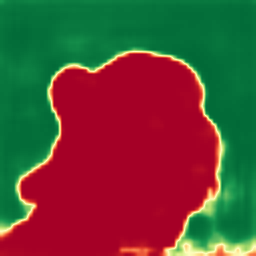} &
\includegraphics[width=0.12\linewidth]{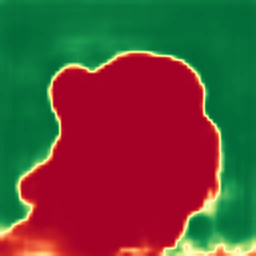} & 

\includegraphics[width=0.12\linewidth]{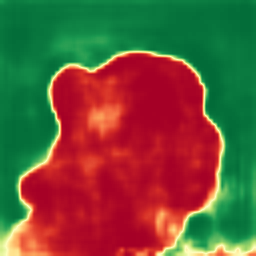} &
\includegraphics[width=0.12\linewidth]{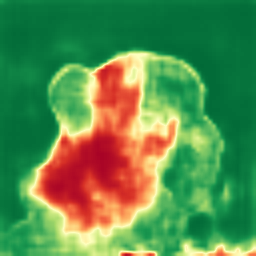} &
\includegraphics[width=0.12\linewidth]{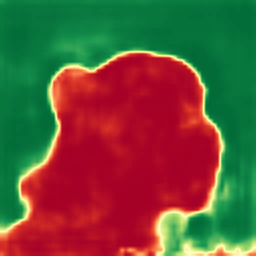} &
\includegraphics[width=0.12\linewidth]{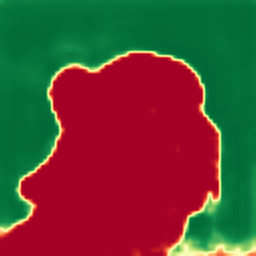}\\ 

\multicolumn{4}{c}{\scriptsize Blurred images} & \multicolumn{4}{c}{\scriptsize JPEG compressed images} \\[3pt]
n=5 & n=15 & n=40 &n=80 & q=50 & q=30 & q=15 & q=5 \\
\includegraphics[width=0.12\linewidth]{Figures/exp5/manual_attack/0286_img_000/blur_block_many__n5_image.png} &
\includegraphics[width=0.12\linewidth]{Figures/exp5/manual_attack/0286_img_000/blur_block_many__n15_image.png} &
\includegraphics[width=0.12\linewidth]{Figures/exp5/manual_attack/0286_img_000/blur_block_many__n40_image.png} &
\includegraphics[width=0.12\linewidth]{Figures/exp5/manual_attack/0286_img_000/blur_block_many__n80_image.png} &

\includegraphics[width=0.12\linewidth]{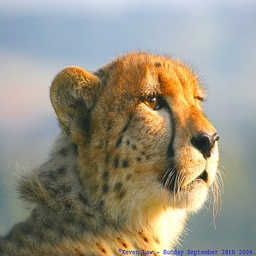} &
\includegraphics[width=0.12\linewidth]{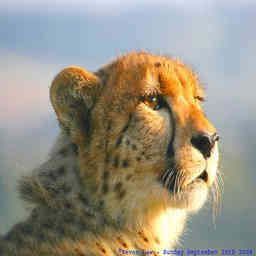} &
\includegraphics[width=0.12\linewidth]{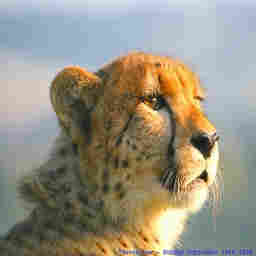} &
\includegraphics[width=0.12\linewidth]{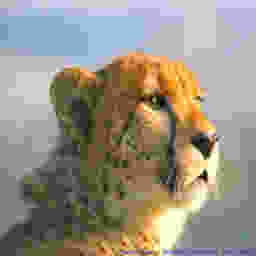} \\[0.5pt]

\multicolumn{8}{c}{\scriptsize Heatmaps} \\[3pt]

\includegraphics[width=0.12\linewidth]{Figures/exp5/manual_attack/0286_img_000/blur_block_many__n5_heatmap.png} &
\includegraphics[width=0.12\linewidth]{Figures/exp5/manual_attack/0286_img_000/blur_block_many__n15_heatmap.png} &
\includegraphics[width=0.12\linewidth]{Figures/exp5/manual_attack/0286_img_000/blur_block_many__n40_heatmap.png} &
\includegraphics[width=0.12\linewidth]{Figures/exp5/manual_attack/0286_img_000/blur_block_many__n80_heatmap.png} & 

\includegraphics[width=0.12\linewidth]{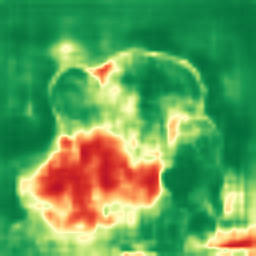} &
\includegraphics[width=0.12\linewidth]{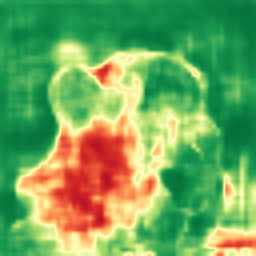} &
\includegraphics[width=0.12\linewidth]{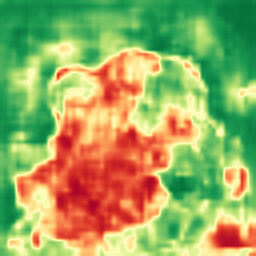} &
\includegraphics[width=0.12\linewidth]{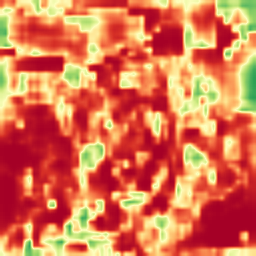}\\ 

Original &Gaussian blur & Box blur & Motion blur & Bicubic resampling & Freq. suppress & Phase smooth & Noise \\
\includegraphics[width=0.12\linewidth]{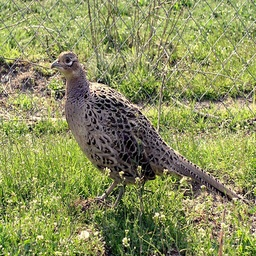} &
\includegraphics[width=0.12\linewidth]{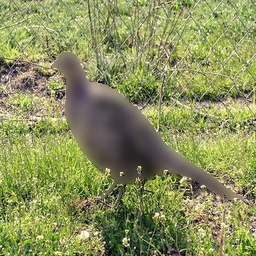} &
\includegraphics[width=0.12\linewidth]{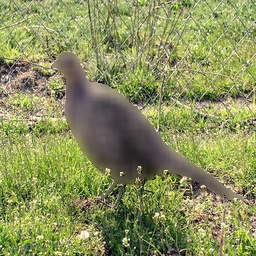} &
\includegraphics[width=0.12\linewidth]{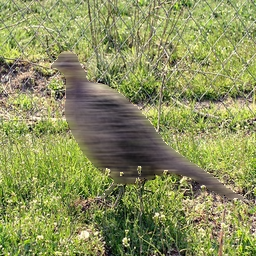} &

\includegraphics[width=0.12\linewidth]{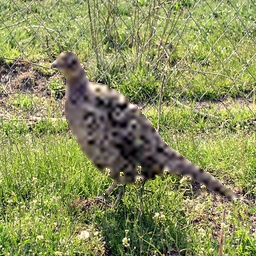} &
\includegraphics[width=0.12\linewidth]{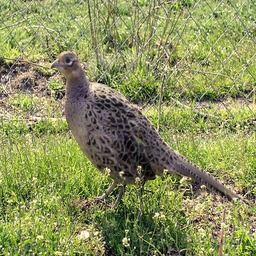} &
\includegraphics[width=0.12\linewidth]{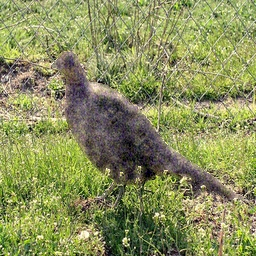} &
\includegraphics[width=0.12\linewidth]{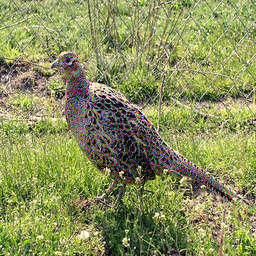} \\[0.5pt]

\multicolumn{8}{c}{\scriptsize Heatmaps} \\[3pt]

\includegraphics[width=0.12\linewidth]{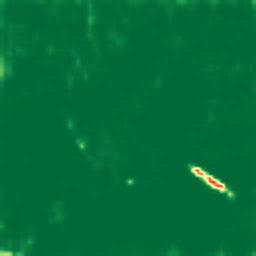} &
\includegraphics[width=0.12\linewidth]{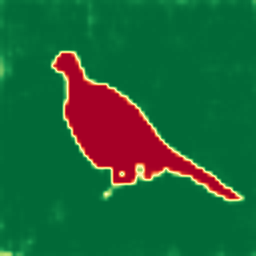} &
\includegraphics[width=0.12\linewidth]{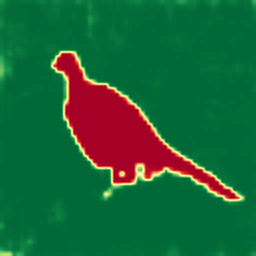} &
\includegraphics[width=0.12\linewidth]{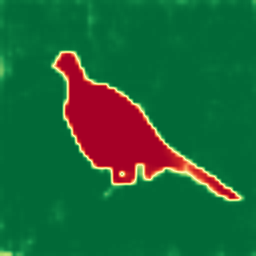} & 

\includegraphics[width=0.12\linewidth]{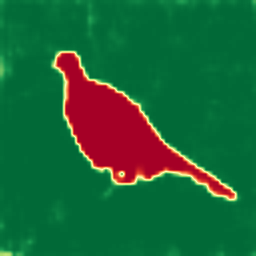} &
\includegraphics[width=0.12\linewidth]{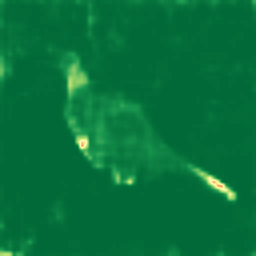} &
\includegraphics[width=0.12\linewidth]{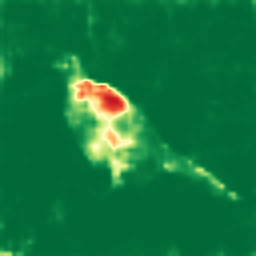} &
\includegraphics[width=0.12\linewidth]{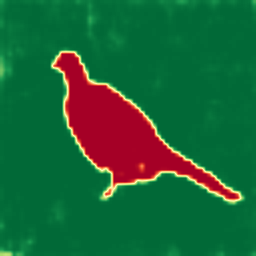}\\ 

\multicolumn{4}{c}{\scriptsize Blurred images} & \multicolumn{4}{c}{\scriptsize JPEG compressed images} \\[3pt]
n=5 & n=15 & n=40 &n=80 & q=50 & q=30 & q=15 & q=5 \\
\includegraphics[width=0.12\linewidth]{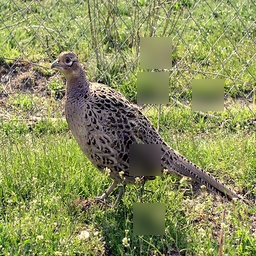} &
\includegraphics[width=0.12\linewidth]{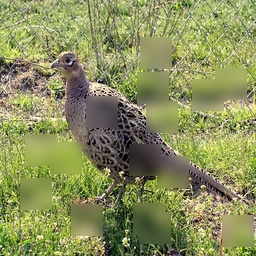} &
\includegraphics[width=0.12\linewidth]{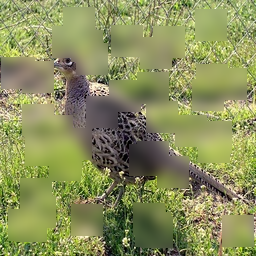} &
\includegraphics[width=0.12\linewidth]{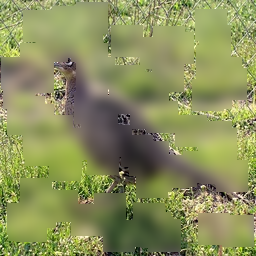} &

\includegraphics[width=0.12\linewidth]{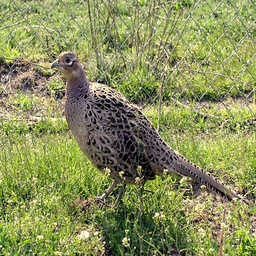} &
\includegraphics[width=0.12\linewidth]{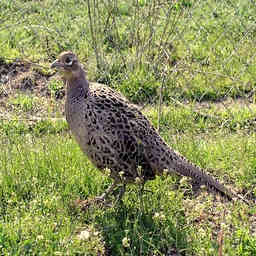} &
\includegraphics[width=0.12\linewidth]{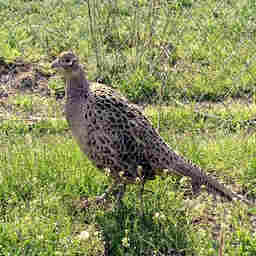} &
\includegraphics[width=0.12\linewidth]{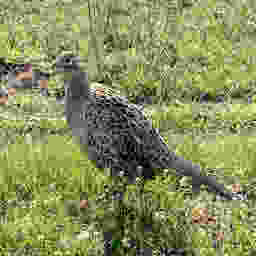} \\[0.5pt]

\multicolumn{8}{c}{\scriptsize Heatmaps} \\[3pt]

\includegraphics[width=0.12\linewidth]{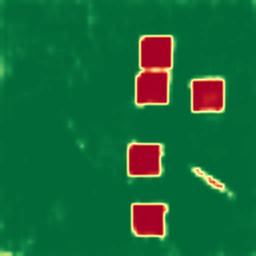} &
\includegraphics[width=0.12\linewidth]{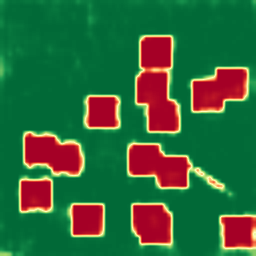} &
\includegraphics[width=0.12\linewidth]{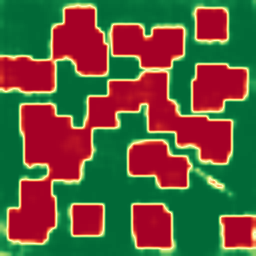} &
\includegraphics[width=0.12\linewidth]{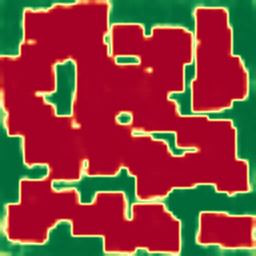} & 

\includegraphics[width=0.12\linewidth]{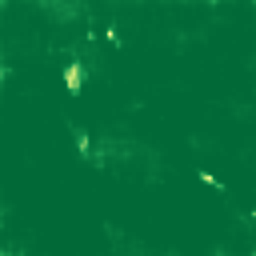} &
\includegraphics[width=0.12\linewidth]{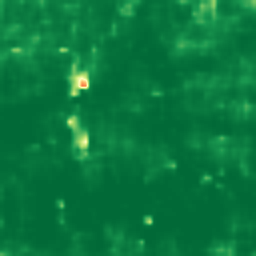} &
\includegraphics[width=0.12\linewidth]{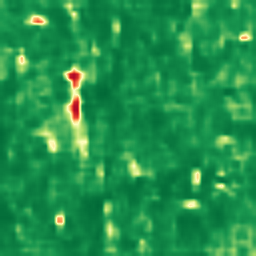} &
\includegraphics[width=0.12\linewidth]{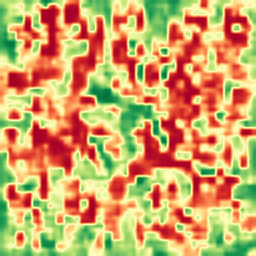}\\ 

\end{tabular}

\caption{ Several manual attacks on the texture-realism backbone. It is sensitive to different degradation operators and JPEG degradations, and the low responses (red regions) correspond to the manually degraded regions.
}
\label{fig:manual_attack_supp}
\end{figure*}

Beyond the generated images evaluated in the main paper, we probe how the texture-realism map of our backbone responds to controlled, manually applied degradations. We refer to Figure ~\ref{fig:manual_attack_supp}, which visualizes the heatmap $\mathbf{M}(I)$ for a real image subjected to each operator in our degradation pool and held-out JPEG compression, where green marks consistent texture and red marks regions judged inconsistent with real-image statistics.

\paragraph{Per-operator response.} We first apply each of the seven degradation operators to the full image. In every case the heatmap turns red over the degraded texture while the background remains green, confirming that TRIG responds to each operator type rather than to a single artifact. The original (undegraded) image yields a nearly uniform green map, and the response is strongest for the operators that most disrupt local texture statistics (Gaussian, box, and motion blur), consistent with the training supervision.

\paragraph{Response to degradation coverage.} Then we apply blurred blocks at increasing coverage ($n=5$ to $n=80$ blocks). The red response grows monotonically with
coverage, indicating that the map reflects the \emph{extent} of degraded texture rather than merely flagging its presence. 

\paragraph{Generalization to held-out degradations.} We also apply JPEG compression at decreasing quality ($q=50$ to $q=5$), which is a degradation \emph{absent} from our training pool. Our backbone nonetheless assigns increasing red response as quality drops and blocking artifacts intensify, showing that the learned representation captures a general notion of inconsistent texture that extends beyond the specific operators used for supervision.

\paragraph{Consistency across content.} We repeat the per-operator comparison on a second image with different content (bird). The response pattern is preserved.

\subsection{Qualitative heatmaps across sources and classes.}
\begin{figure*}[t]
\centering
\setlength{\tabcolsep}{1pt}
\renewcommand{\arraystretch}{0.5}
\newcommand{\rll}[1]{\rotatebox{90}{\scriptsize #1}}  

\begin{tabular}{@{}l*{10}{c}@{}}
\rll{imagenet}
& \includegraphics[width=0.09\linewidth,height=0.09\linewidth]{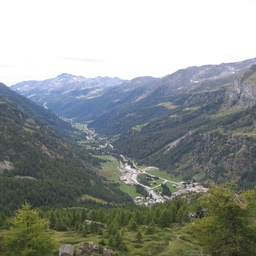} & \includegraphics[width=0.09\linewidth,height=0.09\linewidth]{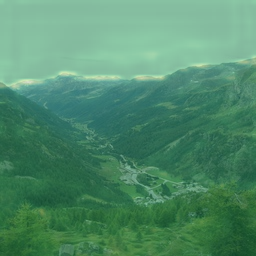} & \includegraphics[width=0.09\linewidth,height=0.09\linewidth]{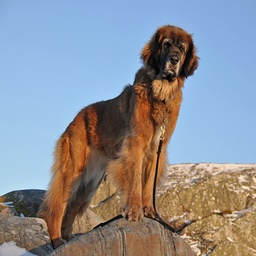}
& \includegraphics[width=0.09\linewidth,height=0.09\linewidth]{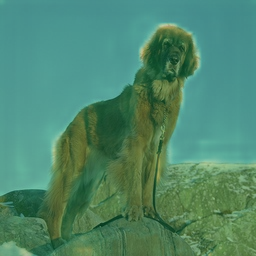} & \includegraphics[width=0.09\linewidth,height=0.09\linewidth]{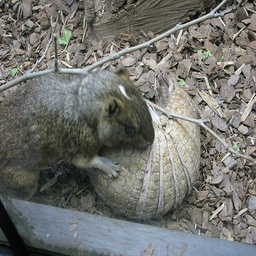} & \includegraphics[width=0.09\linewidth,height=0.09\linewidth]{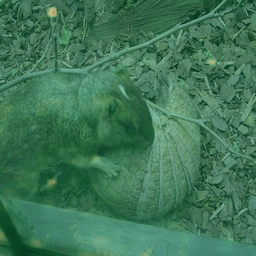}
& \includegraphics[width=0.09\linewidth,height=0.09\linewidth]{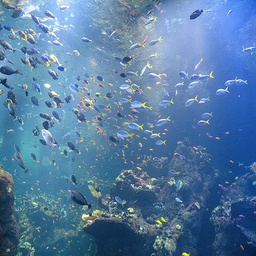} & \includegraphics[width=0.09\linewidth,height=0.09\linewidth]{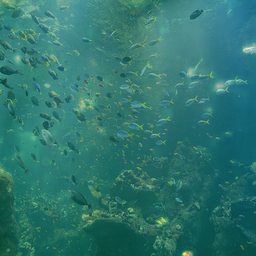} & \includegraphics[width=0.09\linewidth,height=0.09\linewidth]{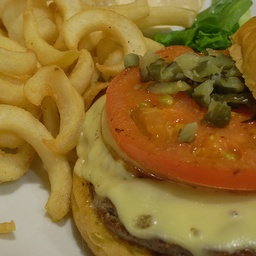}
& \includegraphics[width=0.09\linewidth,height=0.09\linewidth]{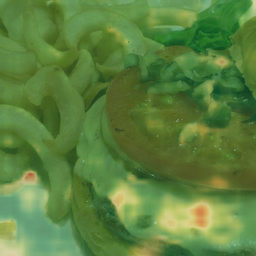} \\

\rll{VAR}
& \includegraphics[width=0.09\linewidth,height=0.09\linewidth]{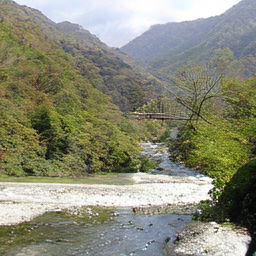} & \includegraphics[width=0.09\linewidth,height=0.09\linewidth]{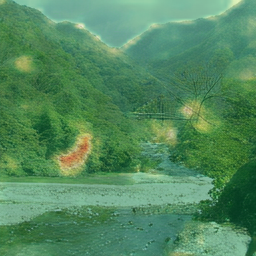} & \includegraphics[width=0.09\linewidth,height=0.09\linewidth]{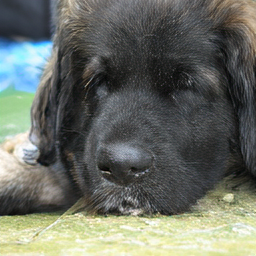}
& \includegraphics[width=0.09\linewidth,height=0.09\linewidth]{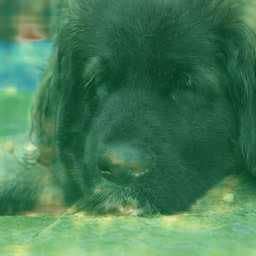} & \includegraphics[width=0.09\linewidth,height=0.09\linewidth]{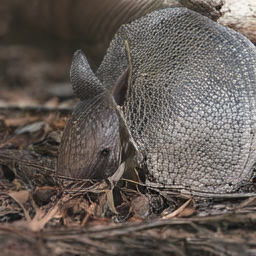} & \includegraphics[width=0.09\linewidth,height=0.09\linewidth]{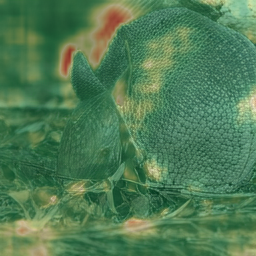}
& \includegraphics[width=0.09\linewidth,height=0.09\linewidth]{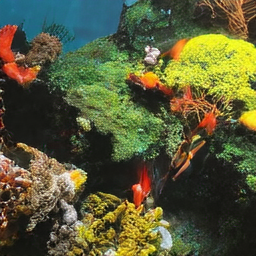} & \includegraphics[width=0.09\linewidth,height=0.09\linewidth]{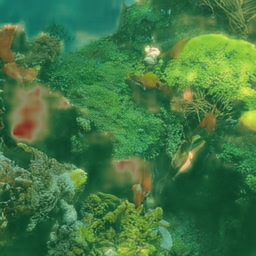} & \includegraphics[width=0.09\linewidth,height=0.09\linewidth]{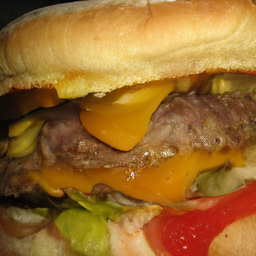}
& \includegraphics[width=0.09\linewidth,height=0.09\linewidth]{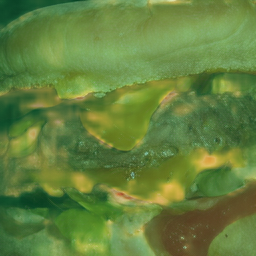} \\

\rll{SiT-ODE}
& \includegraphics[width=0.09\linewidth,height=0.09\linewidth]{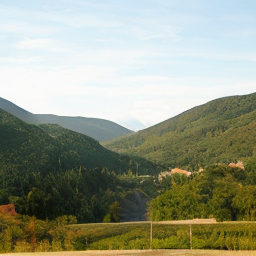} & \includegraphics[width=0.09\linewidth,height=0.09\linewidth]{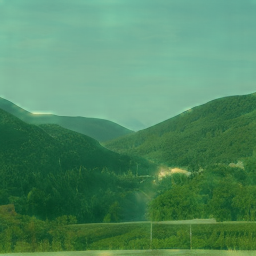} & \includegraphics[width=0.09\linewidth,height=0.09\linewidth]{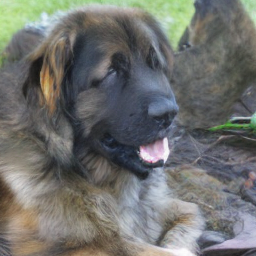}
& \includegraphics[width=0.09\linewidth,height=0.09\linewidth]{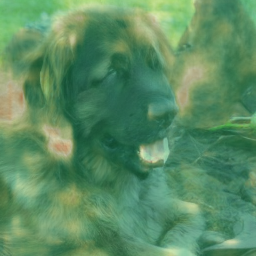} & \includegraphics[width=0.09\linewidth,height=0.09\linewidth]{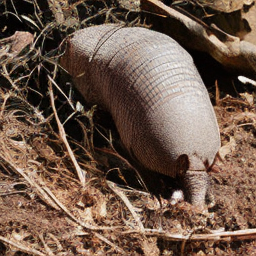} & \includegraphics[width=0.09\linewidth,height=0.09\linewidth]{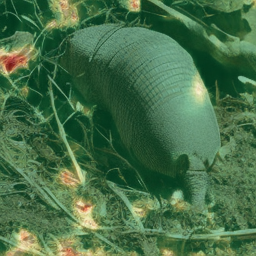}
& \includegraphics[width=0.09\linewidth,height=0.09\linewidth]{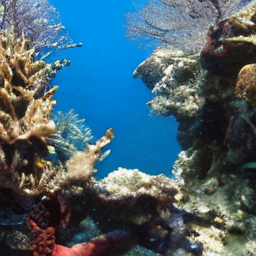} & \includegraphics[width=0.09\linewidth,height=0.09\linewidth]{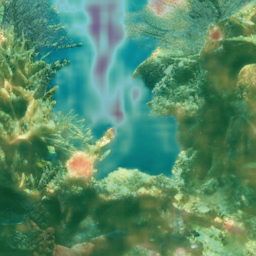} & \includegraphics[width=0.09\linewidth,height=0.09\linewidth]{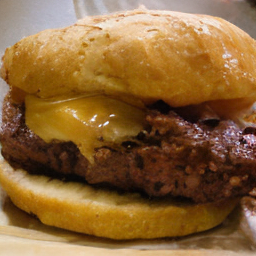}
& \includegraphics[width=0.09\linewidth,height=0.09\linewidth]{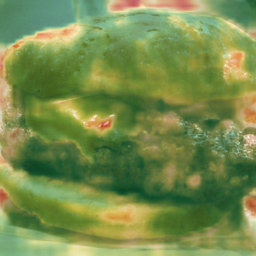} \\

\rll{SiT-SDE}
& \includegraphics[width=0.09\linewidth,height=0.09\linewidth]{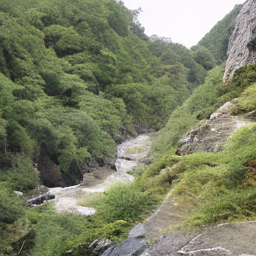} & \includegraphics[width=0.09\linewidth,height=0.09\linewidth]{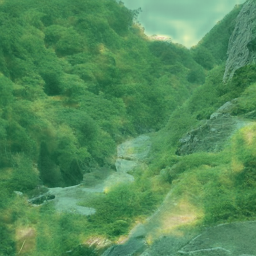} & \includegraphics[width=0.09\linewidth,height=0.09\linewidth]{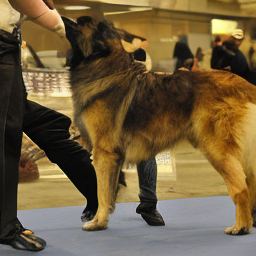}
& \includegraphics[width=0.09\linewidth,height=0.09\linewidth]{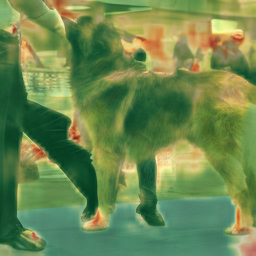} & \includegraphics[width=0.09\linewidth,height=0.09\linewidth]{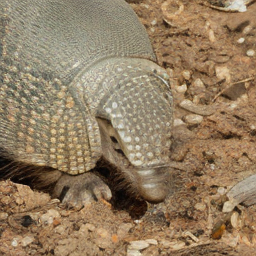} & \includegraphics[width=0.09\linewidth,height=0.09\linewidth]{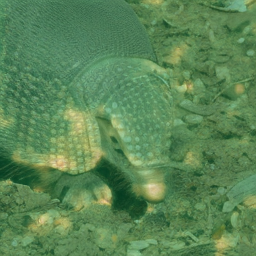}
& \includegraphics[width=0.09\linewidth,height=0.09\linewidth]{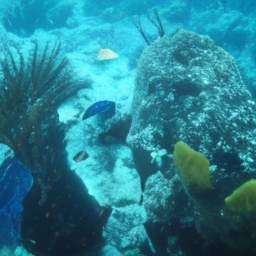} & \includegraphics[width=0.09\linewidth,height=0.09\linewidth]{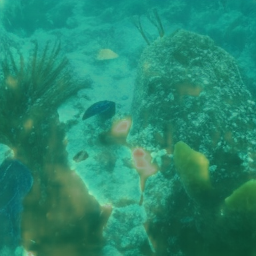} & \includegraphics[width=0.09\linewidth,height=0.09\linewidth]{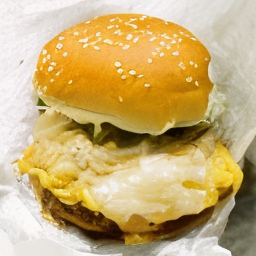}
& \includegraphics[width=0.09\linewidth,height=0.09\linewidth]{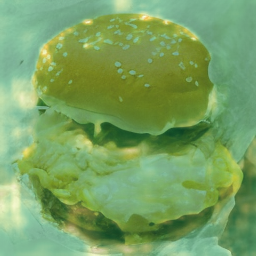} \\

\rll{SoFlow}
& \includegraphics[width=0.09\linewidth,height=0.09\linewidth]{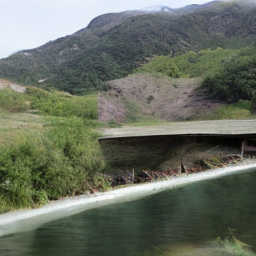} & \includegraphics[width=0.09\linewidth,height=0.09\linewidth]{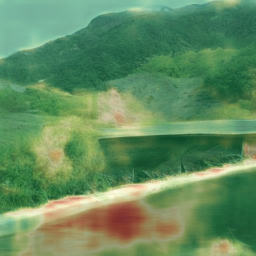} & \includegraphics[width=0.09\linewidth,height=0.09\linewidth]{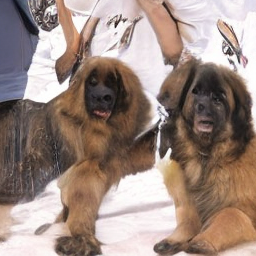}
& \includegraphics[width=0.09\linewidth,height=0.09\linewidth]{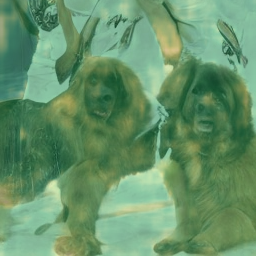} & \includegraphics[width=0.09\linewidth,height=0.09\linewidth]{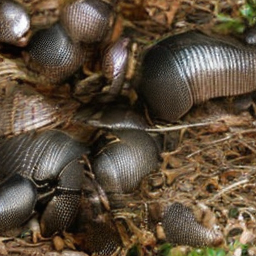} & \includegraphics[width=0.09\linewidth,height=0.09\linewidth]{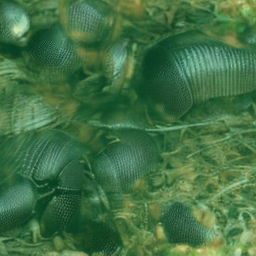}
& \includegraphics[width=0.09\linewidth,height=0.09\linewidth]{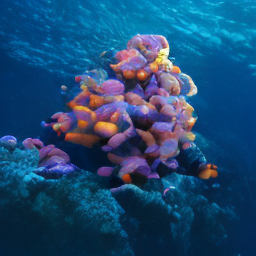} & \includegraphics[width=0.09\linewidth,height=0.09\linewidth]{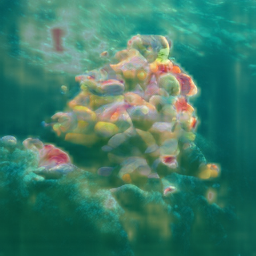} & \includegraphics[width=0.09\linewidth,height=0.09\linewidth]{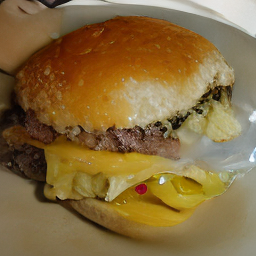}
& \includegraphics[width=0.09\linewidth,height=0.09\linewidth]{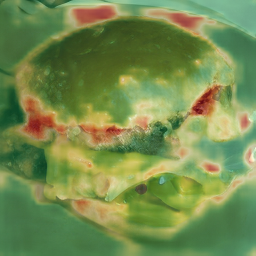} \\

\rll{DiT}
& \includegraphics[width=0.09\linewidth,height=0.09\linewidth]{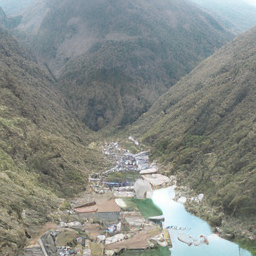} & \includegraphics[width=0.09\linewidth,height=0.09\linewidth]{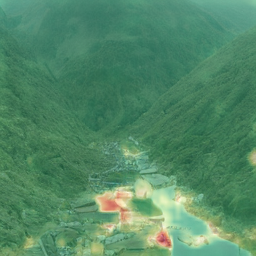} & \includegraphics[width=0.09\linewidth,height=0.09\linewidth]{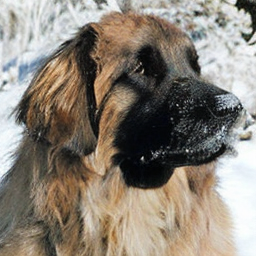}
& \includegraphics[width=0.09\linewidth,height=0.09\linewidth]{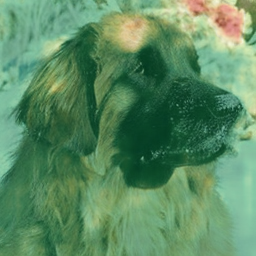} & \includegraphics[width=0.09\linewidth,height=0.09\linewidth]{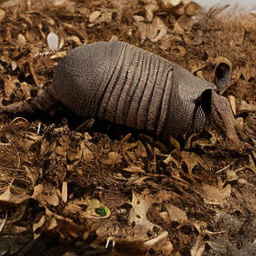} & \includegraphics[width=0.09\linewidth,height=0.09\linewidth]{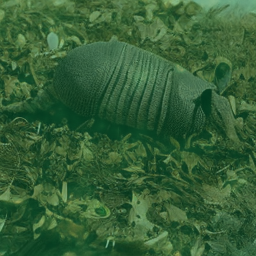}
& \includegraphics[width=0.09\linewidth,height=0.09\linewidth]{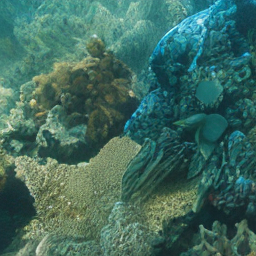} & \includegraphics[width=0.09\linewidth,height=0.09\linewidth]{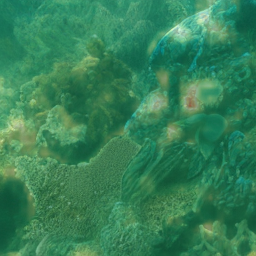} & \includegraphics[width=0.09\linewidth,height=0.09\linewidth]{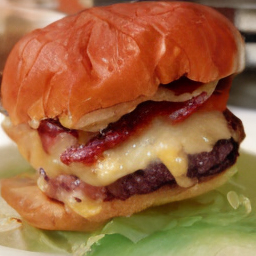}
& \includegraphics[width=0.09\linewidth,height=0.09\linewidth]{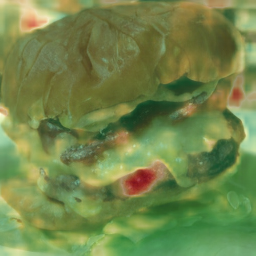} \\

\rll{SelfFlow}
& \includegraphics[width=0.09\linewidth,height=0.09\linewidth]{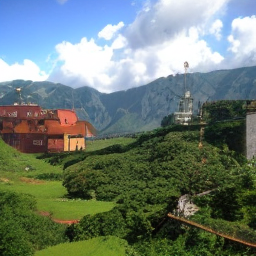} & \includegraphics[width=0.09\linewidth,height=0.09\linewidth]{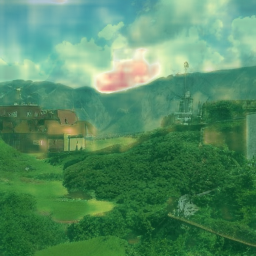} & \includegraphics[width=0.09\linewidth,height=0.09\linewidth]{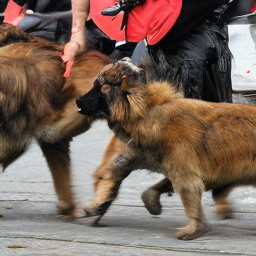}
& \includegraphics[width=0.09\linewidth,height=0.09\linewidth]{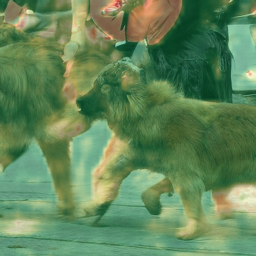} & \includegraphics[width=0.09\linewidth,height=0.09\linewidth]{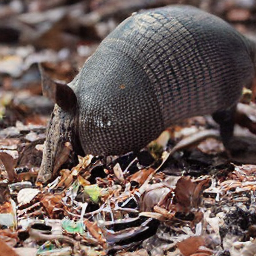} & \includegraphics[width=0.09\linewidth,height=0.09\linewidth]{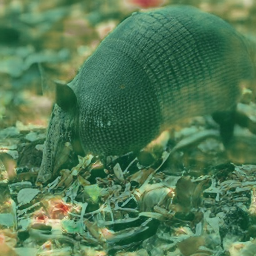}
& \includegraphics[width=0.09\linewidth,height=0.09\linewidth]{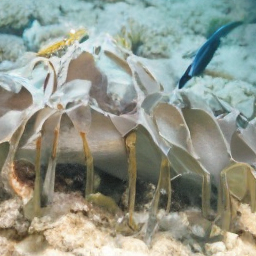} & \includegraphics[width=0.09\linewidth,height=0.09\linewidth]{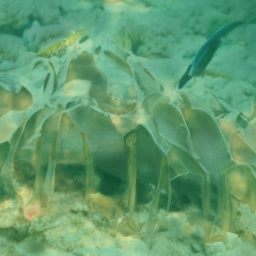} & \includegraphics[width=0.09\linewidth,height=0.09\linewidth]{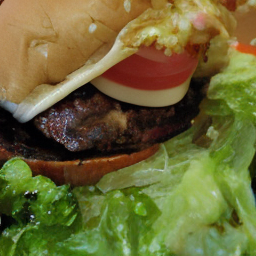}
& \includegraphics[width=0.09\linewidth,height=0.09\linewidth]{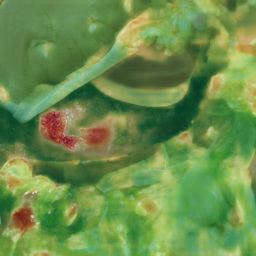} \\

\rll{PMF}
& \includegraphics[width=0.09\linewidth,height=0.09\linewidth]{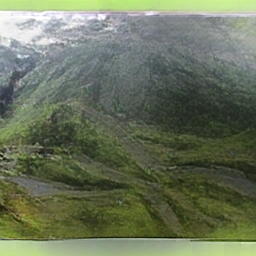} & \includegraphics[width=0.09\linewidth,height=0.09\linewidth]{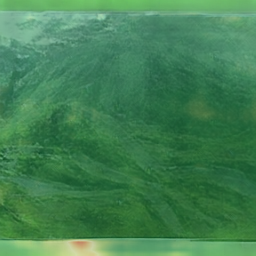} & \includegraphics[width=0.09\linewidth,height=0.09\linewidth]{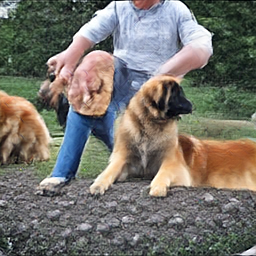}
& \includegraphics[width=0.09\linewidth,height=0.09\linewidth]{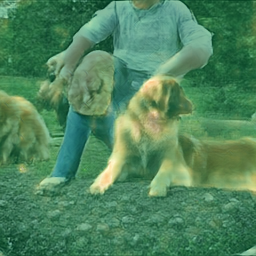} & \includegraphics[width=0.09\linewidth,height=0.09\linewidth]{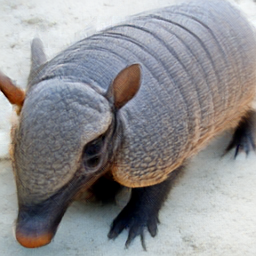} & \includegraphics[width=0.09\linewidth,height=0.09\linewidth]{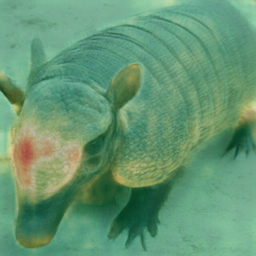}
& \includegraphics[width=0.09\linewidth,height=0.09\linewidth]{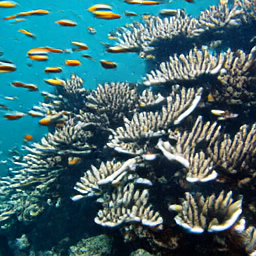} & \includegraphics[width=0.09\linewidth,height=0.09\linewidth]{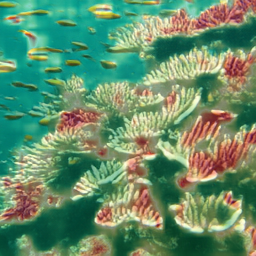} & \includegraphics[width=0.09\linewidth,height=0.09\linewidth]{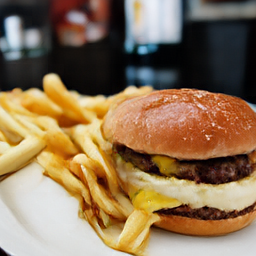}
& \includegraphics[width=0.09\linewidth,height=0.09\linewidth]{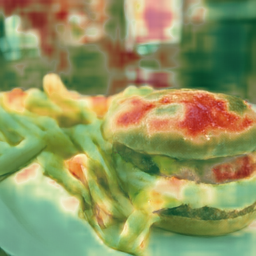} \\

\rll{LlamaGen}
& \includegraphics[width=0.09\linewidth,height=0.09\linewidth]{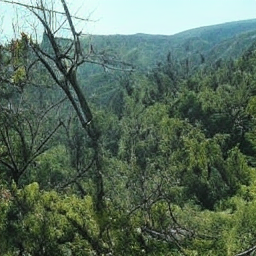} & \includegraphics[width=0.09\linewidth,height=0.09\linewidth]{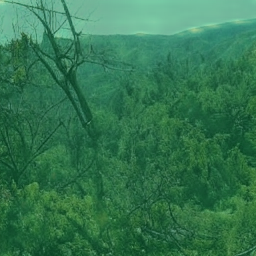} & \includegraphics[width=0.09\linewidth,height=0.09\linewidth]{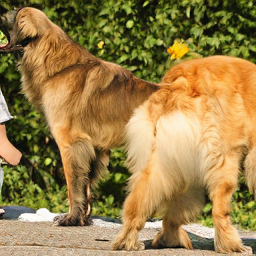}
& \includegraphics[width=0.09\linewidth,height=0.09\linewidth]{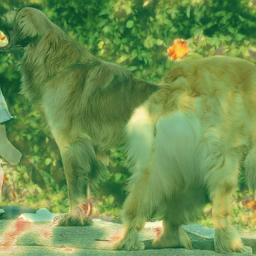} & \includegraphics[width=0.09\linewidth,height=0.09\linewidth]{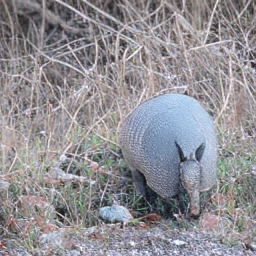} & \includegraphics[width=0.09\linewidth,height=0.09\linewidth]{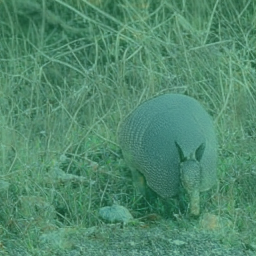}
& \includegraphics[width=0.09\linewidth,height=0.09\linewidth]{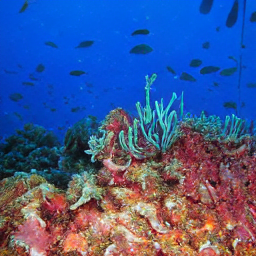} & \includegraphics[width=0.09\linewidth,height=0.09\linewidth]{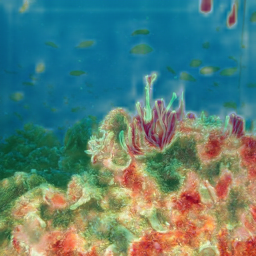} & \includegraphics[width=0.09\linewidth,height=0.09\linewidth]{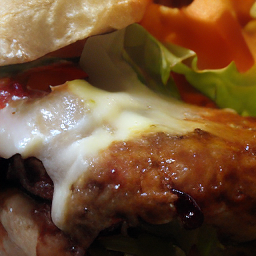}
& \includegraphics[width=0.09\linewidth,height=0.09\linewidth]{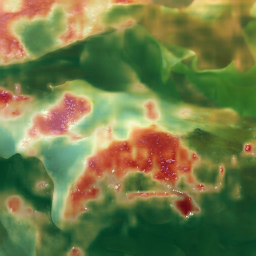} \\

\rll{RAR\_XXL}
& \includegraphics[width=0.09\linewidth,height=0.09\linewidth]{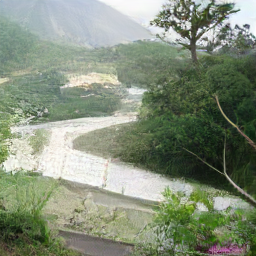} & \includegraphics[width=0.09\linewidth,height=0.09\linewidth]{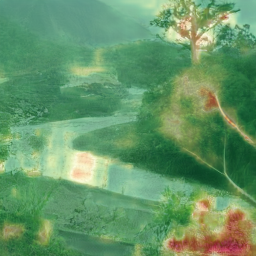} & \includegraphics[width=0.09\linewidth,height=0.09\linewidth]{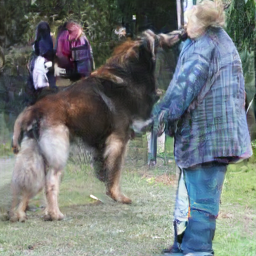}
& \includegraphics[width=0.09\linewidth,height=0.09\linewidth]{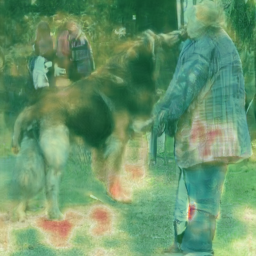} & \includegraphics[width=0.09\linewidth,height=0.09\linewidth]{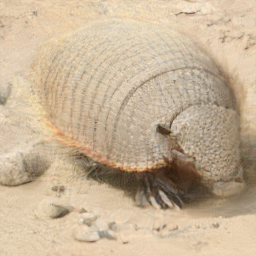} & \includegraphics[width=0.09\linewidth,height=0.09\linewidth]{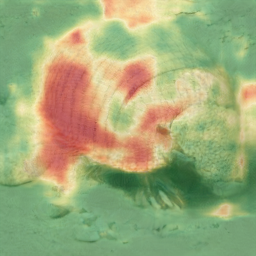}
& \includegraphics[width=0.09\linewidth,height=0.09\linewidth]{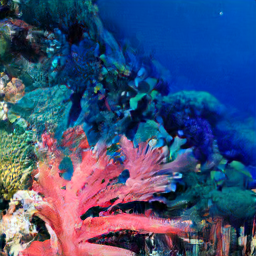} & \includegraphics[width=0.09\linewidth,height=0.09\linewidth]{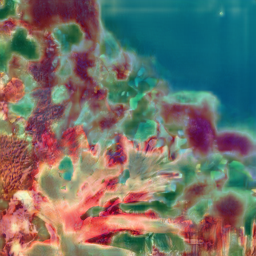} & \includegraphics[width=0.09\linewidth,height=0.09\linewidth]{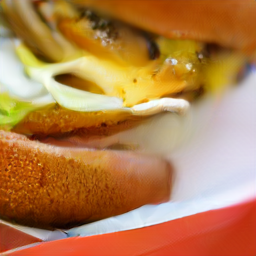}
& \includegraphics[width=0.09\linewidth,height=0.09\linewidth]{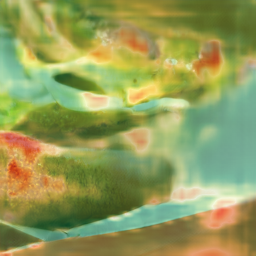} \\

\rll{BigGAN}
& \includegraphics[width=0.09\linewidth,height=0.09\linewidth]{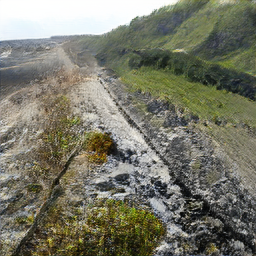} & \includegraphics[width=0.09\linewidth,height=0.09\linewidth]{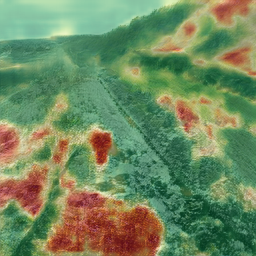} & \includegraphics[width=0.09\linewidth,height=0.09\linewidth]{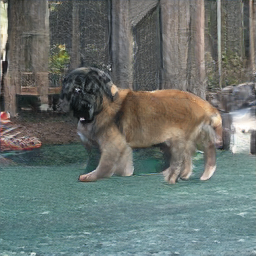}
& \includegraphics[width=0.09\linewidth,height=0.09\linewidth]{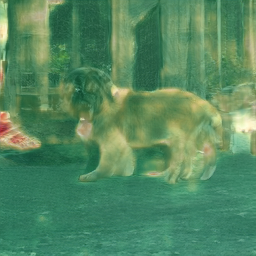} & \includegraphics[width=0.09\linewidth,height=0.09\linewidth]{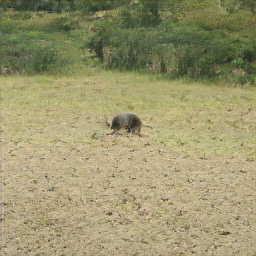} & \includegraphics[width=0.09\linewidth,height=0.09\linewidth]{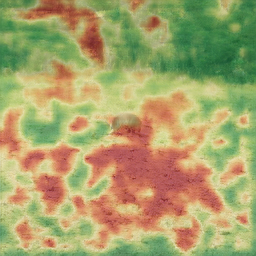}
& \includegraphics[width=0.09\linewidth,height=0.09\linewidth]{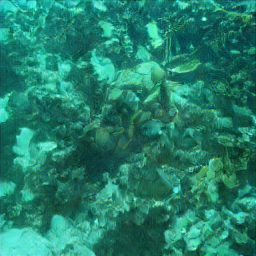} & \includegraphics[width=0.09\linewidth,height=0.09\linewidth]{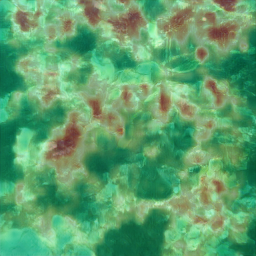} & \includegraphics[width=0.09\linewidth,height=0.09\linewidth]{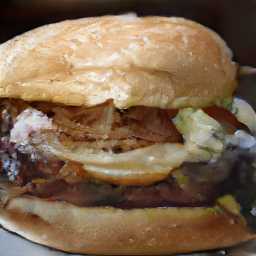}
& \includegraphics[width=0.09\linewidth,height=0.09\linewidth]{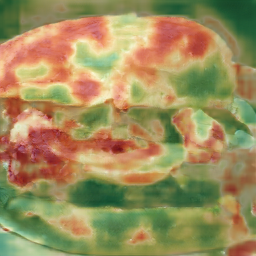} \\

\rll{StyleGAN-XL}
& \includegraphics[width=0.09\linewidth,height=0.09\linewidth]{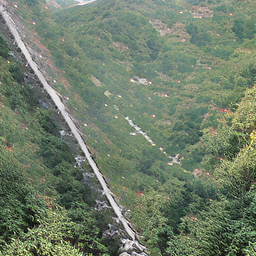} & \includegraphics[width=0.09\linewidth,height=0.09\linewidth]{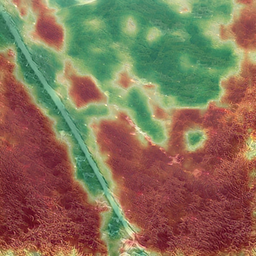} & \includegraphics[width=0.09\linewidth,height=0.09\linewidth]{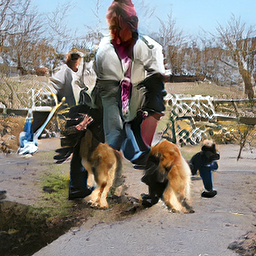}
& \includegraphics[width=0.09\linewidth,height=0.09\linewidth]{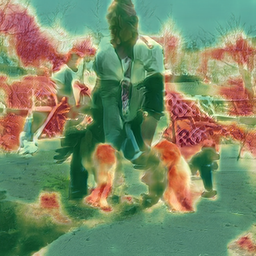} & \includegraphics[width=0.09\linewidth,height=0.09\linewidth]{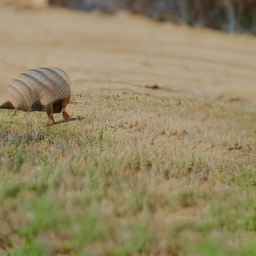} & \includegraphics[width=0.09\linewidth,height=0.09\linewidth]{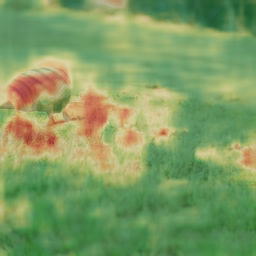}
& \includegraphics[width=0.09\linewidth,height=0.09\linewidth]{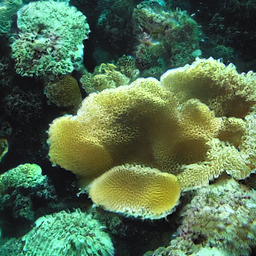} & \includegraphics[width=0.09\linewidth,height=0.09\linewidth]{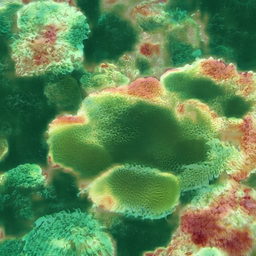} & \includegraphics[width=0.09\linewidth,height=0.09\linewidth]{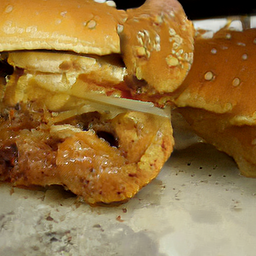}
& \includegraphics[width=0.09\linewidth,height=0.09\linewidth]{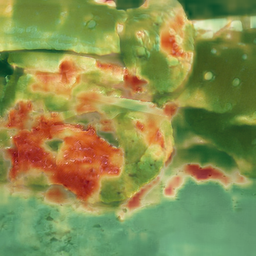} \\

\rll{ADM}
& \includegraphics[width=0.09\linewidth,height=0.09\linewidth]{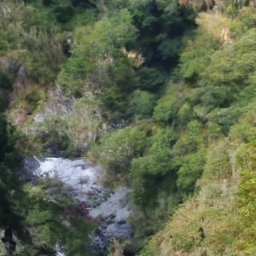} & \includegraphics[width=0.09\linewidth,height=0.09\linewidth]{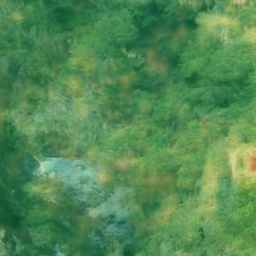} & \includegraphics[width=0.09\linewidth,height=0.09\linewidth]{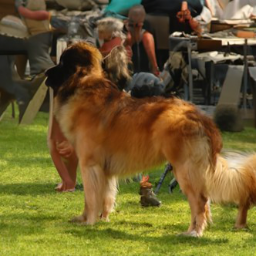}
& \includegraphics[width=0.09\linewidth,height=0.09\linewidth]{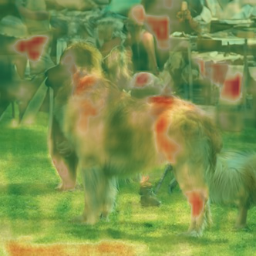} & \includegraphics[width=0.09\linewidth,height=0.09\linewidth]{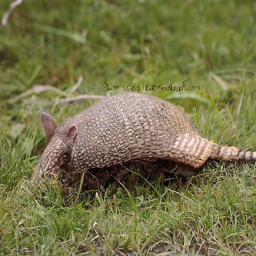} & \includegraphics[width=0.09\linewidth,height=0.09\linewidth]{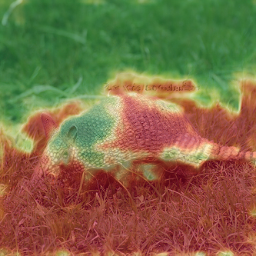}
& \includegraphics[width=0.09\linewidth,height=0.09\linewidth]{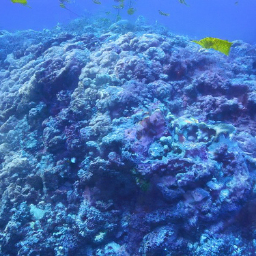} & \includegraphics[width=0.09\linewidth,height=0.09\linewidth]{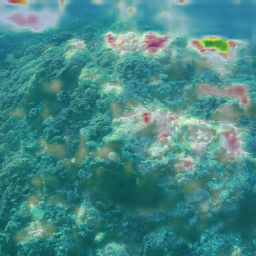} & \includegraphics[width=0.09\linewidth,height=0.09\linewidth]{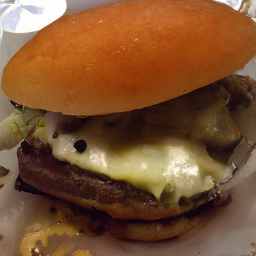}
& \includegraphics[width=0.09\linewidth,height=0.09\linewidth]{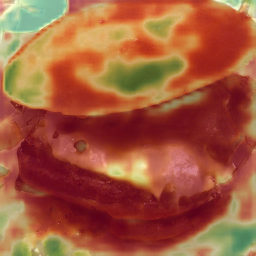} \\

\end{tabular}
\caption{\textbf{Samples across sources (rows) and ImageNet classes (columns).} Top row: real ImageNet; remaining rows: twelve generators. }
\label{fig:sample_grid}
\end{figure*}

Figure~\ref{fig:sample_grid} shows the texture-realism map $\mathbf{M}(I)$ overlaid on samples from every source (rows) across ten ImageNet classes (columns). Several patterns are consistent with our quantitative results. The real ImageNet row (top) is almost uniformly green across all classes, with few red regions, reflecting that real textures are judged consistent throughout the image. The amount and intensity of red responses grow from top to bottom, tracking the TRIG distances in Table 1 of main paper: the diffusion models (SiT, DiT) and VAR remain largely green with only scattered red, whereas the GAN-based generators (StyleGAN-XL, BigGAN) and autoregressive model (RAR) exhibit extensive red over textured surfaces. We also find red responses concentrate on \emph{foreground textured regions}, like the spines of the echidna, dog fur, and the surface of food, rather than on smooth backgrounds such as sky or defocused foliage, suggesting that the backbone does not systematically penalize depth-consistent background defocus, consistent with the foreground/background gap reported in Table 3 of main paper and Figure~\ref{fig:natural_vs_artificial_blur} in this Appendix. These visualizations indicate that our backbone assigns lower responses to textured foreground surfaces that frequently exhibit visible generative artifacts.

\end{document}